\documentclass[letterpaper]{article}
\usepackage{aaai2027}
\usepackage[hyphens]{url}  
\usepackage{graphicx} 
\usepackage{natbib}  
\usepackage{caption} 
\usepackage{amsmath}
\usepackage{amssymb}
\usepackage{booktabs}
\usepackage{multirow}
\usepackage{array}
\usepackage{tabularx}

\nocopyright

\newcommand{\method}{NEMSim}

\newcommand{\best}[1]{\textbf{#1}}
\newcommand{\second}[1]{\underline{#1}}

\title{NEMSim: Learning Control-Conditioned Multi-Event Physical Dynamics via Executable Event-Mechanism Priors}

\author{
Junsong Yu,\textsuperscript{\rm 1}
Junjie Xie,\textsuperscript{\rm 1}
Pengwei Liu,\textsuperscript{\rm 2}
Dong Ni\textsuperscript{\rm 1}\corresponding
}

\affiliations{
\textsuperscript{\rm 1}College of Integrated Circuits, Zhejiang University\\
\textsuperscript{\rm 2}College of Control Science and Engineering, Zhejiang University\\
\texttt{\{musthave,jj.xie,dni\}@zju.edu.cn}, \texttt{liupw@zju.edu.cn}
}

\begin{document}

\maketitle

\begin{abstract}
High-fidelity simulation of control-conditioned multi-event physical
systems is computationally expensive, especially across broad control
spaces and long trajectories.
In these systems, macroscopic evolution emerges from localized discrete events whose intensities and effects depend on process controls and evolving local states, while the available system knowledge is typically expressed as event-attribute descriptions.
Purely data-driven surrogates must infer these event effects from limited trajectory coverage, which can hinder generalization to unseen control regimes. Physics-guided methods instead primarily build on equation-level constraints or differentiable solvers rather than discrete event-rule priors. We therefore propose NEMSim
(\textbf{N}eural \textbf{E}vent-\textbf{M}echanism
\textbf{Sim}ulator), which compiles predefined event-attribute
descriptions into an executable transition structure linking
control-dependent event intensities, prior-guided mechanism attribution, and state-dependent responses.
To enable evaluation of control-conditioned multi-event dynamics with
explicit system knowledge, we construct a 3D KMC-based benchmark pairing high-fidelity trajectories with explicit event rules, standardized splits, and evaluation protocols.
Across three settings, NEMSim reduces Avg. RMSE by
$58.9\%$--$81.3\%$ relative to the strongest baseline in each setting.
It also remains best in the data-efficiency study with training-data
fractions down to $10\%$.
Mechanism analyses further show that these gains arise from
executable rule integration rather than prior access or architecture
alone.
\end{abstract}

\begin{figure*}[!t]
\centering
\includegraphics[
  width=1.0\textwidth,
  height=0.40\textheight
]{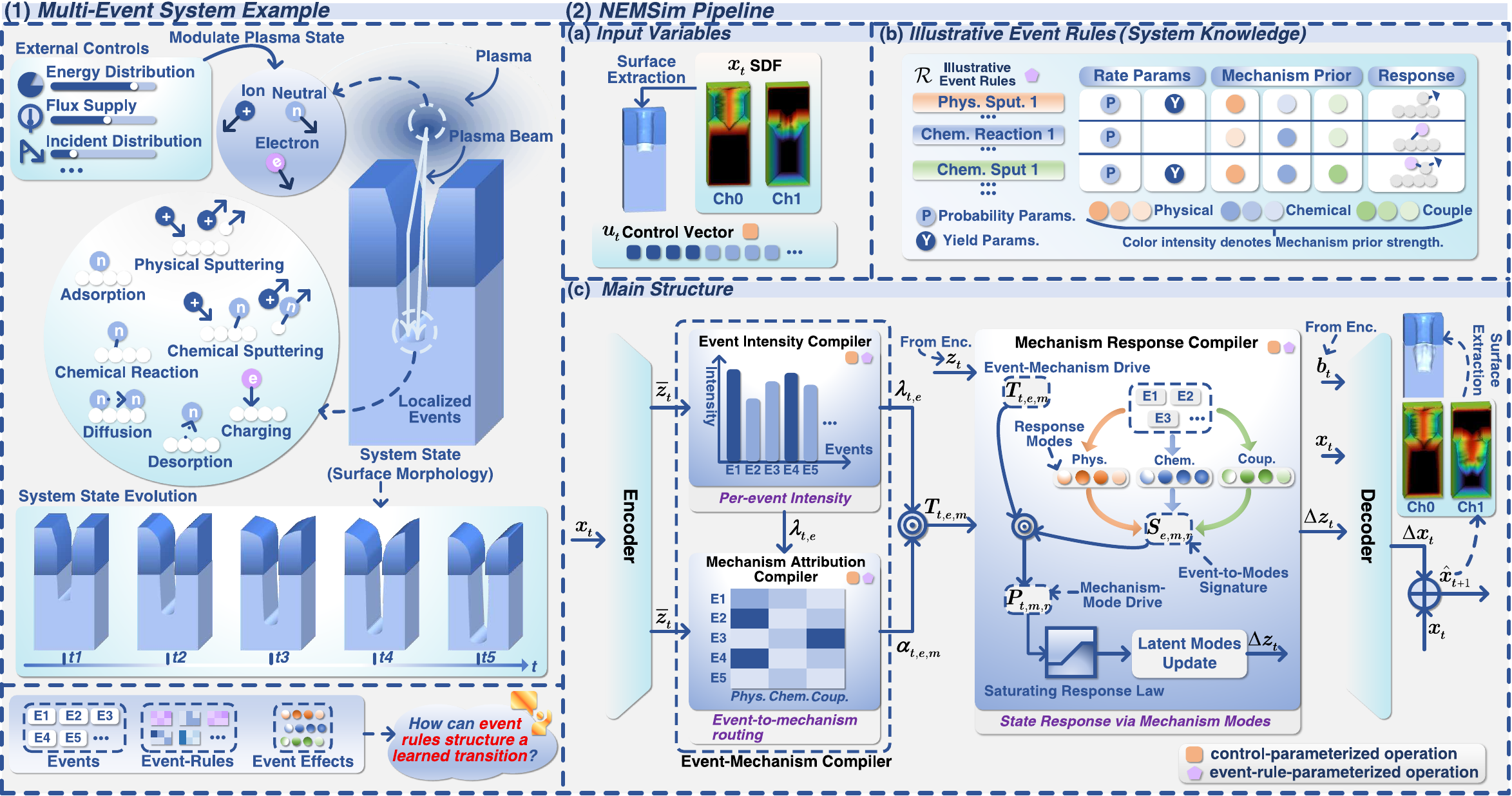}
\caption{
Overview of NEMSim and the representative control-conditioned multi-event system. The left panel uses plasma etching as an example to illustrate controlled multi-event system evolution. The right panel shows the NEMSim pipeline, which compiles $\mathbf{x}_{t}$, $\mathbf{u}_{t}$, and the event rule library into an executable event-mechanism transition structure to predict $\hat{\mathbf{x}}_{t+1}$.
}
\label{fig:overview}
\end{figure*}

\section{Introduction}

High-fidelity simulation is essential for the analysis and design of
control-conditioned multi-event physical systems, whose macroscopic
states evolve through repeated localized events, event competition, and
the accumulation of their effects over time~\cite{bortz1975new,gillespie1977exact,fichthorn1991theoretical}.
The intensities and state effects of these events depend jointly on
process controls and evolving local states.
Generating long trajectories across a broad control space therefore
requires repeatedly resolving numerous state-dependent local events, making
high-fidelity simulation computationally expensive.
This motivates neural surrogates that can predict long-horizon state
evolution across varying process controls.
We study this problem through Kinetic Monte Carlo (KMC)-based 3D surface
morphology simulation, using plasma etching as a representative
application~\cite{economou2000modeling,marcos2003monte}.

Purely data-driven surrogates, including sequence models~\cite{shi2015convolutional,bengio2015scheduled},
graph-based simulators~\cite{sanchez2020learning,pfaff2020learning,brandstetter2022message},
and neural operators~\cite{lu2021learning,li2020fourier,kovachki2023neural},
learn control-conditioned state transitions directly from high-fidelity
trajectories.
For multi-event dynamics, process controls and evolving local states
jointly determine event intensities, while dependencies among events
shape how their effects combine in each state transition.
The resulting control-state-event combinations are difficult to
cover with a finite set of expensive trajectories~\cite{li2021physics,brandstetter2022message}.
This limited coverage makes direct transition learning data-inefficient,
weakens generalization to held-out controls, and amplifies error
accumulation during long autoregressive rollouts.

Physics-guided learning offers a complementary way to reduce reliance
on trajectory coverage by incorporating prior knowledge into learned
dynamics.
Most existing formulations, however, are designed around equation-level
constraints or numerical structures, including governing-equation
residuals, conservation laws, known operators, and differentiable
solvers~\citep{raissi2019physics,li2021physics,belbute2020combining,
rackauckas2020universal,liu2024papm}.
For control-conditioned multi-event systems, available knowledge often
takes a different form: discrete event rules that specify control
dependencies, rate and yield relations, event dependencies, and
event-mechanism relations.
Such rule-level knowledge does not translate naturally into
residual-based constraints or differentiable-solver formulations.
The central challenge is therefore not merely to impose physical
regularization, but to convert discrete event-attribute priors into an
executable and learnable transition structure.

The effects of numerous events can be organized through a limited set of
mechanism families, while each mechanism may produce different local
responses depending on the current state.
This event-mechanism organization motivates \textbf{NEMSim}
(\textbf{N}eural \textbf{E}vent-\textbf{M}echanism
\textbf{Sim}ulator), a neural simulator for control-conditioned
multi-event physical state evolution.
NEMSim compiles predefined event rules into an executable
event-mechanism transition structure.
Given the current state and process controls, it computes event
intensities from rule-specified attributes, attributes event effects to
shared mechanism families under rule-defined priors, and realizes the
resulting mechanism drives through state-dependent response modes. 
This factorization preserves event-specific control dependence, enables
different events to share mechanism-level structure, and retains the
expressiveness required for local state-dependent responses.
We conduct experiments on a 3D KMC-based benchmark.
The results show that executable event-mechanism priors
support accurate 100-step rollouts under unseen controls,
temporal extrapolation and limited training data.

Overall, the main contributions of this work include:
\begin{itemize}
\item \textbf{Problem and method.}
We formulate the problem of learning control-conditioned multi-event
physical dynamics from discrete event-rule priors and introduce NEMSim,
which compiles these priors into an executable
event-mechanism transition structure.

\item \textbf{Benchmark and protocol.}
We construct a standardized 3D KMC-based benchmark for
control-conditioned multi-event dynamics, pairing high-dimensional
trajectories with explicit event rules and providing trajectory-level
splits and protocols for control interpolation, control extrapolation,
temporal extrapolation, data efficiency, and rule robustness.

\item \textbf{Evidence.}
Across 100-step rollouts, \method{} reduces Avg. RMSE by $63.3\%$,
$58.9\%$, and $81.3\%$ under control interpolation, control
extrapolation, and temporal extrapolation, respectively, relative to the
strongest baseline in each setting.
It also remains best with only $10\%$ of the training data, while mechanism analyses
further show that the gains arise from executable rule integration rather than prior access or architecture alone.
\end{itemize}

\section{Related Work}
\noindent\textbf{Purely Data-Driven Methods.}
Data-driven physical simulators use different representations to model
state evolution.
Recurrent models such as ConvLSTM~\citep{shi2015convolutional} and
PredRNN~\citep{wang2017predrnn} learn spatiotemporal dependencies from
gridded sequences through memory propagation.
Graph-based models, including GNS~\citep{sanchez2020learning},
MeshGraphNet~\citep{pfaff2020learning}, and
EAGLE~\citep{janny2023eagle}, model particle or mesh interactions through message passing or attention.
Neural operators such as DeepONet~\citep{lu2021learning},
FNO~\citep{li2020fourier}, U-NO~\citep{rahman2022uno}, and
CNO~\citep{raonic2023convolutional} learn mappings between physical
fields, while GNOT~\citep{hao2023gnot} and
Transolver~\citep{wu2024transolver} further use attention to capture
long-range dependencies.
These methods typically learn direct mappings from observed states and
controls to future states.

\noindent\textbf{Physics-Guided Methods.}
Physics-guided learning embeds prior structure into neural prediction.
PINNs and PINOs impose PDE residuals or equation-level operator
constraints~\citep{raissi2019physics,li2021physics}, while
differentiable-physics and Universal Differential Equation frameworks
embed numerical solvers or known dynamics terms~\citep{belbute2020combining,rackauckas2020universal}.
Chemical Reaction Neural Networks encode mass-action and Arrhenius
kinetics~\citep{ji2021autonomous}, while Mechanistic Neural Networks
learn explicit differential-equation representations~\citep{pervez2024mechanistic}.
PRIMME~\citep{yan2022novel} encodes local transition constraints for
grain growth, while PAPM~\citep{liu2024papm} incorporates operating
conditions and conservation constraints into process surrogates.
These methods primarily express prior knowledge through equations,
kinetic laws, solver structures, or regularization, rather than
heterogeneous event-rule libraries.

\noindent\textbf{Benchmarks for Physical Learning.}
PDEBench~\citep{takamoto2022pdebench},
APEBench~\citep{koehler2024apebench}, and
CFDBench~\citep{luo2023cfdbench} standardize learning from PDE and CFD
trajectories.
A recent phase-field benchmark provides microstructure-evolution
trajectories~\citep{rieger2024setting}.
The underlying systems in these benchmarks are primarily specified
through governing equations, boundary conditions, geometries, or
numerical operators, rather than heterogeneous control-dependent
multi-event descriptions.

\section{Method}

\method{} learns control-conditioned multi-event state transitions by
compiling an Event Rules Library into an executable event-mechanism
transition structure.
We first formulate the one-step prediction problem and formalize the
Event Rules Library, and then describe the architecture that realizes
this factorization.

\subsection{Problem Setting and Notation}

Let $\mathcal{X}$ denote the system-state space and $\mathcal{U}$ the
process-control space.
At transition step $t$, the current state is
$\mathbf{x}_t\in\mathcal{X}$ and the applied process control is
$\mathbf{u}_t\in\mathcal{U}$.
We formulate the task as learning the one-step
control-conditioned transition from
$(\mathbf{x}_t,\mathbf{u}_t)$ to $\mathbf{x}_{t+1}$.
In our benchmark, $\mathbf{x}_t$ is represented as a two-channel 3D
signed-distance field of the evolving morphology.
The control vector $\mathbf{u}_t\in\mathbb{R}^{14}$ contains
supply-rate, energy-scale, and incidence-orientation controls.

\method{} incorporates an Event Rules Library
$\mathcal{R}=\{\mathbf{r}_e\}_{e=1}^{N_e}$ and organizes event effects
across a predefined set of mechanism families
$\mathcal{M}=\{1,\ldots,M\}$, where $N_e$ and $M$ denote the numbers of
event types and mechanism families, respectively.
Each event rule is represented as
\begin{equation}
    \mathbf{r}_e
    =
    \left(
    \mathcal{C}_e,\,
    p_e,\,
    Y_e,\,
    \mathcal{D}_e,\,
    g_e
    \right),
    \label{eq:event_rule_definition}
\end{equation}
where $\mathcal{C}_e$, $p_e$, $Y_e$, $\mathcal{D}_e$, and $g_e$
respectively specify the control bindings, probability relation, yield
relation, product-mediated dependencies, and event-mechanism role.
The role $g_e$ defines the fixed event-mechanism prior
$\boldsymbol{\alpha}^{0}_{e,m}$.
The rule attributes are further encoded as an event descriptor
$\boldsymbol{\phi}_e$ for subsequent mechanism-attribution and
state-response computations.
The Event Rules Library provides fixed, system-specific priors that
structure the forward transition without event- or mechanism-level
supervision.
Complete rule definitions, executable attributes, and their
instantiation for the benchmark system are provided in the
\textbf{supplementary material}
(Secs.~S3.1, S3.2, and S3.3).

Given fixed structural inputs $\mathcal{R}$ and $\mathcal{M}$,
\method{} learns
\begin{equation}
    \hat{\mathbf{x}}_{t+1}
    =
    \mathcal{F}_{\theta}
    \left(
    \mathbf{x}_t,\mathbf{u}_t;
    \mathcal{R},\mathcal{M}
    \right),
    \qquad
    \mathcal{F}_{\theta}:
    \mathcal{X}\times\mathcal{U}\rightarrow\mathcal{X},
    \label{eq:learned_transition}
\end{equation}
where $\mathcal{R}$ and $\mathcal{M}$ provide fixed structural priors
for the transition, while the neural parameters $\theta$ are learned
from state trajectories.

Given
$\mathcal{S}_{\mathrm{train}}
=
\{(\mathbf{x}_t^{(i)},\mathbf{u}_t^{(i)},
\mathbf{x}_{t+1}^{(i)})\}_{i=1}^{N}$,
the model is trained using only one-step state-transition supervision:
\begin{equation}
    \min_{\theta}
    \mathcal{L}_{\mathrm{pred}}(\theta)
    =
    \frac{1}{N}
    \sum_{i=1}^{N}
    \left\|
    \hat{\mathbf{x}}_{t+1}^{(i)}
    -
    \mathbf{x}_{t+1}^{(i)}
    \right\|_{\mathcal{X}}^{2},
    \label{eq:training_loss}
\end{equation}
where $\|\cdot\|_{\mathcal{X}}^{2}$ denotes the mean squared error over
all elements of the normalized state fields.

During long-horizon evaluation, the learned transition is recursively
applied from the true initial state under the trajectory-specific
control:
\begin{equation}
    \hat{\mathbf{x}}_{t+1}
    =
    \mathcal{F}_{\theta}
    \left(
    \hat{\mathbf{x}}_t,\mathbf{u}_t;
    \mathcal{R},\mathcal{M}
    \right),
    \qquad
    \hat{\mathbf{x}}_0=\mathbf{x}_0,
    \label{eq:rollout_transition}
\end{equation}

The transition combines deterministic rule-derived quantities
with learned rule-conditioned corrections and state-dependent responses.
Its learnable components are optimized solely through
$\mathcal{L}_{\mathrm{pred}}$, without intermediate supervision.

\subsection{Model Architecture}

\begin{figure}[!htbp]
    \centering
    \includegraphics[width=\columnwidth]{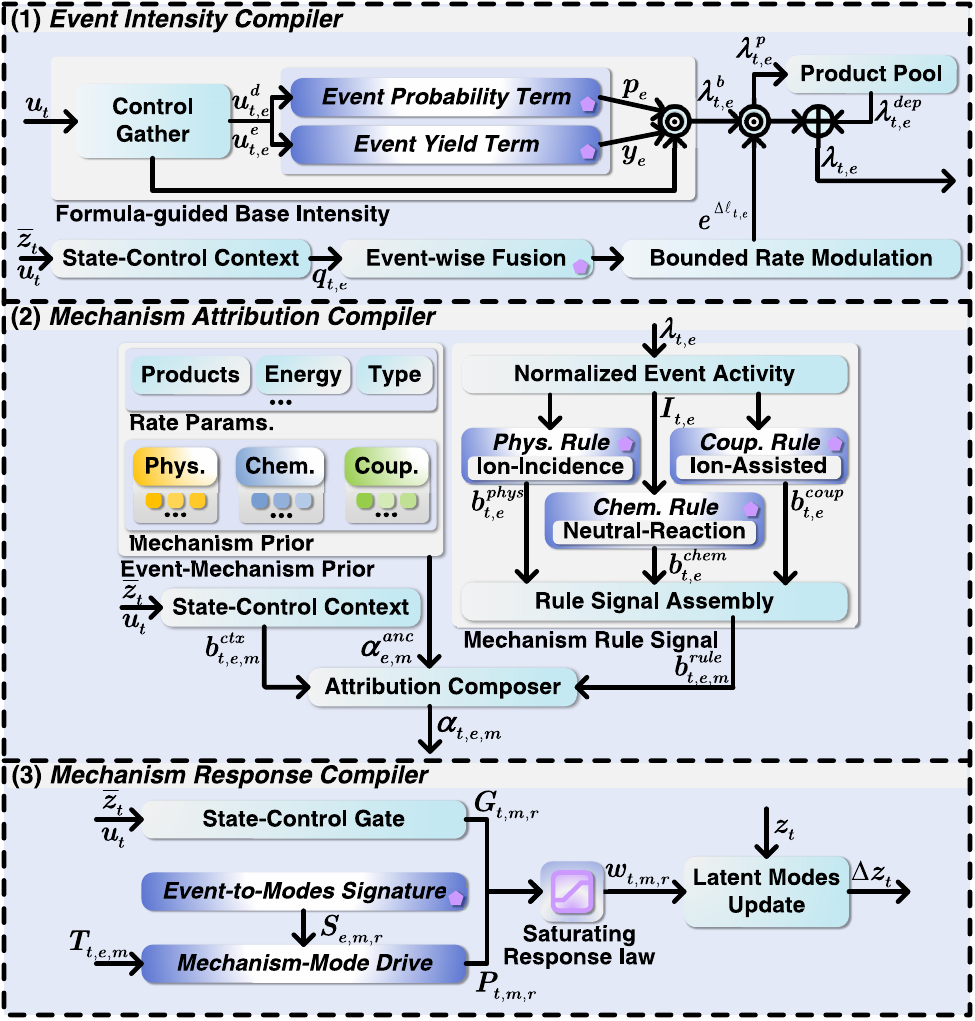}
    \caption{
Detailed architecture of the EIC, MAC and MRC modules in \method{}.
Purple pentagons mark operations parameterized by event-rule attributes.
}
\label{fig:compiler_details}
\end{figure}

As illustrated in Fig.~\ref{fig:overview}, NEMSim predicts the next state through event-intensity, mechanism-attribution, and state-response compilation. Fig.~\ref{fig:compiler_details} details the internal computations of the three compilation stages.

\paragraph{State Encoder.}
The State Encoder maps the current state $\mathbf{x}_t$ to a latent
field, a decoder support feature, and a compact state context:
\begin{equation}
    (\mathbf{z}_t,\mathbf{b}_t,\bar{\mathbf{z}}_t)
    =
    \mathrm{SE}(\mathbf{x}_t),
    \label{eq:se}
\end{equation}
where $\mathbf{z}_t$ supports spatial response computation,
$\bar{\mathbf{z}}_t$ conditions event and mechanism compilation, and
$\mathbf{b}_t$ is reused by the decoder.

\paragraph{Event-Mechanism Compiler.}
The Event-Mechanism Compiler (EMC) compiles event intensities and
event-to-mechanism attribution from the compact state context, process
control, and Event Rules Library:
\begin{equation}
\begin{gathered}
    (\boldsymbol{\lambda}_{t,e},\boldsymbol{\alpha}_{t,e,m})
    =
    \mathrm{EMC}
    (\bar{\mathbf{z}}_{t},\mathbf{u}_{t};\mathcal{R}),\\
    \mathbf{T}_{t,e,m}
    =
    \boldsymbol{\lambda}_{t,e}
    \odot
    \boldsymbol{\alpha}_{t,e,m}.
\end{gathered}
\label{eq:emc_overview}
\end{equation}
Here, $\boldsymbol{\lambda}_{t,e}$ denotes the event-intensity vector,
and $\boldsymbol{\alpha}_{t,e,m}$ denotes the event-to-mechanism
attribution.
Their element-wise product forms the event-mechanism drive
$\mathbf{T}_{t,e,m}$, which is passed to the Mechanism Response
Compiler.

\noindent\textbf{Event Intensity Compilation.}
As shown in Fig.~\ref{fig:compiler_details}(1), EIC determines how
strongly each event acts under the current state and process control.
For each primary event, the control bindings $\mathcal{C}_e$ select the
event-specific supply, energetic, and directional controls
$\mathbf{u}^{f}_{t,e}$, $\mathbf{u}^{e}_{t,e}$, and
$\mathbf{u}^{d}_{t,e}$ from $\mathbf{u}_t$.
The rule-defined base intensity
$\boldsymbol{\lambda}^{b}_{t,e}$ is
\begin{equation}
    \boldsymbol{\lambda}^{b}_{t,e}
    =
    \mathbf{u}^{f}_{t,e}
    \odot
    p_e(\mathbf{u}^{e}_{t,e},\mathbf{u}^{d}_{t,e})
    \odot
    Y_e(\mathbf{u}^{e}_{t,e},\mathbf{u}^{d}_{t,e}).
    \label{eq:base_rate}
\end{equation}
The differentiable probability and yield relations $p_e(\cdot)$ and
$Y_e(\cdot)$ are specified by event rule $\mathbf{r}_e$ and instantiated
from established physics-based and empirical rate/yield models. Their complete functional forms are provided in the
\textbf{supplementary material} (Sec.~S3.4).

To account for transition-specific variation not captured by the fixed
rules, EIC applies a bounded correction in log-rate space:
\begin{equation}
\begin{gathered}
    \Delta\boldsymbol{\ell}_{t,e}
    =
    \delta_{\max}
    \tanh\!\left(
    h_{\mathrm{rate}}(\mathbf{q}_{t,e})
    \right),\\
    \boldsymbol{\lambda}^{p}_{t,e}
    =
    \boldsymbol{\lambda}^{b}_{t,e}
    \odot
    \exp\!\left(
    \Delta\boldsymbol{\ell}_{t,e}
    \right), 
\end{gathered}
\label{eq:primary_intensity}
\end{equation}
where $\mathbf{q}_{t,e}$ is the rule-conditioned state-control context,
$h_{\mathrm{rate}}$ predicts the log-rate correction, and
$\delta_{\max}$ bounds its magnitude.
The resulting $\boldsymbol{\lambda}^{p}_{t,e}$ is a nonnegative primary-event intensity with bounded deviation from the rule-defined base.

For product-mediated events, the Product Pool first aggregates the
products generated by upstream events and then uses their availability
to determine the dependency-driven intensity
$\boldsymbol{\lambda}^{\mathrm{dep}}_{t,e}$ according to
$\mathcal{D}_e$.
This represents the pathway from product-generating events
to downstream product-consuming events.
The final event intensity is
$\boldsymbol{\lambda}_{t,e}
=
\boldsymbol{\lambda}^{p}_{t,e}
+
\boldsymbol{\lambda}^{\mathrm{dep}}_{t,e}$,
which is passed to MAC.
Complete kinetic, bounded-correction, and Product Pool formulations are
provided in the \textbf{supplementary material}
(Secs.~S3.4 and S4.1).

\noindent\textbf{Mechanism Attribution Compilation.}
EIC determines the strengths of the event effects, while MAC determines
how these effects are distributed among the mechanism families.
As shown in Fig.~\ref{fig:compiler_details}(2), MAC assigns the compiled
event intensities to physical, chemical, and coupled mechanisms.

MAC first combines the fixed event-mechanism prior
$\boldsymbol{\alpha}^{0}_{e,m}$ with how well the event attributes match
each mechanism family.
This produces the default attribution
$\boldsymbol{\alpha}^{\mathrm{anc}}_{e,m}$.
The prior specifies the initial mechanism preference, while the default
attribution incorporates the event attributes before state-control
adaptation.

MAC then bounds the event intensities and computes a rule-based
correction $\mathbf{b}^{\mathrm{rule}}_{t,e,m}$:
\begin{equation}
    \mathbf{I}_{t,e}
    =
    \mathrm{Bnd}(\boldsymbol{\lambda}_{t,e}),
    \quad
    \mathbf{b}^{\mathrm{rule}}_{t,e,m}
    =
    \mathrm{RuleCorr}_{m}
    \left(\mathbf{I}_{t,e};\mathbf{r}_{e}\right).
    \label{eq:rule_based_routing}
\end{equation}
Here, $\mathbf{I}_{t,e}$ is the bounded event-intensity signal.
Bounding reduces intensity-scale differences before mechanism
attribution.
$\mathrm{RuleCorr}_{m}(\cdot)$ uses mechanism-relevant attributes and
dependency relations in $\mathbf{r}_e$ to produce
$\mathbf{b}^{\mathrm{rule}}_{t,e,m}$.

A lightweight context correction
$\mathbf{b}^{\mathrm{ctx}}_{t,e,m}$ further uses the event attributes,
compact state context, and process control.
The final attribution is
\begin{equation}
\begin{gathered}
    \boldsymbol{\delta}_{t,e,m}
    =
    \tanh\!\left(
    \mathbf{b}^{\mathrm{rule}}_{t,e,m}
    +
    \kappa\mathbf{b}^{\mathrm{ctx}}_{t,e,m}
    \right),\\
    \boldsymbol{\alpha}_{t,e,m}
    =
    \operatorname{Normalize}_{m}
    \left(
    \boldsymbol{\alpha}^{\mathrm{anc}}_{e,m}
    \odot
    \exp\!\left(
    \rho\boldsymbol{\delta}_{t,e,m}
    \right)
    \right).
\end{gathered}
\label{eq:mechanism_attribution}
\end{equation}
Here, $\boldsymbol{\delta}_{t,e,m}$ is the bounded attribution
correction, $\kappa$ scales the context contribution, and $\rho$ limits
the overall update.
Normalization produces nonnegative attributions that sum to one over
$m$.
This preserves the rule-defined mechanism preference while allowing
the current state and process control to adjust it.
Detailed default-attribution, bounded-intensity, and correction
formulations are provided in the \textbf{supplementary material}
(Secs.~S3.5 and S4.2).

\paragraph{Mechanism Response Compiler.}
MRC converts the event-mechanism drivers
$\mathbf{T}_{t,e,m}$ into a latent state increment.
As shown in Fig.~\ref{fig:compiler_details}(3), it organizes
event-level drivers into mechanism-specific response modes, maps them
to bounded amplitudes, and combines them with state-dependent spatial
response fields.

\noindent\textbf{Event-to-Modes Signature.}
Events routed through the same mechanism family can induce different
response modes.
MRC maps each event descriptor $\boldsymbol{\phi}_e$ to a distribution
over the response modes of every mechanism family:
\begin{equation}
    \mathbf{S}_{e,m,r}
    =
    \operatorname{softmax}_{r}
    \left(
    h_{\mathrm{sig}}(\boldsymbol{\phi}_{e})_{m,r}
    \right).
    \label{eq:event_basis_signature}
\end{equation}
Here, $h_{\mathrm{sig}}$ produces mode-assignment logits, and
$\mathbf{S}_{e,m,r}$ specifies how the driver of event $e$ is
distributed across response modes under mechanism $m$.
The softmax produces nonnegative weights that sum to one over $r$,
retaining event-specific response preferences within shared mechanisms.

\noindent\textbf{Mechanism-Mode Drive.}
The event-mechanism drivers are aggregated into mechanism-mode drives:
\begin{equation}
    \mathbf{P}_{t,m,r}
    =
    \sum_{e=1}^{N_e}
    \mathbf{T}_{t,e,m}
    \odot
    \mathbf{S}_{e,m,r}.
    \label{eq:event_basis_drive}
\end{equation}
Here, $\mathbf{P}_{t,m,r}\geq 0$ represents the aggregated drive of
response mode $r$ under mechanism $m$.
This aggregation compresses the event-level drivers into a compact
mechanism-mode representation.

\noindent\textbf{Saturating Response Law.}
To model finite response capacity under state constraints and event
competition, MRC predicts a state-control gate and a saturating mode
amplitude:
\begin{equation}
\begin{gathered}
    \mathbf{G}_{t,m,r}
    =
    \sigma\!\left(
    h_{\mathrm{sc}}
    ([\bar{\mathbf{z}}_{t},\mathbf{u}_{t}])_{m,r}
    \right),\\
    \mathbf{w}_{t,m,r}
    =
    A_{m,r}
    \frac{\mathbf{P}_{t,m,r}}
    {K_{m,r}+\mathbf{P}_{t,m,r}+\epsilon}
    \odot
    \mathbf{G}_{t,m,r}.
\end{gathered}
\label{eq:response_weight}
\end{equation}
Here, $h_{\mathrm{sc}}$ predicts the realizability gate,
$A_{m,r}>0$ is the response capacity, and $K_{m,r}>0$ is the
saturation scale.
The resulting $\mathbf{w}_{t,m,r}$ is a bounded,
state-control-dependent amplitude for mechanism mode $(m,r)$.

\noindent\textbf{Spatial Response Composition.}
Each mode generates a normalized spatial response field from the current
latent state:
\begin{equation}
    \mathbf{B}_{t,m,r}
    =
    h_{m,r}(\mathbf{z}_{t}),
    \label{eq:latent_basis}
\end{equation}
where $\mathbf{B}_{t,m,r}$ determines the state-dependent response
pattern.
The latent increment is then
\begin{equation}
    \Delta\mathbf{z}_{t}
    =
    \sum_{m=1}^{M}
    \sum_{r=1}^{R_b}
    \mathbf{w}_{t,m,r}
    \odot
    \mathbf{B}_{t,m,r},
    \label{eq:latent_update}
\end{equation}
where $R_b$ is the number of response modes per mechanism.
Thus, $\mathbf{P}_{t,m,r}$ determines which modes are driven,
$\mathbf{w}_{t,m,r}$ determines their realized amplitudes, and
$\mathbf{B}_{t,m,r}$ determines their spatial response patterns.
Event-descriptor construction, network parameterization, field
normalization, and tensor dimensions are provided in the
\textbf{supplementary material} (Secs.~S3.2 and S4.3).

\paragraph{State Increment Decoder.}
The decoder maps the mechanism-induced latent response back to the
state space through a residual transition:
\begin{equation}
\begin{gathered}
    \tilde{\mathbf{z}}_{t+1}
    =
    \mathbf{z}_{t}+\Delta\mathbf{z}_{t},
    \quad
    \Delta\mathbf{x}_{t}
    =
    \mathrm{Dec}
    (\tilde{\mathbf{z}}_{t+1},\mathbf{b}_{t},\mathbf{x}_{t}),\\
    \hat{\mathbf{x}}_{t+1}
    =
    \mathbf{x}_{t}+\Delta\mathbf{x}_{t}.
\end{gathered}
\label{eq:decoder}
\end{equation}
Here, $\mathbf{b}_{t}$ provides support features, while the
residual formulation focuses prediction on transition-induced changes.

\section{Experiments}
We evaluate \method{} on a 3D KMC-based benchmark under three
100-step rollout settings: control interpolation, control extrapolation and temporal extrapolation.
We further examine prior-access and structure controls, training-data
efficiency, robustness to perturbed event rules, mechanism-aligned
counterfactual analysis, and component ablations.

\subsection{Experimental Setup}

\noindent\textbf{Benchmark and protocol.}
The benchmark trajectories are generated using a reference
particle-based Monte Carlo feature-profile simulator.
The dataset contains 500 trajectories, each comprising 101 two-channel
3D signed-distance fields and 100 consecutive transitions on a
$127\times54\times54$ grid.
Each supervised sample is represented as
$(\mathbf{x}_{t},\mathbf{u}_{t},\mathbf{x}_{t+1})$, where
$\mathbf{u}_{t}\in\mathbb{R}^{14}$ contains the process controls and
remains fixed within each trajectory while varying across trajectories.
All splits are constructed at the trajectory level, with no transition
from the same trajectory shared across training, validation, and test
sets.
The Event Rules Library is fixed before dataset splitting and contains
no trajectory-specific event realizations, target states, split
information, or evaluation statistics.
Models are trained with one-step state-transition supervision and
evaluated through 100-step autoregressive rollouts without
ground-truth correction.
Complete benchmark construction, evaluation protocol, and Event Rules
Library details are provided in the \textbf{supplementary material}
(Secs.~S2.1--S2.5 and S3.1).

\noindent\textbf{Evaluation Settings.}
We evaluate all models under three 100-step rollout settings.
\textbf{Control-ID} measures generalization to unseen control
configurations within the training control domain.
\textbf{Control-OOD} evaluates extrapolation along the
Ar$^{+}$ and Cl$^{+}$ incident-flux dimensions selected
through final-state sensitivity analysis, with their boundary
regions held out from training.
We consider OOD-Ar$^{+}$, OOD-Cl$^{+}$, and OOD-Joint, where
the latter holds out boundary regions along both control
dimensions.
\textbf{Temporal Extrapolation} restricts training to
transitions from only the first $50\%$ or $60\%$ of each
training trajectory and evaluates full-horizon prediction on
trajectory-disjoint test cases.
Detailed sensitivity analysis and split construction are provided in the \textbf{supplementary material} (Sec.~S2.5).

\noindent\textbf{Baselines.}
We compare \method{} with five representative spatiotemporal and
neural-operator baselines adapted to 3D state-transition prediction:
ConvLSTM~\cite{shi2015convolutional}, FNO~\cite{li2020fourier},
U-NO~\cite{rahman2022uno}, CNO~\cite{raonic2023convolutional}, and
DeepONet~\cite{lu2021learning}.
The standard baselines receive the same state-control inputs
$(\mathbf{x}_{t},\mathbf{u}_{t})$ and use identical trajectory-level
splits, one-step supervision, and autoregressive evaluation protocols.
To isolate the effect of prior access, we further construct CNO+Prior
and DeepONet+Prior, which receive deterministic RuleFeat representations
derived from the same Event Rules Library used by \method{}.
Architectural adaptations, RuleFeat construction, and training details
are provided in the \textbf{supplementary material}
(Secs.~S5.3 and S6).

\noindent\textbf{Evaluation Metrics.}
We report average rollout RMSE (Avg. RMSE), final-step RMSE
(Final RMSE), final-step Chamfer distance (CD), and the number of
trainable parameters (Params.).
Avg. RMSE is the mean of the per-step RMSE values over all
100 rollout steps, whereas Final RMSE measures the field error at the
final step.
CD is computed between the surfaces extracted from the
predicted and ground-truth final-step fields and measures final-surface
discrepancy.
Params. is reported in millions to compare model compactness.
Detailed metric definitions, channel aggregation, and surface-extraction
settings are provided in the \textbf{supplementary material}
(Sec.~S2.5).

\begin{table*}[!t]
\centering
{\footnotesize
\begin{tabular}{
@{}l
@{\hspace{4pt}}c
@{\hspace{3pt}}c
@{\hspace{3pt}}c
@{\hspace{5pt}}c
@{\hspace{3pt}}c
@{\hspace{3pt}}c
@{\hspace{5pt}}c
@{\hspace{3pt}}c
@{\hspace{3pt}}c
@{\hspace{5pt}}c
@{}
}
\toprule
\multirow[c]{2}{*}[-2pt]{Method}
& \multicolumn{3}{c}{Control-ID}
& \multicolumn{3}{c}{OOD-Joint}
& \multicolumn{3}{c}{Vis-50\%}
& \multirow[c]{2}{*}[-2pt]{
    \shortstack[c]{Params. $\downarrow$\\(M)}
} \\
\cmidrule(lr){2-4}
\cmidrule(lr){5-7}
\cmidrule(lr){8-10}
& Avg. RMSE $\downarrow$
& Final RMSE $\downarrow$
& CD $\downarrow$
& Avg. RMSE $\downarrow$
& Final RMSE $\downarrow$
& CD $\downarrow$
& Avg. RMSE $\downarrow$
& Final RMSE $\downarrow$
& CD $\downarrow$
& \\
\midrule

ConvLSTM
& 0.2730 & 0.4557 & 3.847
& 0.3657 & 0.6513 & 3.360
& 0.7117 & 0.8835 & 3.596
& \second{0.634} \\

FNO
& 0.5263 & 0.7510 & 3.838
& 0.4318 & 0.5639 & 2.865
& 0.4747 & 0.5995 & 3.398
& 1.358 \\

U-NO
& 0.4526 & 0.5403 & 3.849
& 0.5673 & 0.7861 & 3.567
& 0.4131 & 0.5069 & 3.515
& 2.432 \\

CNO
& 0.1841 & 0.3027 & 2.005
& 0.2552 & 0.4078 & 2.282
& 0.1041 & 0.2275 & 1.307
& 1.734 \\

DeepONet
& \second{0.0365}
& \second{0.0842}
& \second{0.987}
& 0.1243
& \second{0.2120}
& 1.122
& 0.1207
& 0.2004
& 1.202
& 1.720 \\

\addlinespace[2pt]

CNO+Prior
& -- & -- & --
& 0.1992 & 0.3513 & 1.555
& \second{0.0880}
& 0.2051
& 1.253
& 1.751 \\

DeepONet+Prior
& -- & -- & --
& \second{0.1165}
& 0.2161
& \second{1.088}
& 0.1300
& \second{0.1682}
& \second{1.026}
& 1.736 \\

\addlinespace[2pt]

\textbf{\method{}} (ours)
& \best{0.0134}
& \best{0.0316}
& \best{0.519}
& \best{0.0511}
& \best{0.1141}
& \best{0.851}
& \best{0.0195}
& \best{0.0659}
& \best{0.621}
& \best{0.253} \\

\bottomrule
\end{tabular}
}
\caption{
Predictive performance over 100-step autoregressive rollouts.
Results for OOD-Ar$^{+}$, OOD-Cl$^{+}$, and Vis-60\%
are provided in the \textbf{supplementary material}
(Sec.~S5.1 and Table~S11).
Computational-efficiency profiling is provided in the
\textbf{supplementary material}
(Sec.~S5.2 and Table~S12).
}
\label{tab:main_results}
\end{table*}

\subsection{Main Results}

\paragraph{Performance Comparison.}
Table~\ref{tab:main_results} reports the 100-step autoregressive rollout
performance. Our observations from the results are as follows:

\textbf{Firstly,} \method{} achieves the most accurate 100-step rollouts
under unseen in-distribution control configurations.
Compared with DeepONet, it reduces Avg. RMSE, Final RMSE, and CD by
$63.3\%$, $62.5\%$, and $47.4\%$, respectively, while using only
$0.253$M trainable parameters. These results demonstrate consistent improvements in both rollout-wide field prediction and final-state geometry.

\textbf{Secondly,} \method{} maintains its performance advantage under
both control and temporal extrapolation.
Relative to the strongest standard baseline in each setting, it reduces
Avg. RMSE by $58.9\%$ under OOD-Joint and by $81.3\%$ under Vis-50\%.
Consistent reductions in Avg. RMSE, Final RMSE, and CD demonstrate
improved generalization to held-out control regions and later evolution
stages.

\textbf{Thirdly,} feature-level access to information derived from the same Event Rules Library does not close the performance gap.
As shown in Table~\ref{tab:main_results}, \method{} reduces Avg. RMSE
relative to the strongest prior-augmented baseline by $56.1\%$ under
OOD-Joint and by $77.8\%$ under Vis-50\%.
These results show that the gains cannot be attributed to feature-level prior access alone, but arise from operationalizing the rules through the executable event-mechanism transition structure.

\begin{figure}[!h]
    \centering
    \includegraphics[width=0.9\columnwidth]
    {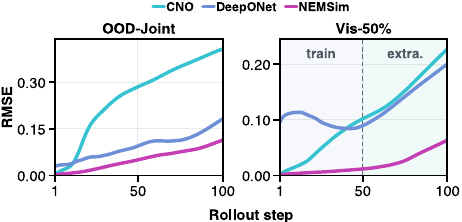}
    \caption{
    Mean stepwise RMSE over 100-step autoregressive rollouts under
    OOD-Joint and Vis-50\%.
    }
    \label{fig:rollout_curve}
\end{figure}

\begin{figure}[!h]
    \centering
    \includegraphics[width=0.9\columnwidth]{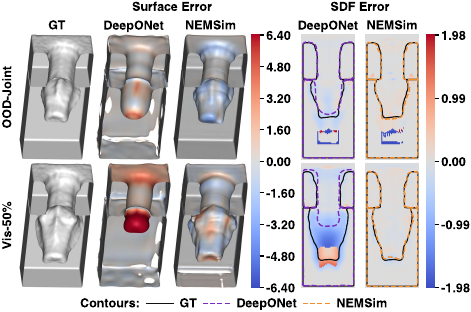}
    \caption{
    Visualization of final-step rollout predictions under OOD-Joint and Vis-50\%.
    Surface errors denote signed deviations from the ground-truth surface, and SDF errors are shown on representative slices. Additional qualitative results are provided in the \textbf{supplementary material} (Sec.~S7).
    }
    \label{fig:qualitative_results}
\end{figure}

\paragraph{Visualization.}
As shown in Fig.~\ref{fig:rollout_curve}, \method{} maintains lower
stepwise RMSE than CNO and DeepONet under both OOD-Joint and Vis-50\%.
Its advantage remains pronounced after the Vis-50\% rollout enters the
temporally extrapolative stage, indicating reduced long-horizon error
accumulation.
Fig.~\ref{fig:qualitative_results} further shows that DeepONet produces substantial surface deviations and contour misalignment near the trench bottom and sidewalls, where rollout errors accumulate most visibly. In contrast, \method{} better preserves these challenging regions, with final surfaces and SDF contours more closely aligned with the ground truth.

\begin{figure}[!h]
    \centering
    \includegraphics[width=0.9\columnwidth]
    {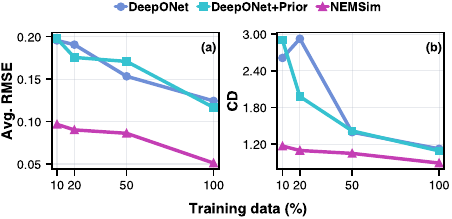}
    \caption{
    Data efficiency under OOD-Joint across fractions of
    training trajectories.
    (a) Average RMSE over 100-step autoregressive rollouts.
    (b) Final-step Chamfer distance.
    }
    \label{fig:data_efficiency}
\end{figure}

\subsection{Data Efficiency}

To assess data efficiency, we train \method{}, DeepONet, and
DeepONet+Prior using $10\%$, $20\%$, $50\%$, and $100\%$ of the
training trajectories under the same OOD-Joint protocol.
As shown in Fig.~\ref{fig:data_efficiency}, \method{} achieves the lowest
Avg. RMSE and final-step CD at every data fraction.
With only $10\%$ of the training trajectories, its Avg. RMSE remains
below $0.10$ and is lower than both DeepONet variants trained with the
complete training partition.
The sustained advantage over DeepONet+Prior shows that feature-level
prior access alone cannot explain the data-efficiency gains, supporting
the role of the executable event-mechanism transition structure.

\subsection{Mechanism Analysis}

\paragraph{Prior Ablation and Robustness.}
Table~\ref{tab:mechanism_robustness} examines whether rule semantics
provide predictive value beyond the event-mechanism architecture and how
performance changes under imperfect prior information.
Structure-only replaces all rule semantics with fixed anonymous
embeddings while retaining the same model topology and parameter
capacity.
Mechanism Shuffle randomly selects $30\%$ of the events and reassigns
their mechanism priors to different mechanism families;
Event Drop randomly selects $30\%$ of the event rules and disables them
by setting their intensities to zero; and Rate/Yield Noise independently
scales each positive rate and yield parameter by a factor sampled from
$\mathcal{U}(0.7,1.3)$.

The full model outperforms Structure-only under the same model topology
and parameter capacity, showing that rule semantics provide predictive
value beyond architectural structure alone.
All ablated and perturbed variants degrade performance relative to the
full model, confirming that the encoded rule knowledge is functionally
used.
The perturbed variants nevertheless retain nontrivial predictive
performance, indicating partial tolerance to imperfect rule information.

\begin{table}[!htbp]
\centering
{\footnotesize
\begin{tabular*}{0.96\columnwidth}{
@{\extracolsep{\fill}}
lccc
@{}
}
\specialrule{\heavyrulewidth}{0pt}{1.2pt}
Variant
& Avg. RMSE $\downarrow$
& Final RMSE $\downarrow$
& CD $\downarrow$ \\
\specialrule{\lightrulewidth}{0.8pt}{1.2pt}

\textbf{\method{}} (Full)
& \best{0.0511}
& \best{0.1141}
& \best{0.851} \\

Structure-only
& 0.0684
& 0.1525
& 0.944 \\

Mechanism Shuffle
& \second{0.0683}
& \second{0.1497}
& 0.896 \\

Event Drop
& 0.0892
& 0.2167
& \second{0.892} \\

Rate/Yield Noise
& 0.0944
& 0.1812
& 1.117 \\

\specialrule{\heavyrulewidth}{1.2pt}{0pt}
\end{tabular*}
}
\caption{
Event-rule ablation and perturbation results under OOD-Joint.
Detailed configurations are provided in the
\textbf{supplementary material} (Sec.~S5.5).
}
\label{tab:mechanism_robustness}
\end{table}

\begin{figure}[!htbp]
    \centering
    \includegraphics[width=0.95\columnwidth]
    {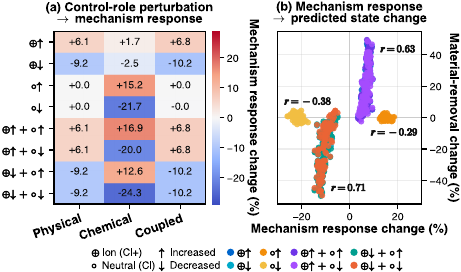}
    \caption{
    Mechanism-aligned responses under counterfactual control-role
    perturbations.
    (a) Changes in the learned physical, chemical, and coupled responses.
    (b) Relationships between selected mechanism-response changes and material-removal changes.
    }
    \label{fig:mechanism_analysis}
\end{figure}

\paragraph{Mechanism-Aligned Response Analysis.}

We fix the input state $\mathbf{x}_{t}$ and counterfactually perturb the
Cl$^{+}$ and neutral Cl flux controls relative to their nominal values.
Each flux is independently scaled to $0.5\times$ or $1.5\times$, covering
four single-control and four joint perturbations, while the unperturbed
setting $(1.0\times,1.0\times)$ serves as the reference.
We then measure the resulting changes in the learned mechanism responses
and predicted material removal.

As shown in Fig.~\ref{fig:mechanism_analysis}(a), the perturbations
produce control-role-specific and directionally consistent responses.
Increasing the Cl$^{+}$ flux primarily strengthens the physical and
coupled responses, consistent with enhanced ion bombardment and
ion-assisted removal, whereas increasing the neutral Cl flux selectively
strengthens the chemical response, consistent with enhanced
surface-chemical activity.
Decreasing either flux produces the opposite trend, while joint
perturbations combine the two response patterns. Fig.~\ref{fig:mechanism_analysis}(b) further shows that, under
Cl$^{+}$-related interventions, stronger coupled responses correspond to greater material removal, whereas neutral-Cl-only
interventions exhibit weaker immediate correlations, as expected from
their primarily chemical role. Together, these counterfactual results provide evidence that \method{} learns control-specific internal responses that are physically aligned and reflected in state changes, marking a concrete step toward interpretable neural simulation.

\subsection{Ablation Studies} 
We individually remove EIC, MAC, and MRC while keeping all other architectural and training settings unchanged. Removing any single component consistently degrades Avg. RMSE, Final RMSE, and CD across all three settings, increasing Avg. RMSE by $2.4\times$--$4.6\times$, with the largest absolute degradation under OOD-Joint. These results show that the complete EIC-MAC-MRC pathway is functionally necessary for the observed rollout performance, particularly under extrapolation.
Complete results are provided in the \textbf{supplementary material} (Sec.~S5.7).

\section{Conclusion and Limitations}

We proposed \method{}, a neural simulator that compiles predefined
event-attribute descriptions into an executable event--mechanism
transition structure for control-conditioned multi-event physical
dynamics.
By organizing event intensity, mechanism routing, and state-dependent
response composition, \method{} reduces Avg. RMSE by
$58.9\%$--$81.3\%$ across three 100-step rollout settings and remains
superior with only $10\%$ of the training data.
Prior-controlled comparisons, rule perturbations, ablations, and
mechanism-aligned analyses further support the benefit of executable
rule integration beyond prior access or architectural structure alone.

The current evaluation is limited to one 3D KMC-based plasma-etching
benchmark, reflecting the scarcity of public benchmarks for
control-conditioned multi-event physical dynamics.
\method{} also produces deterministic rollouts and assumes predefined
event-attribute descriptions. Future work will broaden the evaluation, develop probabilistic rollouts for uncertainty estimation, and support data-driven event discovery and rule refinement.

\bibliography{references}

\clearpage
\raggedbottom

\setcounter{secnumdepth}{2}
\setcounter{section}{0}
\setcounter{subsection}{0}
\setcounter{subsubsection}{0}
\setcounter{figure}{0}
\setcounter{table}{0}
\setcounter{equation}{0}

\renewcommand{\thesection}{S\arabic{section}}
\renewcommand{\thesubsection}{\thesection.\arabic{subsection}}
\renewcommand{\thesubsubsection}
{\thesubsection.\arabic{subsubsection}}

\renewcommand{\thefigure}{S\arabic{figure}}
\renewcommand{\thetable}{S\arabic{table}}
\renewcommand{\theequation}{S\arabic{equation}}

\begin{center}
    {\LARGE\bfseries Supplementary Material}\\[6pt]
\end{center}

\vspace{1em}


\section{Physical-System Background}
\label{sec:supp_physical_background}

\subsection{Plasma Etching as a Controlled Multi-Event System}
\label{sec:supp_plasma_system}

Plasma etching is a surface-processing technique in which a plasma supplies ions, neutral reactive species, and electrons to an evolving material surface.
External operating conditions regulate the particle supply, energy distribution, and incidence characteristics of these species, thereby controlling how strongly different surface processes are activated~\cite{economou2000modeling,marcos2003monte}.
In the benchmark considered in this work, these conditions are represented by process-control variables associated with supply rate, energy scale, and incidence orientation.

Surface evolution results from numerous localized events, including adsorption, desorption, charging, physical sputtering, chemical reaction, ion-assisted reaction, diffusion, and redeposition.
The occurrence and effect of each event depend on both the applied controls and the local surface state, such as material composition, site availability, and surface geometry.
Moreover, events can compete for the same local sites or interact through assistance and product-mediated dependencies.
For example, energetic ions can directly induce physical removal or assist chemical reactions, while reaction products generated upstream may later contribute to redeposition.

Although each event acts locally, repeated event occurrence,
competition, and coupling accumulate over time to produce
macroscopic 3D morphology evolution.
Plasma etching therefore provides a representative
control-conditioned multi-event system in which control changes
alter not only the magnitude of state updates, but also the
relative strengths and interactions of different event pathways.
This structure motivates a surrogate model that explicitly
organizes the learned transition through event intensity,
mechanism attribution, and state-dependent response composition.

\subsection{Particle-Based Monte Carlo Surface-Evolution Simulation}
\label{sec:supp_kmc_simulation}

The high-fidelity trajectories are generated using a self-developed
particle-based Monte Carlo feature-profile simulator, where surface
evolution emerges from repeated stochastic interactions between incident
particles and the evolving surface configuration.

Let $\mathcal{S}$ denote the microscopic surface configuration and
$\mathbf{u}$ denote the applied process controls.
At each simulation step, incident particles are sampled according to the
control-dependent distributions determined by $\mathbf{u}$, including
particle flux, energy distribution, and incident-angle distribution.
Each sampled particle is transported toward the current surface and
interacts with local surface sites according to predefined event rules.

For an incident particle interacting with local surface state
$\mathcal{S}_{\ell}$, the possible outcomes are determined by the
corresponding event attributes:
\begin{equation}
    o
    =
    \Phi(\mathcal{S}_{\ell},\mathbf{u};\mathbf{r}_{e}),
\end{equation}
where $\mathbf{r}_{e}$ contains the event-specific attributes, including
rate-related parameters, yield relations, dependency information, and
event-response characteristics.
The realized interaction may remove surface materials, generate reactive
species, modify local composition, or change neighboring-site
availability.

After each microscopic interaction, the surface configuration is updated:
\begin{equation}
    \mathcal{S}
    \leftarrow
    \Phi_{o}(\mathcal{S}).
\end{equation}

Because each local interaction modifies the evolving surface state,
subsequent particle trajectories and reaction probabilities depend on the
updated configuration, resulting in state-dependent event competition and
coupling.

A macroscopic morphology state is recorded after a predefined number of
particle interactions:
\begin{equation}
    \mathbf{x}_{t}
    =
    \Psi(\mathcal{S}(\tau_t)),
\end{equation}
where $\Psi(\cdot)$ converts the microscopic surface configuration into
the multi-channel 3D signed-distance fields used by the learning models.

Each transition
$(\mathbf{x}_{t},\mathbf{u}_{t},\mathbf{x}_{t+1})$
therefore summarizes the accumulated effects of numerous localized
particle-surface interactions rather than a single event realization.
Repeated simulation generates high-dimensional morphology trajectories
under different control conditions.

Although the simulator explicitly executes microscopic particle
interactions, the learning models are not provided with particle
trajectories, event locations, realized event identities, microscopic
states, or intermediate simulation variables.
The Event Rules Library only provides fixed event-attribute descriptions
that define possible event categories and their structural relations.

\subsection{Localized Events and Mechanism Families}
\label{sec:supp_event_mechanisms}

The simulated surface evolves through a catalog of localized events, each of which modifies a site or a small neighborhood of the current microscopic configuration.
Representative events include adsorption, desorption, charging, physical sputtering, chemical reaction, ion-assisted chemical removal, diffusion, and product redeposition.
An event is admissible only when its local prerequisites are satisfied, such as the presence of a removable material, an available adsorption site, a required reactant, or a compatible neighboring configuration.
Its propensity and local effect are further modulated by the applied process controls and the current surface state.

The event catalog describes \emph{what local transition can occur}, whereas the mechanism taxonomy describes \emph{through which underlying effect the event changes the system}.
We group the event effects into three mechanism families,
\begin{equation}
    \mathcal{M}
    =
    \{
        \mathrm{physical},
        \mathrm{chemical},
        \mathrm{coupled}
    \}.
\end{equation}
The physical family represents effects dominated by momentum or energy transfer, incidence geometry, and charged-particle interactions.
Typical examples include direct physical sputtering and control-dependent ion-impact responses.
The chemical family represents adsorption, surface reaction, desorption, and other transformations governed primarily by reactive-species availability and local chemical state.
The coupled family represents processes that require interactions between physical and chemical factors, such as ion-assisted reactions, chemically enhanced sputtering, and product-mediated dependencies.

Event types and mechanism families are not related by a one-to-one assignment.
A single event may act through more than one mechanism, while different events may share the same underlying mechanism.
For example, an ion-assisted removal event can contain both a chemical contribution associated with surface reaction and a coupled contribution associated with energetic ion assistance.
Conversely, several distinct ion-induced events may share a physical energy-transfer mechanism while differing in their rate parameters, products, or local response patterns.
This many-to-few organization separates event-specific
driving strengths from reusable mechanism-level effects.

For each event rule $\mathbf{r}_e$, the Event Rules Library therefore specifies a prior event--mechanism relation
\begin{equation}
    \boldsymbol{\alpha}^{0}_{e}
    =
    \left[
        \alpha^{0}_{e,1},
        \ldots,
        \alpha^{0}_{e,M}
    \right],
\end{equation}
where $\alpha^{0}_{e,m}$ indicates whether and to what extent event $e$ is supported by mechanism family $m$.
These priors are constructed from the predefined event attributes, including particle type, energetic or directional dependence, reaction role, product dependence, and ion-assistance flags.
They are fixed before model training and encode structural support rather than event-occurrence labels or dynamically observed mechanism targets.

The separation between localized events and shared mechanism families is important for control-conditioned morphology evolution.
External controls first alter which events are activated and how strongly they occur, while the resulting event effects are expressed through a smaller set of physical, chemical, and coupled response pathways.
Their repeated local execution and competition ultimately produce the observed macroscopic 3D state transition.

\section{Benchmark Construction}
\label{sec:supp_benchmark}

We construct a benchmark for learning
control-conditioned 3D multi-event physical state
transitions from high-fidelity reference trajectories.
The benchmark targets a setting that is not explicitly
covered by existing physical-learning resources:
high-dimensional morphology evolution produced by the
accumulation and interaction of localized discrete events
under externally varying process controls.
It is designed not merely as a collection of simulated
trajectories, but as an integrated evaluation resource
combining state-transition data, process-control metadata,
a machine-readable Event Rules Library, predefined data
splits, and standardized long-horizon rollout protocols.

The current benchmark instantiation contains 500 trajectories,
each comprising 100 transition steps on a
$54\times54\times127$ three-dimensional grid.
Each transition is represented by
\begin{equation}
    \left(
        \mathbf{x}_t,
        \mathbf{u}_t,
        \mathbf{x}_{t+1}
    \right),
\end{equation}
where $\mathbf{x}_t$ and $\mathbf{x}_{t+1}$ are consecutive
morphology states and
$\mathbf{u}_t\in\mathbb{R}^{14}$ contains the process
controls applied during the transition.
The Event Rules Library is shared across trajectories and
provides fixed system-level metadata rather than
trajectory-specific event-occurrence or mechanism labels.

The benchmark defines three complementary evaluation
regimes.
Control-ID measures interpolation to unseen control
configurations within the training control domain;
Control-OOD measures generalization to control regions
explicitly excluded from training; and Temporal
Extrapolation measures whether a one-step transition learned
from early trajectory stages can be recursively applied to
later unseen stages.
All evaluation settings use trajectory-level separation and
100-step autoregressive rollout without ground-truth state
correction.

The following subsections describe the reference simulator
and physical configuration, trajectory-generation and
control-sampling procedure, state and control
representations, dataset statistics and quality-control
checks, split and evaluation protocols, and the planned
release package.
The detailed event catalog, kinetic parameters, dependency
relations, and mechanism-family priors are documented
separately in Sec.~\ref{sec:supp_event_rules}.

\subsection{High-Fidelity Simulator and Physical Configuration}
\label{sec:supp_simulator}

\noindent\textbf{Reference Simulator and Modeling Scope.}
The reference trajectories are generated using a
three-dimensional particle-based Monte Carlo feature-profile
simulator for chlorine-based plasma etching.
The simulator resolves individual particle injection and
transport, stochastic particle--surface interactions, local
multi-species material updates, secondary-product transport,
and three-dimensional morphology evolution.
Local material changes accumulated from particle interactions
are transferred to a level-set solver, which advances the
macroscopic Si and resist interfaces.

The simulator serves as the high-fidelity reference relative
to the learned surrogate.
It should not be interpreted as experimental ground truth or
as an exact continuous-time kinetic Monte Carlo solver.
In particular, the implementation does not construct a global
event-propensity table, sample events according to normalized
reaction rates, or advance time using a Gillespie or
residence-time increment.
Instead, its basic stochastic unit is an incident particle
that is propagated through the current geometry and locally
interacts with the first accessible surface.
The reaction classes and kinetic parameterization are
consistent with established Monte Carlo feature-profile
models for chlorine-based plasma etching
\cite{huard2018nanoscale,zhang2017investigation}, while the
benchmark trajectories follow the implemented simulator
configuration described below.

\noindent\textbf{Simulator Availability.}
The reference simulator is an in-house implementation
developed for industrial research collaboration and is
subject to confidentiality restrictions.
Its complete source code and selected engineering details
therefore cannot be publicly released.
We disclose the physical configuration, particle-sampling
and transport procedure, stochastic surface-interaction
process, morphology-update scheme, snapshot cadence, and
output representation required to understand the scope and
construction of the benchmark.
The simulator is used only for offline generation of the
fixed reference trajectories and is unavailable to all
learning methods during training and evaluation.
To support reproducible machine-learning evaluation, the
benchmark release will include the generated trajectories,
process controls, preprocessing code, data-split manifests,
evaluation metrics, baseline configurations, and model
implementations.

\noindent\textbf{Physical Domain and Initial Geometry.}
The simulator maintains a multi-species voxel domain of size
\begin{equation}
    N_x\times N_y\times N_z
    =
    52\times52\times125,
\label{eq:supp_native_simulation_grid}
\end{equation}
with array axes corresponding to $(x,y,z)$.
A nominal isotropic voxel spacing of
$2~\mathrm{nm}$ is used for the geometric scale and
flux-to-particle conversion, giving a nominal physical extent
of
\begin{equation}
    104\times104\times250~\mathrm{nm}^{3}.
\end{equation}
The level-set solver itself operates in lattice coordinates
with unit numerical spacing; therefore, reported signed
distances are converted and normalized at the representation
stage rather than interpreted directly as physical distances
in nanometers.

All benchmark cases use the same deterministic initial
geometry.
The lower $80$ voxel layers are initialized as bulk Si.
A $40$ voxel resist layer is placed above the substrate,
occupying $80\leq z<120$, except for a centered cylindrical
opening of radius $10$ voxels.
The uppermost five voxel layers form an initially empty
incident region.
The nominal substrate thickness, resist thickness, and
opening diameter are therefore
$160~\mathrm{nm}$,
$80~\mathrm{nm}$, and
$40~\mathrm{nm}$, respectively.

The lateral boundaries are periodic:
a particle leaving the domain in $x$ or $y$ is wrapped to the
opposite side.
The vertical boundary is open, and particles leaving through
either end of the $z$ axis are terminated.
A voxel is considered surface-exposed when it contains
non-vacuum material and at least one of its six axis-aligned
neighbors is vacuum.
Local surface normals are estimated from the surrounding
material configuration and are used for incidence-angle
evaluation, reflection, product emission, and morphology
updates.

\begin{table*}[!t]
\centering
\footnotesize
\setlength{\tabcolsep}{5pt}
\renewcommand{\arraystretch}{1.08}
\begin{tabular}{
    p{0.17\textwidth}
    p{0.22\textwidth}
    p{0.53\textwidth}
}
\toprule
Configuration
& Implemented value
& Interpretation \\
\midrule

Native material domain
&
$52\times52\times125$
&
Multi-species voxel domain in $(x,y,z)$ ordering. \\

Nominal voxel spacing
&
$2\times2\times2~\mathrm{nm}$
&
Used for geometric scale and flux-to-particle conversion;
level-set evolution uses lattice units. \\

Initial Si region
&
$z<80$
&
Bulk silicon substrate with nominal thickness
$160~\mathrm{nm}$. \\

Initial resist region
&
$80\leq z<120$
&
Resist layer with a centered cylindrical opening. \\

Opening radius
&
$10$ voxels
&
Nominal radius $20~\mathrm{nm}$ and diameter
$40~\mathrm{nm}$. \\

Incident region
&
$120\leq z<125$
&
Initially empty region above the resist surface. \\

Particle boundaries
&
Periodic in $x$ and $y$; open in $z$
&
Lateral escape is wrapped; vertical escape terminates the
particle trajectory. \\

Surface definition
&
Six-neighbor exposure
&
A non-vacuum voxel is surface exposed when an axis-aligned
neighbor is vacuum. \\

Primary species
&
Cl, Cl$^{+}$, Cl$_2^{+}$, Ar$^{+}$
&
Species are sampled according to control-dependent incident
fluxes. \\

Native saved fields
&
$2\times54\times54\times127$
&
Separate padded mask and Si level-set fields.
The model-compatible representation is described in
Sec.~\ref{sec:supp_representation}. \\
\bottomrule
\end{tabular}
\caption{
Physical and numerical configuration of the particle-based
Monte Carlo reference simulator.
}
\label{tab:supp_simulator_configuration}
\end{table*}

\noindent\textbf{Particle Injection and Transport.}
The primary incident species are
Cl, Cl$^{+}$, Cl$_2^{+}$, and Ar$^{+}$.
For a given set of process controls, the four incident fluxes are
converted into integer particle rates using the top-surface
area and a fixed scaling factor.
Each primary particle is then sampled from the normalized
species-rate distribution.

Particles are launched from the upper boundary at
\begin{equation}
    z=N_z-\epsilon,
\end{equation}
with $x$ and $y$ sampled uniformly over the complete top
plane.
For incident species $s$, the polar direction follows a
cosine-power distribution
\begin{equation}
    p_s(\Omega)
    \propto
    \cos^{m_s}\theta,
\label{eq:supp_particle_angular_distribution}
\end{equation}
where the exponent $m_s$ is supplied by the process controls.
The azimuthal angle is sampled uniformly over
$[0,2\pi)$.
Larger values of $m_s$ produce more narrowly directed
incidence around the vertical direction.

The energy of each ion is sampled from a
species-specific Gaussian distribution,
\begin{equation}
    E_s
    \sim
    \max
    \left(
        \mathcal{N}
        \left(
            \mu_s,\sigma_s^2
        \right),
        0
    \right),
\label{eq:supp_particle_energy_sampling}
\end{equation}
where the mean and standard deviation are determined by the
current controls.
Neutral Cl particles use a fixed incident energy of
$0.2~\mathrm{eV}$.

After initialization, each particle is propagated
stepwise in continuous coordinates along its sampled
direction.
Periodic wrapping is applied in the two lateral directions,
and a particle is terminated if it escapes through the open
vertical boundary.
The implementation limits each particle trajectory to at most
$10{,}000$ propagation steps.

\noindent\textbf{Local Stochastic Surface Interactions.}
Each voxel stores amounts of multiple local species rather
than a single material label.
The represented surface states include Si, progressively
chlorinated SiCl$_x$ states, resist, and vacuum.
When a particle reaches a surface voxel, a target species is
sampled from the local voxel contents.
The simulator then evaluates the reaction channels applicable
to the projectile--target pair using the configured
probability, energy-threshold, angular, and yield models.

Neutral Cl interactions include progressive surface
chlorination and lower-probability abstraction or removal
branches.
Ion-induced interactions include direct physical sputtering,
chemically enhanced removal of chlorinated surface species,
and resist erosion.
For the Cl$^{+}$--SiCl pair, the implementation explicitly
constructs competing physical and chemical channels.
When their total probability is below one, the remaining
probability corresponds to a non-reactive outcome; otherwise,
the two channels are selected according to their normalized
relative probabilities.
The current implementation uses a single applicable reaction
channel for the other projectile--target cases.
The complete benchmark event catalog and kinetic parameters
are reported in
Sec.~\ref{sec:supp_event_rules}.

Reaction yields are converted into discrete material-removal
counts through stochastic sampling.
A successful event modifies the local species amounts and can
remove material, change its chlorination state, or generate
secondary products.
If no reaction occurs, an ion undergoes specular reflection
with an energy reduction determined by the projectile--target
mass relation.
Non-reactive neutral particles and secondary products undergo
diffuse reflection.

\noindent\textbf{Explicit Secondary-Product Transport.}
Selected ion-induced removal events generate volatile or
redepositable SiCl, SiCl$_2$, and SiCl$_3$ products.
Unlike the reduced Product Pool used inside the neural
compiler, these products are explicitly represented as
secondary particles in the reference simulator.
Each generated product is assigned an energy of
$2~\mathrm{eV}$, zero charge, and an outward direction
sampled from a cosine-weighted hemisphere around the local
surface normal.

Generated products are placed in a first-in--first-out queue
and propagated using the same particle-transport procedure as
the primary particles.
Upon reaching a supported solid surface, a product deposits
with probability
\begin{equation}
    p_{\mathrm{stick}}=0.02.
\end{equation}
A product that does not deposit undergoes diffuse reflection
and continues to be tracked.
The simulator therefore includes explicit secondary-product
transport with simplified sticking chemistry rather than a
spatially homogeneous redeposition source.

\noindent\textbf{Flux-Triggered Morphology Evolution.}
Particle interactions first update the local multi-species
voxel state.
The accumulated material changes are subsequently converted
into narrow-band velocity fields for separate Si and mask
level sets.
After a flux-derived number of incident ions has been
processed, both level sets are advanced by one macro step of
\begin{equation}
    \Delta t_{\mathrm{macro}}
    =
    0.02~\mathrm{s}.
\end{equation}
Velocity extension, local smoothing, and level-set
reinitialization are applied during this update.

The macro-time variable is calibrated from the incident ion
flux rather than sampled from microscopic reaction
propensities.
Specifically, one macro step is triggered after processing
approximately
\begin{equation}
    \left\lfloor
        R_{\mathrm{ion}}
        \Delta t_{\mathrm{macro}}
    \right\rfloor
\end{equation}
incident ions, where $R_{\mathrm{ion}}$ is the total
control-dependent ion rate.
Because neutral particles are interleaved with ions through
stochastic species sampling, the total number of primary
particles and successful local reactions within a macro step
varies across process controls and stochastic realizations.

\noindent\textbf{Saved Reference States and Task Boundary.}
The simulator is advanced from macro-time $0$ to
$10~\mathrm{s}$ and records a state every
$0.1~\mathrm{s}$.
Each trajectory therefore contains
\begin{equation}
    \left\{
        \mathbf{x}_t
    \right\}_{t=0}^{100},
\end{equation}
giving 101 saved states and 100 consecutive benchmark
transitions.
Each saved transition spans five level-set macro steps and
aggregates many incident particles, stochastic surface
interactions, material updates, and secondary-product
trajectories.
It is therefore a coarse-grained morphology transition rather
than a single microscopic event.

The simulator natively stores separate padded Si and mask
level-set fields at
$54\times54\times127$ resolution.
It does not provide microscopic event identities, event
locations, particle histories, reaction outcomes, or internal
random seeds as learning targets.
The transformation of these native fields into the
two-channel, model-compatible benchmark state representation
is described in
Sec.~\ref{sec:supp_representation}.

\subsection{Trajectory Generation and Control-Space Sampling}
\label{sec:supp_data_generation}

\noindent\textbf{Process-Control Parameterization.}
Each simulator trajectory is conditioned on a
14-dimensional process-control vector,
\begin{equation}
    \mathbf{u}^{\mathrm{raw}}
    =
    \left[
        \mathbf{u}^{\mathrm{flux}},
        \mathbf{u}^{\mathrm{angle}},
        \mathbf{u}^{\mathrm{energy}}
    \right]
    \in
    \mathbb{R}^{14},
\label{eq:supp_raw_control_vector}
\end{equation}
which controls the incident particle supply, directional
distribution, and ion-energy distribution.
The flux component is
\begin{equation}
\begin{aligned}
    \mathbf{u}^{\mathrm{flux}}
    =
    \bigl[
        &\Gamma_{\mathrm{Ar}^{+}},
        \Gamma_{\mathrm{Cl}^{+}},
        \Gamma_{\mathrm{Cl}_{2}^{+}},
        \Gamma_{\mathrm{Cl}}
    \bigr],
\end{aligned}
\label{eq:supp_flux_controls}
\end{equation}
where $\Gamma_s$ denotes the incident molar flux of species
$s$.
The directional component is
\begin{equation}
\begin{aligned}
    \mathbf{u}^{\mathrm{angle}}
    =
    \bigl[
        &m_{\mathrm{Ar}^{+}},
        m_{\mathrm{Cl}^{+}},
        m_{\mathrm{Cl}_{2}^{+}},
        m_{\mathrm{Cl}}
    \bigr],
\end{aligned}
\label{eq:supp_angle_controls}
\end{equation}
where $m_s$ is the cosine-power exponent in the incident
angular distribution
$p_s(\Omega)\propto\cos^{m_s}\theta$.
Larger values of $m_s$ correspond to more vertically
concentrated incidence.
The ion-energy component is
\begin{equation}
\begin{aligned}
    \mathbf{u}^{\mathrm{energy}}
    =
    \bigl[
        &\mu_{\mathrm{Cl}^{+}},
        \sigma_{\mathrm{Cl}^{+}},
        \mu_{\mathrm{Cl}_{2}^{+}},
        \sigma_{\mathrm{Cl}_{2}^{+}},\\
        &\mu_{\mathrm{Ar}^{+}},
        \sigma_{\mathrm{Ar}^{+}}
    \bigr],
\end{aligned}
\label{eq:supp_energy_controls}
\end{equation}
where $\mu_s$ and $\sigma_s$ define the mean and standard
deviation of the truncated Gaussian energy distribution for
ion species $s$.
Neutral Cl uses a fixed incident energy of
$0.2~\mathrm{eV}$ and therefore has no energy-control
variables.

The complete process-control vector remains fixed throughout each
trajectory.
Accordingly, each trajectory represents the morphology
evolution induced by one stationary process condition,
whereas variation across trajectories arises from
distinct control configurations and stochastic
particle--surface interactions.

\noindent\textbf{Latin Hypercube Control-Space Design.}
To obtain broad coverage of the $14$-dimensional control
space under a finite simulation budget, we generated
$N_{\mathrm{control}}=500$ process-control configurations.
Using a fixed random seed of $42$, we generated the control
design matrix
\begin{equation}
    \boldsymbol{\Xi}
    =
    \operatorname{LHS}(500,14)
    \in
    [0,1]^{500\times14},
\label{eq:supp_lhs_design}
\end{equation}
where the $i$th row
$\boldsymbol{\xi}^{(i)}\in[0,1]^{14}$
represents one sampled control configuration.
Latin hypercube sampling stratifies every control dimension
into 500 intervals and selects one sample from each interval,
providing more systematic marginal coverage than independent
unstratified sampling.

Each unit-hypercube coordinate is transformed according to
the numerical scale of its corresponding process variable:
\begin{equation}
    u^{(i)}_j
    =
    \mathcal{T}_j
    \left(
        \xi^{(i)}_j
    \right).
\label{eq:supp_control_mapping}
\end{equation}
The four incident fluxes are mapped logarithmically because
their configured ranges span substantially different
magnitudes.
Ion-energy means and standard deviations are mapped linearly,
while the cosine-power angular parameters are mapped to
bounded integer values.
Table~\ref{tab:supp_control_space} reports the complete
control-space definition.

\begin{table*}[!t]
\centering
\footnotesize
\setlength{\tabcolsep}{4pt}
\renewcommand{\arraystretch}{1.08}

\begin{tabular}{
    c
    l
    l
    l
    l
}
\toprule
Index
& Process variable
& Physical role
& Sampling transformation
& Configured range \\
\midrule

1
& $\Gamma_{\mathrm{Ar}^{+}}$
& Ar$^{+}$ incident flux
& Logarithmic LHS
& $10^{-4}$--$7\times10^{-3}~
   \mathrm{mol\,m^{-2}\,s^{-1}}$ \\

2
& $\Gamma_{\mathrm{Cl}^{+}}$
& Cl$^{+}$ incident flux
& Logarithmic LHS
& $10^{-4}$--$7\times10^{-3}~
   \mathrm{mol\,m^{-2}\,s^{-1}}$ \\

3
& $\Gamma_{\mathrm{Cl}_{2}^{+}}$
& Cl$_2^{+}$ incident flux
& Logarithmic LHS
& $10^{-6}$--$2\times10^{-4}~
   \mathrm{mol\,m^{-2}\,s^{-1}}$ \\

4
& $\Gamma_{\mathrm{Cl}}$
& Neutral Cl incident flux
& Logarithmic LHS
& $10^{-3}$--$2\times10^{-2}~
   \mathrm{mol\,m^{-2}\,s^{-1}}$ \\

5
& $m_{\mathrm{Cl}}$
& Neutral Cl angular concentration
& Bounded integer LHS
& $0$--$20$ \\

6
& $m_{\mathrm{Cl}^{+}}$
& Cl$^{+}$ angular concentration
& Bounded integer LHS
& $50$--$3000$ \\

7
& $m_{\mathrm{Cl}_{2}^{+}}$
& Cl$_2^{+}$ angular concentration
& Bounded integer LHS
& $50$--$3000$ \\

8
& $m_{\mathrm{Ar}^{+}}$
& Ar$^{+}$ angular concentration
& Bounded integer LHS
& $50$--$3000$ \\

9
& $\mu_{\mathrm{Cl}^{+}}$
& Cl$^{+}$ mean energy
& Linear LHS
& $160$--$500~\mathrm{eV}$ \\

10
& $\sigma_{\mathrm{Cl}^{+}}$
& Cl$^{+}$ energy deviation
& Linear LHS
& $5$--$10~\mathrm{eV}$ \\

11
& $\mu_{\mathrm{Cl}_{2}^{+}}$
& Cl$_2^{+}$ mean energy
& Linear LHS
& $160$--$500~\mathrm{eV}$ \\

12
& $\sigma_{\mathrm{Cl}_{2}^{+}}$
& Cl$_2^{+}$ energy deviation
& Linear LHS
& $5$--$15~\mathrm{eV}$ \\

13
& $\mu_{\mathrm{Ar}^{+}}$
& Ar$^{+}$ mean energy
& Linear LHS
& $160$--$500~\mathrm{eV}$ \\

14
& $\sigma_{\mathrm{Ar}^{+}}$
& Ar$^{+}$ energy deviation
& Linear LHS
& $5$--$15~\mathrm{eV}$ \\

\bottomrule
\end{tabular}

\caption{
Definition and sampling ranges of the 14-dimensional process
control space.
Flux variables are sampled through logarithmic mappings,
angular exponents through bounded integer mappings, and
ion-energy statistics through linear mappings.
}
\label{tab:supp_control_space}
\end{table*}

The resulting control vectors are mutually distinct and are
assigned deterministic control-configuration identifiers from
$0$ to $499$. The control-space design is generated independently of the
subsequent neural-model training, data splitting, and
evaluation results.
It therefore defines a fixed simulator design rather than a
model-dependent selection of favorable process conditions.

\noindent\textbf{Control-Conditioned Trajectory Generation.}
Each of the 500 process-control configurations is simulated
independently from the common initial geometry described in
Sec.~\ref{sec:supp_simulator}.
The process controls determine the primary-species sampling
rates, cosine-power angular distributions, and ion-energy
distributions used throughout the simulator run.
All 500 simulator cases completed successfully and produced
500 complete trajectories, with one trajectory corresponding
to each unique control configuration.

For each control configuration, the simulator evolves the
morphology from macro-time $0$ to $10~\mathrm{s}$.
Separate Si and resist level-set fields are saved at
$0.1~\mathrm{s}$ intervals, yielding
\begin{equation}
    \mathcal{X}^{(i)}
    =
    \left\{
        \mathbf{x}^{(i)}_t
    \right\}_{t=0}^{100}
\label{eq:supp_simulated_trajectory}
\end{equation}
for control configuration $i$.
Each trajectory therefore contains 101 morphology states and
forms 100 consecutive transitions:
\begin{equation}
    \left\{
        \left(
            \mathbf{x}^{(i)}_t,
            \mathbf{u}^{(i)},
            \mathbf{x}^{(i)}_{t+1}
        \right)
    \right\}_{t=0}^{99}.
\label{eq:supp_trajectory_transitions}
\end{equation}
Because the process controls remain fixed within each
trajectory, we use $\mathbf{u}^{(i)}$ rather than
$\mathbf{u}^{(i)}_t$ in
Eq.~\eqref{eq:supp_trajectory_transitions}.
The time-indexed notation $\mathbf{u}_t$ used in the general
problem formulation permits time-varying controls, whereas
the present benchmark instantiation uses stationary control
conditions.

Each saved transition spans five
$0.02~\mathrm{s}$ level-set macro steps.
The number of incident particles processed during this
interval is not fixed across control configurations.
It depends on the control-specific ion fluxes and on the
stochastic interleaving of neutral and ionic particles.
Likewise, the number and spatial distribution of successful
surface reactions vary across trajectories.
A benchmark transition should therefore be interpreted as an
aggregate morphology update produced by many microscopic
particle interactions, rather than as one particle collision
or one elementary event.

\noindent\textbf{Stochastic Realizations and Reproducibility.}
The Latin hypercube design comprising 500 process-control
configurations is deterministic and can be regenerated from
the reported sampling ranges, transformations, and random
seed.
Within each simulator run, particle species, positions,
directions, energies, transport outcomes, and local surface
reactions are sampled stochastically.
Each control configuration is associated with one
independently generated stochastic trajectory.

The internal particle-level random seeds were not retained,
and exact regeneration of identical microscopic particle
histories is therefore not claimed.
This does not affect the reproducibility of the downstream
machine-learning benchmark: the generated trajectories and
their associated raw control vectors are fixed and will be
released directly.
The transformation of the native simulator outputs and raw
process controls into the model-ready benchmark tensors is
described in Sec.~\ref{sec:supp_representation}.

\subsection{State, Control, and Transition Representation}
\label{sec:supp_representation}

\noindent\textbf{State Representation.}
The particle-based Monte Carlo simulator maintains a
multi-species voxel state internally, but the released
benchmark does not expose the microscopic voxel contents as
learning inputs.
Instead, each saved morphology state is represented by two
three-dimensional signed-distance fields corresponding to the
resist mask and Si interfaces.

The simulator natively stores each level-set field in
$(x,y,z)$ array order with a spatial size of
$54\times54\times127$.
During dataset construction, the arrays are transposed to the
neural-learning layout
\begin{equation}
    (x,y,z)
    \longrightarrow
    (z,y,x),
\label{eq:supp_axis_transposition}
\end{equation}
giving a spatial tensor size of
$127\times54\times54$ in $(D,H,W)$ order.
Let
$\phi^{\mathrm{mask}}_t$
and
$\phi^{\mathrm{Si}}_t$
denote the corresponding level-set fields at transition step
$t$.
Their zero level sets define the resist and Si interfaces,
respectively.

To bound the field magnitude while retaining the interface
geometry, each signed-distance field is scaled by
$d_{\max}=15$ voxels and clipped to $[-1,1]$:
\begin{equation}
    \bar{\phi}^{\,c}_t
    =
    \operatorname{clip}
    \left(
        \frac{\phi^{c}_t}{d_{\max}},
        -1,
        1
    \right),
    \qquad
    c\in
    \left\{
        \mathrm{mask},
        \mathrm{Si}
    \right\},
\label{eq:supp_sdf_normalization}
\end{equation}
The normalized fields are stored in single precision and
stacked in a fixed channel order:
\begin{equation}
    \mathbf{x}_t
    =
    \begin{bmatrix}
        \bar{\phi}^{\,\mathrm{mask}}_t\\
        \bar{\phi}^{\,\mathrm{Si}}_t
    \end{bmatrix}
    \in
    \mathbb{R}^{2\times127\times54\times54}.
\label{eq:supp_state_representation}
\end{equation}
Channel $0$ represents the resist-mask interface and channel
$1$ represents the Si interface.
The two channels at each step are generated at the same
simulator macro-time and are treated jointly as the system
state.

\noindent\textbf{Control Transformation and Normalization.}
Each trajectory is associated with the raw
$14$-dimensional process-control vector
$\mathbf{u}^{\mathrm{raw}}\in\mathbb{R}^{14}$
defined in Sec.~\ref{sec:supp_data_generation}.
The four incident-flux variables span multiple orders of
magnitude and are therefore transformed logarithmically,
whereas the remaining angular and ion-energy variables retain
their original numerical values.
For component $j$, the transformed control is
\begin{equation}
    \widetilde{u}_j
    =
    \begin{cases}
        \log_{10}
        \left(
            u^{\mathrm{raw}}_j+\epsilon_u
        \right),
        & j\in\mathcal{F},\\[3pt]
        u^{\mathrm{raw}}_j,
        & j\notin\mathcal{F},
    \end{cases}
\label{eq:supp_control_transformation}
\end{equation}
where
\begin{equation}
    \mathcal{F}
    =
    \left\{
        1,2,3,4
    \right\},
    \qquad
    \epsilon_u=10^{-12},
\end{equation}
corresponding to the Ar$^{+}$, Cl$^{+}$,
Cl$_2^{+}$, and neutral-Cl fluxes.

Each transformed component is subsequently mapped to
$[-1,1]$ using the predefined benchmark range:
\begin{equation}
    u_j
    =
    2
    \frac{
        \widetilde{u}_j
        -
        \widetilde{u}^{\min}_j
    }{
        \widetilde{u}^{\max}_j
        -
        \widetilde{u}^{\min}_j
    }
    -1.
\label{eq:supp_control_normalization}
\end{equation}
For the four flux variables, the transformed bounds are
\begin{equation}
\begin{gathered}
    \widetilde{u}^{\min}_j
    =
    \log_{10}
    \left(
        u^{\min}_j+\epsilon_u
    \right),\\
    \widetilde{u}^{\max}_j
    =
    \log_{10}
    \left(
        u^{\max}_j+\epsilon_u
    \right).
\end{gathered}
\label{eq:supp_flux_control_bounds}
\end{equation}
For the remaining ten variables,
$\widetilde{u}^{\min}_j=u^{\min}_j$
and
$\widetilde{u}^{\max}_j=u^{\max}_j$.
The bounds
$u^{\min}_j$ and $u^{\max}_j$
are the fixed control-space limits reported in
Table~\ref{tab:supp_control_space}.

These normalization bounds are specified before dataset
splitting and are not estimated from the training,
validation, or test trajectories.
All evaluation settings therefore use the same control
transformation without introducing split-dependent
statistics.
No clipping is applied after
Eq.~\eqref{eq:supp_control_normalization}.
All controls in the current benchmark lie within the
predefined global ranges, while the Control-OOD protocols
exclude selected subregions from training as described in
Sec.~\ref{sec:supp_protocols}.

\noindent\textbf{Transition Samples.}
For trajectory $i$, the process controls remain fixed over all
saved states.
A supervised transition sample is therefore represented as
\begin{equation}
    \left(
        \mathbf{x}^{(i)}_t,
        \mathbf{u}^{(i)},
        \mathbf{x}^{(i)}_{t+1}
    \right),
    \qquad
    t=0,\ldots,99,
\label{eq:supp_transition_sample}
\end{equation}
where
$\mathbf{u}^{(i)}\in[-1,1]^{14}$
is the normalized control vector associated with trajectory
$i$.
Adjacent states are separated by
$0.1~\mathrm{s}$ of simulator macro-time.

Because each of the 500 trajectories contains 101 states,
the complete benchmark contains
\begin{equation}
    500\times100
    =
    50{,}000
\end{equation}
one-step state transitions before data splitting.
The learning task is to approximate the transition
\begin{equation}
    \left(
        \mathbf{x}^{(i)}_t,
        \mathbf{u}^{(i)}
    \right)
    \longmapsto
    \mathbf{x}^{(i)}_{t+1}.
\label{eq:supp_benchmark_transition}
\end{equation}
All models are trained using consecutive one-step transition
pairs.
During autoregressive evaluation, the predicted state is
recursively used as the input to the subsequent transition.

The general formulation in the main paper uses
$\mathbf{u}_t$ to permit time-varying process controls.
In the present benchmark instantiation, however,
\begin{equation}
    \mathbf{u}^{(i)}_t
    =
    \mathbf{u}^{(i)},
    \qquad
    t=0,\ldots,99,
\end{equation}
so each trajectory represents a stationary control condition.

\noindent\textbf{Learning Inputs and Supervision Boundary.}
All evaluated methods receive the same state and
process-control tensors defined above.
NEMSim additionally uses the fixed Event Rules Library as
structural prior information, as described in
Sec.~\ref{sec:supp_event_rules}.
The library is shared across all trajectories and contains no
trajectory-specific event occurrences, reaction locations, or
target mechanism variables.

No learning method receives microscopic particle
trajectories, sampled surface-event identities, event counts,
reaction outcomes, secondary-particle histories, local
material-update logs, or random-number-generator states.
Training supervision is provided exclusively through the
saved next-state field
$\mathbf{x}^{(i)}_{t+1}$.
Thus, the benchmark evaluates coarse-grained
control-conditioned morphology-transition learning rather
than reconstruction of individual microscopic simulator
events.

\subsection{Dataset Statistics and Quality Control}
\label{sec:supp_dataset_statistics}

\noindent\textbf{Benchmark Summary.}
The final benchmark contains 500 complete trajectories
generated from 500 mutually distinct control configurations.
Each trajectory contains 101 synchronized morphology states
recorded at macro-times
\begin{equation}
    0,\,
    0.1,\,
    \ldots,\,
    10.0~\mathrm{s},
\end{equation}
and therefore provides 100 consecutive state transitions.
Before applying any benchmark split, the complete dataset
contains
\begin{equation}
    N_{\mathrm{state}}
    =
    500\times101
    =
    50{,}500
\label{eq:supp_total_states}
\end{equation}
saved states and
\begin{equation}
    N_{\mathrm{trans}}
    =
    500\times100
    =
    50{,}000
\label{eq:supp_total_transitions}
\end{equation}
one-step transition samples.

Each saved state is represented by two normalized
three-dimensional signed-distance fields with tensor shape
\begin{equation}
    2\times127\times54\times54,
\end{equation}
and every trajectory is associated with one
14-dimensional process-control vector.
Table~\ref{tab:supp_dataset_summary} summarizes the resulting
benchmark configuration.

\begin{table}[!t]
\centering
\footnotesize
\setlength{\tabcolsep}{4pt}
\renewcommand{\arraystretch}{1.06}
\begin{tabular}{ll}
\toprule
Property & Value \\
\midrule
Number of trajectories & 500 \\
Unique control configurations & 500 \\
Saved states per trajectory & 101 \\
Transitions per trajectory & 100 \\
Total saved states & 50,500 \\
Total one-step transitions & 50,000 \\
State channels & 2 \\
State tensor shape & $2\times127\times54\times54$ \\
Process-control dimension & 14 \\
Simulated macro-time & $0$--$10~\mathrm{s}$ \\
Snapshot interval & $0.1~\mathrm{s}$ \\
\bottomrule
\end{tabular}
\caption{
Summary statistics of the constructed benchmark before
train, validation, and test splitting.
}
\label{tab:supp_dataset_summary}
\end{table}

\noindent\textbf{Control and Trajectory Coverage.}
Each of the 500 control configurations corresponds to exactly
one independently generated trajectory.
The control-configuration identifiers span $0$ to $499$, and
no two configurations share the same $14$-dimensional
control vector.

Because the control configurations are generated using Latin
hypercube sampling, every control dimension is stratified
across its configured range.
The resulting trajectories therefore cover different
combinations of incident-particle fluxes, angular
concentration parameters, and ion-energy statistics.
This design provides systematic marginal coverage of the
14-dimensional control space under a finite simulation
budget, while not implying dense coverage of all possible
joint control combinations.

All trajectories share the same initial geometry and
material configuration.
Consequently, variations in the resulting morphology
sequences arise from differences in the process controls and
from stochastic particle injection, transport, and local
surface interactions.
The dataset therefore isolates control-conditioned evolution
without introducing trajectory-dependent changes in the
initial structure.

\noindent\textbf{Trajectory Integrity Checks.}
Each simulator case is validated before inclusion in the
benchmark.
A valid case must contain exactly 101 resist-mask frames and
101 Si frames with consecutive frame identifiers from
$0$ to $100$.
For every saved step, the paired level-set fields must have
identical timestamps and spatial dimensions.
The spatial shape must also remain unchanged throughout the
complete trajectory.

The construction pipeline verifies the following conditions:
the saved tensors contain only finite values; the signed
distances remain within a prescribed numerical sanity bound;
timestamps are nondecreasing; the two state channels are
temporally aligned; and the associated process metadata
contains all 14 control variables within their configured
global ranges.
Control identifiers and control vectors are additionally
checked for uniqueness.

All 500 simulator cases completed successfully and were
retained as complete benchmark trajectories.
The resulting dataset contains no missing transition step,
channel-time mismatch, inconsistent frame dimension, or
duplicate control configurations.
The preprocessing procedure subsequently applies the common
axis transformation, signed-distance truncation, and control
normalization described in
Sec.~\ref{sec:supp_representation}.

\noindent\textbf{Split-Independent Dataset Construction.}
All trajectory validation and preprocessing operations are
performed before the construction of the training,
validation, and test sets.
No quality-control threshold, signed-distance transformation,
or control-normalization parameter is estimated separately
from any benchmark split.
The same fixed dataset representation is therefore used
across Control-ID, Control-OOD, and Temporal Extrapolation.
The trajectory-level splits and the corresponding rollout
protocols are specified in
Sec.~\ref{sec:supp_protocols}.

\subsection{Data Splits, Rollout Protocols, and Metrics}
\label{sec:supp_protocols}

\noindent\textbf{General Split Principles.}
All benchmark partitions are constructed at the trajectory
level.
No state or transition from the same simulator trajectory is
shared across the training, validation, and test sets.
For each evaluation setting, the split manifest is fixed
before model training and is shared by NEMSim and all
baselines.
The validation trajectories are used for checkpoint
selection, whereas the test trajectories are reserved for
final evaluation.

The state and control transformations described in
Sec.~\ref{sec:supp_representation} are fixed independently of
the data split.
In particular, the control-normalization bounds are not
estimated from the training, validation, or test subsets.
All methods use the same trajectory states, process controls,
one-step targets, split manifests, and autoregressive
evaluation horizon.
NEMSim additionally uses the fixed Event Rules Library as its
intended structural prior.

Table~\ref{tab:supp_split_summary} summarizes the complete
full-500 protocols.
The number of training samples is obtained by multiplying the
number of training trajectories by the number of visible
one-step transitions per trajectory.

\begin{table*}[!t]
\centering
\scriptsize
\setlength{\tabcolsep}{3.0pt}
\renewcommand{\arraystretch}{1.08}
\begin{tabular}{
    l
    r
    r
    r
    r
    r
    r
    p{0.29\textwidth}
}
\toprule
Setting
& Train
& Val.
& Test
& Unused
& Trans./traj.
& Train samples
& Evaluation condition \\
\midrule

Control-ID
& 300
& 100
& 100
& 0
& 100
& 30,000
& Unseen control configurations within the benchmark control
domain. \\

OOD-Ar$^{+}$
& 280
& 70
& 150
& 0
& 100
& 28,000
& Ar$^{+}$ incident flux outside its central training
window. \\

OOD-Cl$^{+}$
& 279
& 70
& 151
& 0
& 100
& 27,900
& Cl$^{+}$ incident flux outside its central training
window. \\

OOD-Joint
& 202
& 50
& 53
& 195
& 100
& 20,200
& Ar$^{+}$ and Cl$^{+}$ fluxes simultaneously outside
their central training windows. \\

Vis-50\%
& 300
& 100
& 100
& 0
& 50
& 15,000
& Only the first 50 transitions are visible during one-step
training. \\

Vis-60\%
& 300
& 100
& 100
& 0
& 60
& 18,000
& Only the first 60 transitions are visible during one-step
training. \\

\bottomrule
\end{tabular}
\caption{
Trajectory-level splits for the complete 500-trajectory
benchmark.
``Trans./traj.'' denotes the number of one-step transitions
available to the training dataset.
All final evaluations use complete 100-step autoregressive
rollouts.
}
\label{tab:supp_split_summary}
\end{table*}

\noindent\textbf{Control-ID Protocol.}
Control-ID evaluates interpolation to control configurations
that are not observed during training but remain within the
benchmark control domain.
The fixed partition contains 300 training trajectories,
100 validation trajectories, and 100 test trajectories,
corresponding to a $60/20/20$ trajectory-level split.
Because every trajectory is generated from a unique
$14$-dimensional control vector, no identical control vector
is shared across the three subsets.
All 100 consecutive transitions of each training trajectory
are available for one-step supervision.

\noindent\textbf{Selection of OOD Control Axes.}
The OOD axes are selected using an analysis of the
simulator-generated morphologies, independently of all
learned-model results.
For each of the 14 process variables, we measure its
association with five final-state morphology quantities:
etch depth, top-opening width, neck width, bottom width, and
bowing.
For morphology quantity $m$ and process variable $j$, the
importance score is defined as
\begin{equation}
    I_{j,m}
    =
    0.7
    \left|
        \beta_{j,m}
    \right|
    +
    0.3
    \left|
        \rho_{j,m}
    \right|,
\label{eq:supp_control_importance}
\end{equation}
where $\beta_{j,m}$ is the standardized coefficient obtained
from multivariate linear regression and $\rho_{j,m}$ is the
Spearman correlation between the process variable and the
morphology quantity.
The overall score is obtained by averaging $I_{j,m}$ across
the five quantities.

Across all 500 trajectories, the two highest-ranked variables
are the Ar$^{+}$ incident flux and the Cl$^{+}$ incident
flux, with mean normalized importance contributions of
$34.13\%$ and $29.68\%$, respectively.
Together they account for $63.82\%$ of the mean importance
and are therefore used as the OOD-Ar$^{+}$ and
OOD-Cl$^{+}$ axes.
This selection depends only on the generated benchmark
trajectories and not on the performance of NEMSim or any
baseline.

\noindent\textbf{Control-OOD Protocol.}
The OOD regions are defined in a canonical unit coordinate.
For transformed control component $\widetilde{u}_j$, let
\begin{equation}
    q_j
    =
    \frac{
        \widetilde{u}_j
        -
        \widetilde{u}^{\min}_j
    }{
        \widetilde{u}^{\max}_j
        -
        \widetilde{u}^{\min}_j
    },
    \qquad
    q_j\in[0,1],
\label{eq:supp_ood_unit_coordinate}
\end{equation}
where the flux variables use the logarithmic transformation
defined in Eq.~\eqref{eq:supp_control_transformation}.
The central training window and the held-out boundary region
are
\begin{equation}
\begin{gathered}
    \mathcal{I}_{\mathrm{in}}
    =
    [0.15,0.85],\\
    \mathcal{I}_{\mathrm{out}}
    =
    [0,0.15)
    \cup
    (0.85,1].
\end{gathered}
\label{eq:supp_ood_intervals}
\end{equation}
Thus, the central $70\%$ of the selected control coordinate
forms the training and validation pool, while the lower and
upper boundary regions are reserved for OOD evaluation.

For OOD-Ar$^{+}$, the selected coordinate is the
Ar$^{+}$ incident flux
$\Gamma_{\mathrm{Ar}^{+}}$.
Trajectories satisfying
$q_{\mathrm{Ar}^{+}}\in\mathcal{I}_{\mathrm{in}}$
form the training and validation pool, whereas trajectories
satisfying
$q_{\mathrm{Ar}^{+}}\in\mathcal{I}_{\mathrm{out}}$
form the test set.
The resulting partition contains 280 training,
70 validation, and 150 test trajectories.
The test set contains 75 trajectories from the lower boundary
and 75 from the upper boundary.

OOD-Cl$^{+}$ is constructed analogously using the
Cl$^{+}$ incident flux
$\Gamma_{\mathrm{Cl}^{+}}$.
It contains 279 training, 70 validation, and 151 test
trajectories, with 76 test trajectories in the lower boundary
and 75 in the upper boundary.
The Cl$^{+}$ OOD coordinate refers specifically to the ionic
flux $\Gamma_{\mathrm{Cl}^{+}}$ and not to the neutral-Cl
flux $\Gamma_{\mathrm{Cl}}$.

For OOD-Joint, define the indicators
\begin{equation}
\begin{aligned}
    A
    &=
    \left[
        q_{\mathrm{Ar}^{+}}
        \in
        \mathcal{I}_{\mathrm{in}}
    \right],\\
    C
    &=
    \left[
        q_{\mathrm{Cl}^{+}}
        \in
        \mathcal{I}_{\mathrm{in}}
    \right].
\end{aligned}
\label{eq:supp_joint_ood_indicators}
\end{equation}
Trajectories satisfying $A\land C$ form the training and
validation pool.
Trajectories satisfying
$\neg A\land\neg C$, for which both selected fluxes lie
outside their training windows, form the joint-OOD test set.
Cases satisfying $A\oplus C$, for which only one selected
flux lies outside its training window, are excluded from this
setting.

The resulting OOD-Joint partition contains 202 training,
50 validation, 53 test, and 195 unused trajectories.
Let $L$ denote $q<0.15$ and $H$ denote $q>0.85$.
For the ordered pair
$(\mathrm{Ar}^{+},\mathrm{Cl}^{+})$, the joint test set
contains 14 $(L,L)$ cases, 15 $(L,H)$ cases,
13 $(H,L)$ cases, and 11 $(H,H)$ cases.
The test set therefore covers all four combinations of lower
and upper boundary extrapolation.

\noindent\textbf{Temporal Extrapolation Protocol.}
Temporal Extrapolation evaluates whether a transition learned
from early trajectory stages remains valid when recursively
applied to later stages.
The trajectory-level partition contains 300 training,
100 validation, and 100 test trajectories.

For Vis-$p\%$, the one-step training dataset exposes only the
first $T_p$ transitions:
\begin{equation}
    \mathcal{D}_{\mathrm{train}}^{(p)}
    =
    \left\{
        \left(
            \mathbf{x}^{(i)}_t,
            \mathbf{u}^{(i)},
            \mathbf{x}^{(i)}_{t+1}
        \right)
        \,\middle|\,
        t=0,\ldots,T_p-1
    \right\}.
\label{eq:supp_temporal_training_set}
\end{equation}
For Vis-50\%, $T_p=50$, so the source-state indices are
$t=0,\ldots,49$ and the target-frame indices are
$1,\ldots,50$.
For Vis-60\%, $T_p=60$, so the corresponding source and
target indices are $0,\ldots,59$ and $1,\ldots,60$.

The one-step validation and training-time prefix-test
datasets use the same visible-prefix restriction.
However, the validation rollout monitor used for checkpoint
selection reads the complete trajectory and rolls out from
$\mathbf{x}_0$ to $\mathbf{x}_{100}$.
The final test evaluation likewise covers all 100
transitions.
Consequently, the reported results include later morphology
stages that are not exposed through one-step supervision.

\noindent\textbf{Autoregressive Rollout Protocol.}
For test trajectory $i$, prediction is initialized from the
ground-truth initial state:
\begin{equation}
    \widehat{\mathbf{x}}^{(i)}_0
    =
    \mathbf{x}^{(i)}_0.
\end{equation}
The learned transition is then recursively applied as
\begin{equation}
    \widehat{\mathbf{x}}^{(i)}_{t+1}
    =
    \mathcal{G}_{\theta}
    \left(
        \widehat{\mathbf{x}}^{(i)}_t,
        \mathbf{u}^{(i)}
    \right),
    \qquad
    t=0,\ldots,99.
\label{eq:supp_autoregressive_rollout}
\end{equation}
Every predicted state is directly fed back as the input to the
next transition.
No teacher forcing, ground-truth correction, periodic reset,
SDF reinitialization, thresholding, spatial filtering,
projection, or prediction clipping is applied during rollout.
The normalized process controls are likewise evaluated
without clipping.
Surface extraction and point sampling are used only for
metric computation and do not modify the predicted states.

\noindent\textbf{Field-Error Metrics.}
Let $N_{\mathrm{test}}$ denote the number of test
trajectories, let $H=100$ denote the rollout horizon, and let
\begin{equation}
    V
    =
    2\times127\times54\times54
\end{equation}
denote the number of elements in one state tensor.
For trajectory $i$ and rollout step $t$, the field mean
squared error is
\begin{equation}
    e_{i,t}
    =
    \frac{1}{V}
    \left\|
        \widehat{\mathbf{x}}^{(i)}_t
        -
        \mathbf{x}^{(i)}_t
    \right\|_2^2.
\label{eq:supp_case_step_mse}
\end{equation}
The test-set MSE and RMSE at step $t$ are
\begin{equation}
\begin{gathered}
    \operatorname{MSE}_t
    =
    \frac{1}{N_{\mathrm{test}}}
    \sum_{i=1}^{N_{\mathrm{test}}}
    e_{i,t},\\
    \operatorname{RMSE}_t
    =
    \sqrt{
        \operatorname{MSE}_t
    }.
\end{gathered}
\label{eq:supp_step_rmse}
\end{equation}
The rollout-averaged metric is
\begin{equation}
    \operatorname{Avg.\ RMSE}
    =
    \frac{1}{H}
    \sum_{t=1}^{H}
    \operatorname{RMSE}_t,
\label{eq:supp_avg_rmse}
\end{equation}
and the final-step metric is
\begin{equation}
    \operatorname{Final\ RMSE}
    =
    \operatorname{RMSE}_{H}.
\label{eq:supp_final_rmse}
\end{equation}
Avg.~RMSE therefore averages the 100 independently computed
step-wise RMSE values.
It is not obtained by averaging the MSE values across time
before applying a single square root.
Both metrics are computed directly in the normalized
two-channel SDF representation.

\noindent\textbf{Final-Surface Chamfer Distance.}
Final-state geometric fidelity is evaluated by extracting the
zero-level surface from each SDF channel separately.
The mask channel is evaluated over depth indices
$80$ through $121$, and the Si channel over indices
$0$ through $80$, with both endpoints included.
Marching cubes is applied at level zero using unit voxel
spacing.

For each predicted and ground-truth surface, 20,000 points
are sampled from the triangular mesh with probability
proportional to triangle area.
The sampling seed is determined by the trajectory index,
rollout step, and state channel.
If a cropped SDF contains no zero crossing, the voxel centers
with the smallest absolute SDF values are used as a
deterministic fallback.

For two point sets $\mathcal{P}$ and $\mathcal{Q}$, the
symmetric channel-wise Chamfer distance is
\begin{equation}
\begin{aligned}
    \operatorname{CD}
    \left(
        \mathcal{P},
        \mathcal{Q}
    \right)
    =
    \frac{1}{2}
    \Bigg[
    &
    \frac{1}{|\mathcal{P}|}
    \sum_{\mathbf{p}\in\mathcal{P}}
    \min_{\mathbf{q}\in\mathcal{Q}}
    \left\|
        \mathbf{p}-\mathbf{q}
    \right\|_2\\
    +
    &
    \frac{1}{|\mathcal{Q}|}
    \sum_{\mathbf{q}\in\mathcal{Q}}
    \min_{\mathbf{p}\in\mathcal{P}}
    \left\|
        \mathbf{q}-\mathbf{p}
    \right\|_2
    \Bigg].
\end{aligned}
\label{eq:supp_chamfer_distance}
\end{equation}
The final CD of one trajectory is the mean of the mask and Si
channel distances.
The reported value is then averaged across all test
trajectories.
The nearest-neighbor distances are Euclidean and are not
squared.
Because surface extraction uses spacing $(1,1,1)$, CD is
reported in voxel units.

\noindent\textbf{Parameters.}
We report the number of trainable scalar parameters in millions:
\begin{equation}
    \operatorname{Params.}
    =
    \frac{1}{10^{6}}
    \sum_{p\in\mathcal{P}_{\mathrm{train}}}
    \operatorname{numel}(p),
\label{eq:supp_params}
\end{equation}
where $\mathcal{P}_{\mathrm{train}}$ contains the parameter tensors
optimized during training.
Shared parameter objects are counted only once, while fixed rule tensors
and other non-trainable buffers are excluded.

\subsection{Release Package and Reproducibility}
\label{sec:supp_release}

\noindent\textbf{Release Scope.}
The complete 500-trajectory dataset is not included in the
arXiv source package because of its size.
We plan to publicly release the complete benchmark package
described below.
The complete release is designed to reproduce the
machine-learning task, including data loading, preprocessing,
model training, checkpoint selection, and final 100-step
rollout evaluation.
It will contain the complete set of 500 generated trajectories,
their associated process controls, the machine-readable Event
Rules Library, fixed data-split manifests, preprocessing
utilities, metric implementations, baseline configurations,
and the NEMSim implementation.

The released trajectories are fixed reference data.
Consequently, reproducing the benchmark results does not
require access to the particle-based Monte Carlo simulator
during either training or evaluation.
All evaluated methods operate only on the released state
fields, process controls, and, for rule-aware models, the
fixed Event Rules Library.

\noindent\textbf{Benchmark Data Package.}
Each released case is assigned a unique identifier that is
mapped one-to-one to its process-control identifier.
The trajectory package stores the model-ready state sequence,
the corresponding raw and transformed process controls, the
saved macro-time coordinates, and the metadata required to
interpret the tensor representation.
The principal released fields are summarized in
Table~\ref{tab:supp_release_schema}.

\begin{table*}[!t]
\centering
\footnotesize
\setlength{\tabcolsep}{5pt}
\renewcommand{\arraystretch}{1.08}
\begin{tabular}{
    p{0.18\textwidth}
    p{0.24\textwidth}
    p{0.50\textwidth}
}
\toprule
Released component
& Representation
& Description \\
\midrule

State trajectories
&
$\boldsymbol{\Phi}
\in
\mathbb{R}^{101\times2\times127\times54\times54}$
&
Normalized mask and Si signed-distance fields for one
complete trajectory, ordered as
$(T,C,D,H,W)$. \\

Raw controls
&
$\mathbf{u}^{\mathrm{raw}}
\in
\mathbb{R}^{14}$
&
Original process-control values in their physical or numerical
units. \\

Transformed controls
&
$\widetilde{\mathbf{u}}
\in
\mathbb{R}^{14}$
&
Control vector after logarithmic transformation of the four
flux dimensions and before range normalization. \\

Time coordinates
&
$\mathbf{t}
\in
\mathbb{R}^{101}$
&
Simulator macro-times associated with the saved states. \\

Case metadata
&
Case and control identifiers
&
Identifiers, parameter order, tensor dimensions, channel
semantics, and representation metadata. \\

Event Rules Library
&
Machine-readable rule records
&
Fixed event attributes, kinetic parameters, dependency
relations, and mechanism-family priors shared across all
trajectories. \\

Split manifests
&
Case-identifier lists
&
Fixed training, validation, test, and unused sets for
Control-ID, OOD-Ar$^{+}$, OOD-Cl$^{+}$, OOD-Joint,
Vis-50\%, and Vis-60\%. \\

\bottomrule
\end{tabular}
\caption{
Principal components of the released benchmark package.
}
\label{tab:supp_release_schema}
\end{table*}

The package will also include the global control ranges,
control ordering, channel ordering, signed-distance
truncation constant, and all transformations defined in
Sec.~\ref{sec:supp_representation}.
Both raw and transformed controls are retained so that users
can inspect the original process conditions and independently
reconstruct the model-input normalization.

\noindent\textbf{Fixed Split Manifests.}
The exact trajectory-level partitions described in
Sec.~\ref{sec:supp_protocols} will be distributed as explicit
case manifests rather than regenerated implicitly at runtime.
Separate manifests will be provided for Control-ID,
OOD-Ar$^{+}$, OOD-Cl$^{+}$, OOD-Joint, and Temporal
Extrapolation.
For OOD-Joint, the manifest will also identify trajectories
excluded because only one of the two selected fluxes lies
outside the central training window.

Using fixed manifests prevents implementation-dependent
differences in case assignment and ensures that all models
are trained and evaluated on identical trajectory sets.
The temporal protocols additionally specify the visible
transition indices for Vis-50\% and Vis-60\%, while retaining
the complete trajectories for autoregressive rollout
evaluation.

\noindent\textbf{Preprocessing and Evaluation Code.}
The release will contain the code required to reproduce the
benchmark representation from the stored trajectory fields.
This includes axis ordering, signed-distance truncation and
scaling, control transformation, range normalization, and
construction of consecutive one-step transition pairs.

The evaluation package will implement the exact procedures
reported in Sec.~\ref{sec:supp_protocols}, including
100-step autoregressive rollout, step-wise and final RMSE,
zero-level surface extraction, area-weighted surface-point
sampling, symmetric Chamfer Distance, and trainable-parameter
counting.
The released evaluators will use the same settings for all
methods and will not apply prediction clipping, teacher
forcing, state correction, or architecture-specific surface
postprocessing during rollout.

\noindent\textbf{Model and Baseline Reproducibility.}
The release will include the NEMSim architecture, Event Rules
Library parser, training configuration, and evaluation
configuration used for the reported experiments.
Configuration files for ConvLSTM, FNO, U-NO, CNO, and
DeepONet will be included together with the corresponding
training and rollout interfaces.
The knowledge-augmented RuleFeat variants will also be
provided so that comparisons under matched access to the
Event Rules Library can be reproduced.

For every reported setting, the package will specify the
model configuration, optimization parameters, checkpoint
selection criterion, data manifest, visible temporal prefix
when applicable, and final evaluation command.
The same released preprocessing and metric implementations
will be used across NEMSim and all baselines.

\noindent\textbf{Simulator Disclosure Boundary.}
The particle-based Monte Carlo reference simulator is an
in-house implementation developed for industrial research
collaboration and is subject to confidentiality restrictions.
Its complete source code and selected engineering details
therefore cannot be released.
The release does not claim bitwise reproducibility of the
underlying particle-level simulator or exact regeneration of
the microscopic stochastic histories.

Instead, reproducibility is defined at the benchmark level.
The complete generated trajectories, process controls,
representation code, rule metadata, split manifests,
training configurations, and evaluation procedures are fixed
and are planned for inclusion in the public benchmark release.
The proprietary simulator is not invoked by NEMSim, any
baseline, or the evaluation pipeline.
Thus, all reported machine-learning comparisons can be
reproduced from the released package without access to the
trajectory-generation software.

\noindent\textbf{Versioning and Integrity.}
The public release will assign a version identifier to the
benchmark and preserve the full-500 trajectory set and split
manifests used in this paper.
Checksums will be provided for the principal data archives
and manifest files.
A dataset card will document the benchmark scope, state and
control semantics, physical and numerical configuration,
known limitations, simulator disclosure boundary, and
recommended citation.

Future extensions, such as additional control configurations,
geometries, or physical systems, will be released under new
version identifiers rather than modifying the benchmark splits
used for the results reported here.

\section{Event Rules Library}
\label{sec:supp_event_rules}

\subsection{Provenance and Role of the Event Rules Library}
\label{sec:supp_rule_provenance}

The Event Rules Library is a system-specific, machine-readable
catalog of the localized transitions considered possible in the
benchmark physical system.
It is constructed before model training by organizing the
event logic and parameter roles used by the benchmark simulator
into a unified event schema.
The current implementation represents the library as
\begin{equation}
    \mathcal{R}
    =
    \left\{
        \mathbf{r}_e
    \right\}_{e=1}^{N_e},
\end{equation}
where each rule $\mathbf{r}_e$ describes one event channel,
including its reactant or projectile, target surface species,
control bindings, probability and yield relations,
product-mediated dependencies, and mechanism role.

\noindent\textbf{Physical and Simulator Provenance.}
The event taxonomy is based on the chlorine-silicon surface
processes represented in the benchmark simulator, including
neutral chlorination and abstraction, ion-induced physical
sputtering, chemically enhanced removal of chlorinated
surface species, resist erosion, volatile product generation,
and product redeposition.
These event classes and their kinetic interpretation are
consistent with established Monte Carlo feature-profile
reaction mechanisms for chlorine-based plasma etching
\cite{huard2018nanoscale,zhang2017investigation}.
In particular, the source mechanisms distinguish
energy-independent neutral reactions from ion-induced
processes governed by threshold energy, reference
probability, and incidence-angle dependence.

The published reaction mechanisms provide the physical basis
for the event categories, parameter roles, and kinetic
functional forms, rather than being copied verbatim.
The numerical values and event decomposition used by
\method{} follow the implemented benchmark configuration,
which adapts the source mechanisms to the species set,
control representation, product handling, and state variables
used in this work.
Accordingly, the machine-readable configuration and its
compiler define the event rules used in all experiments.

\noindent\textbf{Separation of Event Structure and Process Controls.}
The library separates intrinsic event structure from the
process controls under which an event occurs.
An event rule specifies the relevant control roles, such as
particle supply, incident energy, and incidence orientation,
together with the corresponding probability and yield
relations.
The numerical values of these controls are supplied by
$\mathbf{u}_t$ at each transition.
Thus, changing the process controls changes the compiled
event intensities without changing the identity or schema of
the available events.
This separation follows the feature-profile modeling view
that the surface-reaction mechanism is fixed for a given
reactant-surface system, whereas the process controls
determine the magnitudes, energies, and angular
distributions of the incident species
\cite{zhang2017investigation}.

\noindent\textbf{Current Benchmark Instantiation.}
The implemented library contains $N_e=37$ event channels.
Among them, 22 are primary events directly driven by
externally supplied neutral or ion controls, and 15 are
product-mediated deposition events driven by the internal
product pool.
The primary catalog includes six neutral-Cl surface events
and sixteen ion-induced channels associated with
Cl$^{+}$, Cl$_2^{+}$, and Ar$^{+}$.
The product-mediated catalog represents the redeposition of
SiCl, SiCl$_2$, and SiCl$_3$ onto the five modeled target
classes.
Events are assigned semantic groups denoted as
\texttt{chemical}, \texttt{physical}, or \texttt{both};
the last group denotes event channels whose effects require
joint physical and chemical interpretation.
The complete event catalog and its parameterization are
reported in Sec.~\ref{sec:supp_event_catalog}.

\noindent\textbf{Compilation into Fixed Structural Inputs.}
Before training, the event configuration is parsed and
compiled into fixed tensor representations:
\begin{equation}
\begin{aligned}
    \operatorname{Compile}(\mathcal{R})
    =
    \bigl\{&
        \mathbf{R},
        \boldsymbol{\Phi}^{\mathrm{evt}},
        \boldsymbol{\alpha}^{0},
        \mathbf{R}^{\mathrm{mech}},\\
        &
        \mathbf{W}^{\mathrm{src}},
        \mathbf{W}^{\mathrm{sink}},
        \mathbf{s}^{\mathrm{dep}}
    \bigr\},
\end{aligned}
\label{eq:supp_event_library_compilation}
\end{equation}
where $\mathbf{R}\in\mathbb{R}^{N_e\times22}$ is the compiled
event-parameter matrix,
$\boldsymbol{\Phi}^{\mathrm{evt}}$ is the event-descriptor
matrix,
$\boldsymbol{\alpha}^{0}\in\mathbb{R}^{N_e\times M}$ contains
the static event-mechanism priors, and
$\mathbf{R}^{\mathrm{mech}}\in\{0,1\}^{N_e\times M}$
contains the deterministic mechanism-role signals.
The source and sink tensors
$\mathbf{W}^{\mathrm{src}}$ and
$\mathbf{W}^{\mathrm{sink}}$ encode product-generation and
product-consumption relations, while
$\mathbf{s}^{\mathrm{dep}}$ contains the sticking
coefficients of product-mediated events.

These tensors are registered as fixed model buffers rather
than trainable parameters.
They remain unchanged across trajectories, training epochs,
data splits, and rollout steps.
Consequently, the model cannot modify the event catalog,
kinetic constants, control bindings, product graph,
or static mechanism priors to fit a particular trajectory.

\noindent\textbf{Role in the Executable Transition.}
The compiled library is used at three stages of \method{}.
First, EIC reads the control bindings and kinetic attributes
in $\mathbf{R}$ to gather event-specific controls, evaluate
rule-defined base intensities, and propagate generated products
through the Product Pool.
Second, MAC uses the event descriptors
$\boldsymbol{\Phi}^{\mathrm{evt}}$, the static prior
$\boldsymbol{\alpha}^{0}$, the deterministic role signals
$\mathbf{R}^{\mathrm{mech}}$, and the final event intensities
to construct prior-guided, context-dependent mechanism
attributions.
Product-mediated effects enter MAC through the
dependency-driven event intensities compiled by EIC, while
the fixed source and sink attributes contribute to the
mechanism-role signals.
Third, MRC uses each event descriptor
$\boldsymbol{\phi}_e$ to construct the Event-to-Modes
Signature, through which event-mechanism drives are aggregated
into mechanism-mode drives.
The library therefore affects the executed forward transition
rather than serving only as auxiliary metadata or an
additional input feature.

\noindent\textbf{No Intermediate Supervision.}
The Event Rules Library specifies which event types are
possible and how their attributes enter the model, but it
does not reveal which microscopic events occurred during an
observed transition.
In particular, the training data do not provide event
identities, event locations, event counts, microscopic
propensities, Product Pool values, mechanism labels,
mechanism attributions, or response-mode targets.
Only the state transition
$(\mathbf{x}_t,\mathbf{u}_t,\mathbf{x}_{t+1})$
is supervised.
Event intensities, mechanism attributions, and mechanism-mode
responses are computed as latent intermediate quantities, and
their learnable components are trained only through the final
state-transition objective.

The library should therefore be interpreted as a fixed
structural interface between available system knowledge and
the learned transition model.
It constrains how controls, events, mechanisms, and responses
are composed, while the state-control-dependent corrections
and latent response fields are learned from trajectory data.

\subsection{Rule Schema and Executable Attributes}
\label{sec:supp_rule_schema}

Each entry follows the abstract event-rule representation introduced
in the main paper:
\begin{equation}
    \mathbf{r}_e
    =
    \left(
        \mathcal{C}_e,\,
        p_e,\,
        Y_e,\,
        \mathcal{D}_e,\,
        g_e
    \right).
\label{eq:supp_rule_definition}
\end{equation}
For implementation, the corresponding attributes and metadata are
stored in the validated specification
\begin{equation}
    \widetilde{\mathbf{r}}_e
    =
    \left(
        \mathbf{r}^{\mathrm{id}}_e,\,
        \mathbf{r}^{\mathrm{ctrl}}_e,\,
        \mathbf{r}^{\mathrm{kin}}_e,\,
        \mathbf{r}^{\mathrm{dep}}_e,\,
        \mathbf{r}^{\mathrm{prod}}_e
    \right),
\label{eq:supp_rule_decomposition}
\end{equation}
where the five components respectively store event identity,
control bindings, kinetic attributes, dependency attributes, and
generated products.
The compiler maps this implementation specification to
$\mathcal{C}_e$, $p_e$, $Y_e$, $\mathcal{D}_e$, and $g_e$.
The schema distinguishes primary events, whose rule-defined base
intensities are computed from process controls, from product-mediated
deposition events, whose dependency-driven intensities are computed
from intermediate species generated by upstream events.

\noindent\textbf{Event Identity and Semantics.}
Every event rule contains four required identity fields:
\texttt{name}, \texttt{group}, \texttt{projectile}, and
\texttt{target}.
The event name provides a unique identifier, while the
projectile and target specify the incoming species and the
surface species on which the event acts.
The current projectile vocabulary is
\begin{equation}
\begin{aligned}
    \mathcal{P}
    =
    \{&
    \mathrm{Cl},
    \mathrm{Cl}_2,
    \mathrm{Cl}^{+},
    \mathrm{Cl}_2^{+},
    \mathrm{Ar}^{+},\\
    &\mathrm{SiCl},
    \mathrm{SiCl}_2,
    \mathrm{SiCl}_3
    \},
\end{aligned}
\label{eq:supp_projectile_vocab}
\end{equation}
and the target vocabulary is
\begin{equation}
    \mathcal{T}
    =
    \{
        \mathrm{Si},
        \mathrm{SiCl},
        \mathrm{SiCl}_2,
        \mathrm{SiCl}_3,
        \mathrm{Resist}
    \}.
\label{eq:supp_target_vocab}
\end{equation}

The \texttt{group} field takes one of
\texttt{chemical}, \texttt{physical}, or \texttt{both}.
It provides the semantic group used to construct the static
mechanism prior.
The optional \texttt{effect} field records whether the event
primarily represents reaction, etching, sputtering, or
deposition.
Although this field does not directly define the
event-intensity relation, it contributes to the compiled
semantic descriptor used by MAC and MRC.

\noindent\textbf{Primary-Event Control Bindings.}
A primary event must specify a \texttt{flux\_key}, which
identifies the supply component of the process-control vector
used to drive the event.
Energetic events may additionally specify
\texttt{angle\_key}, \texttt{E\_mean\_key}, and
\texttt{E\_std\_key}.
These string-valued keys are resolved at compilation time
into fixed indices of the 14-dimensional control vector.

For event $e$, the compiled control bindings are
\begin{equation}
    \boldsymbol{\iota}_e
    =
    \left(
        \iota^{f}_e,
        \iota^{\theta}_e,
        \iota^{E}_e,
        \iota^{\sigma_E}_e
    \right),
\label{eq:supp_control_bindings}
\end{equation}
where the four entries identify the supply,
incidence-orientation, mean-energy, and energy-spread
channels, respectively.
An unavailable control role is represented by the sentinel
index $-1$ and is ignored by the corresponding kinetic
formula.
The index vector $\boldsymbol{\iota}_e$ is the compiled
implementation of the control-binding attribute
$\mathcal{C}_e$ used in the main paper.
These bindings are fixed structural relations rather than
learned event-control associations.

\noindent\textbf{Kinetic Attributes.}
Each primary event specifies both a probability model and a
yield model.
The probability model is selected by
\texttt{prob\_model} from
\begin{equation}
    \mathcal{Q}
    =
    \{
        \mathrm{const},
        \mathrm{probModel}
    \}.
\end{equation}
A constant-probability event stores a scalar probability
$p_e^{\mathrm{const}}$, whereas an energy-dependent event stores
\begin{equation}
    \boldsymbol{\eta}^{p}_e
    =
    \left(
        E^{\mathrm{th}}_e,\,
        E^{\mathrm{ref}}_e,\,
        p^0_e,\,
        n_e,\,
        a_e
    \right),
\label{eq:supp_probability_attributes}
\end{equation}
where $E^{\mathrm{th}}_e$ is the activation threshold,
$E^{\mathrm{ref}}_e$ is the reference energy,
$p^0_e$ is the reference probability,
$n_e$ is the energy exponent, and
$a_e\in\{\mathrm{None},\mathrm{P},\mathrm{C}\}$
selects no angular correction, a physical-sputtering angular
law, or a chemical-sputtering angular law.

The yield model is selected by \texttt{yield\_model} from
\begin{equation}
    \mathcal{Y}
    =
    \{
        \mathrm{const},
        \mathrm{Yamamura}
    \}.
\end{equation}
A constant-yield event stores $Y^{\mathrm{const}}_e$.
An angular sputtering-yield event stores
\begin{equation}
    \boldsymbol{\eta}^{Y}_e
    =
    \left(
        E^{Y,\mathrm{th}}_e,\,
        \theta^{\max}_e,\,
        Y^{\max}_e,\,
        Y^0_e
    \right),
\label{eq:supp_yield_attributes}
\end{equation}
which respectively denote the yield threshold, the incidence
angle of maximum yield, the maximum yield, and the
normal-incidence yield.
These parameters are evaluated by the differentiable kinetic
forms described in
Sec.~\ref{sec:supp_kinetic_models} and
Sec.~\ref{sec:supp_eic}.

\noindent\textbf{Product and Dependency Attributes.}
A primary event may include a variable-length
\texttt{products} list.
Each product entry is represented by
\begin{equation}
    \boldsymbol{\chi}^{\mathrm{prod}}_{e,j}
    =
    \left(
        s_{e,j},
        \xi_{e,j}
    \right),
\label{eq:supp_product_entry}
\end{equation}
where $s_{e,j}$ is the generated species and
$\xi_{e,j}$ is either a fixed numerical coefficient or the
special value \texttt{yield}.
The latter indicates that the amount supplied to the Product
Pool is proportional to the event-specific sputtering yield. 

A product-mediated deposition event does not require an
external flux, probability model, or yield model.
Instead, it specifies
\texttt{dep\_model=from\_products\_pool} and
\begin{equation}
    \boldsymbol{\eta}^{\mathrm{dep}}_e
    =
    \left(
        s^{\mathrm{dep}}_e,\,
        p^{\mathrm{stick}}_e
    \right),
\label{eq:supp_deposition_attributes}
\end{equation}
where $s^{\mathrm{dep}}_e$ identifies the required intermediate
species and $p^{\mathrm{stick}}_e$ is its sticking coefficient
on the target surface.
This distinction explicitly represents the directed
source-product-sink relations among events. Together, the generated-product entries and deposition attributes
implement the product-mediated dependency attribute
$\mathcal{D}_e$ used in the main paper.

\noindent\textbf{Compilation into the Event-Parameter Matrix.}
The human-readable rules are validated and compiled into a
fixed event-parameter matrix
\begin{equation}
    \mathbf{R}
    =
    \left[
        \mathbf{R}_1;
        \ldots;
        \mathbf{R}_{N_e}
    \right]
    \in
    \mathbb{R}^{N_e\times 22}.
\label{eq:supp_event_matrix}
\end{equation}
Each row is organized as
\begin{equation}
    \mathbf{R}_e
    =
    \left[
        \mathbf{R}^{\mathrm{id}}_e,
        \mathbf{R}^{\mathrm{ctrl}}_e,
        \mathbf{R}^{\mathrm{prob}}_e,
        \mathbf{R}^{\mathrm{yield}}_e,
        \mathbf{R}^{\mathrm{dep}}_e
    \right],
\label{eq:supp_event_matrix_blocks}
\end{equation}
with block dimensions
\begin{equation}
    4+4+6+6+2=22.
\end{equation}
The identity block contains encoded group, projectile, target,
and deposition-model identifiers.
The string-valued event name is retained as metadata and is
not stored in $\mathbf{R}$.
The control block contains the four control indices in
Eq.~\eqref{eq:supp_control_bindings}.
The probability and yield blocks contain the model
identifiers and kinetic parameters, while the dependency
block contains the required product-species identifier and
sticking coefficient.

\begin{table*}[!t]
\centering
\footnotesize
\setlength{\tabcolsep}{4pt}
\renewcommand{\arraystretch}{1.18}

\begin{tabular}{
p{0.15\textwidth}
p{0.29\textwidth}
p{0.48\textwidth}
}
\toprule
Attribute group
& Source fields
& Executable role \\
\midrule

Event identity
&
\texttt{name}, \texttt{group},
\texttt{projectile}, \texttt{target},
\texttt{effect}
&
Defines event identity, species semantics, target type,
semantic mechanism support, and response-role flags. \\

Control binding
&
\texttt{flux\_key}, \texttt{angle\_key},
\texttt{E\_mean\_key}, \texttt{E\_std\_key}
&
Selects the event-specific supply, incidence, and energy
channels gathered from $\mathbf{u}_t$ by EIC. \\

Probability
&
\texttt{prob\_model},
\texttt{prob\_params}
&
Selects a constant or energy-angle-dependent interaction
probability and provides
$p^{\mathrm{const}}$ or
$E^{\mathrm{th}}$, $E^{\mathrm{ref}}$, $p^0$, $n$, and
the angular-law type. \\

Yield
&
\texttt{yield\_model},
\texttt{yield\_params}
&
Selects a constant or Yamamura-style yield and provides
$Y^{\mathrm{const}}$ or the yield threshold,
maximum-yield angle, maximum yield, and normal-incidence
yield. \\

Product source
&
\texttt{products.species},
\texttt{products.coeff}
&
Defines which intermediate species are generated and whether
their amount is fixed or proportional to the event yield. \\

Product sink
&
\texttt{dep\_model},
\texttt{dep\_params.species},
\texttt{dep\_params.p\_stick}
&
Defines product-mediated deposition events and their required
species and sticking coefficients. \\

\bottomrule
\end{tabular}

\caption{
Event-rule attributes and their executable roles in
\method{}.
}
\label{tab:supp_rule_schema}
\end{table*}

\noindent\textbf{Compiled Event Descriptor.}
The event-parameter matrix is used directly by EIC, but MAC
and MRC require a compact semantic representation that does
not contain raw control indices.
For this purpose, the compiler constructs an 18-dimensional
normalized parameter vector
\begin{equation}
    \bar{\mathbf{r}}_e
    \in
    \mathbb{R}^{18},
\label{eq:supp_normalized_event_params}
\end{equation}
containing normalized identity identifiers, probability
parameters, yield parameters, and dependency parameters.

The compiler additionally derives a 17-dimensional binary
meta-feature vector
\begin{equation}
    \mathbf{m}_e
    \in
    \{0,1\}^{17}.
\label{eq:supp_event_meta_flags}
\end{equation}
These flags indicate whether an event is product-mediated,
ion-driven, or neutral-driven; whether it uses physical or
chemical angular dependence; whether energy and Yamamura
yield models are present; whether the event represents
reaction, etching, or deposition; whether it generates or
consumes products; and whether its target is resist or a
chlorinated silicon species.

The complete event descriptor is
\begin{equation}
    \boldsymbol{\phi}_e
    =
    \left[
        \bar{\mathbf{r}}_e,
        \mathbf{m}_e
    \right]
    \in
    \mathbb{R}^{35}.
\label{eq:supp_event_descriptor}
\end{equation}
Stacking all descriptors gives
\begin{equation}
    \boldsymbol{\Phi}^{\mathrm{evt}}
    =
    \left[
        \boldsymbol{\phi}_1;
        \ldots;
        \boldsymbol{\phi}_{N_e}
    \right]
    \in
    \mathbb{R}^{N_e\times 35}.
\label{eq:supp_event_descriptor_matrix}
\end{equation}
Each event descriptor $\boldsymbol{\phi}_e$ is used by MAC
and by the Event-to-Modes Signature in MRC.
EIC instead uses the compact rate descriptor
$\boldsymbol{\psi}_e$ defined in
Sec.~\ref{sec:supp_eic}.

\noindent\textbf{Compiled Product Graph.}
The variable-length product lists are also converted into
fixed tensors.
The product-source matrix is
\begin{equation}
    \mathbf{W}^{\mathrm{src}}
    \in
    \mathbb{R}_{\geq 0}^{N_e\times N_s},
\label{eq:supp_source_matrix}
\end{equation}
where $N_s=8$ is the current projectile and product
vocabulary size. The dependency attribute $\mathcal{D}_e$ is compiled into the
corresponding rows of $\mathbf{W}^{\mathrm{src}}$,
$\mathbf{W}^{\mathrm{sink}}$, and $\mathbf{s}^{\mathrm{dep}}$.
These tensors jointly encode product generation, product
requirements, and sticking coefficients.
Its entry $W^{\mathrm{src}}_{e,q}$ indicates that event $e$
can generate species $q$.
The product-sink matrix is
\begin{equation}
    \mathbf{W}^{\mathrm{sink}}
    \in
    \{0,1\}^{N_e\times N_s},
\label{eq:supp_sink_matrix}
\end{equation}
where $W^{\mathrm{sink}}_{e,q}=1$ identifies the intermediate
species required by deposition event $e$.
The associated sticking coefficients are stored in
\begin{equation}
    \mathbf{s}^{\mathrm{dep}}
    =
    \left[
        p^{\mathrm{stick}}_1,
        \ldots,
        p^{\mathrm{stick}}_{N_e}
    \right].
\label{eq:supp_stickiness_vector}
\end{equation}

The source and sink tensors encode structural connectivity.
During execution, EIC retains the original coefficient type
for each source relation and computes the actual product
amount using either its fixed coefficient or the current
event yield.
Thus, the compiled graph specifies which dependencies are
allowed, while their transition-specific magnitudes remain
state-control dependent.

\noindent\textbf{Static Mechanism Prior.}
Finally, the semantic group of each event is converted into a
smoothed static mechanism prior
\begin{equation}
    \boldsymbol{\alpha}^{0}_e
    \in
    \Delta^{M-1},
\end{equation}
where the current benchmark uses $M=3$ mechanism families ordered as
physical, chemical, and coupled.
For smoothing coefficient $\epsilon_{\alpha}$, the prior assigned to
the rule-defined family $g_e$ is
\begin{equation}
    \alpha^{0}_{e,m}
    =
    \begin{cases}
        1-(M-1)\epsilon_{\alpha},
        & m=g_e,\\[1mm]
        \epsilon_{\alpha},
        & m\neq g_e.
    \end{cases}
\label{eq:supp_static_family_prior}
\end{equation}
The default implementation uses
$\epsilon_{\alpha}=0.05$.
The construction and interpretation of this prior are detailed
in Sec.~\ref{sec:supp_mechanism_mapping}.

\noindent\textbf{Validation and Immutability.}
Compilation validates all required fields, species names,
control keys, probability models, yield models, angular-law
identifiers, and product relations.
Unsupported or incomplete rules raise explicit errors rather
than being silently ignored.
After compilation,
$\mathbf{R}$,
$\boldsymbol{\Phi}^{\mathrm{evt}}$,
$\boldsymbol{\alpha}^{0}$,
$\mathbf{R}^{\mathrm{mech}}$,
$\mathbf{W}^{\mathrm{src}}$,
$\mathbf{W}^{\mathrm{sink}}$, and
$\mathbf{s}^{\mathrm{dep}}$
are registered as fixed model buffers.
They are not optimized by gradient descent and remain
unchanged throughout training and autoregressive rollout.

\subsection{Benchmark Event Catalog and Parameterization}
\label{sec:supp_event_catalog}

The benchmark Event Rules Library contains $N_e=37$
configured event channels.
The catalog is organized into 22 primary events whose
intensities are driven by external process controls and
15 product-mediated deposition events whose intensities are
derived from the internal Product Pool.
The primary catalog comprises six neutral-chlorine surface
events and sixteen ion-induced events associated with
Cl$^{+}$, Cl$_2^{+}$, and Ar$^{+}$.
The catalog is implementation specific: its numerical values,
event decomposition, and product assignments follow the
released configuration used in all experiments.
Established chlorine--silicon feature-profile mechanisms are
used as physical references for the reaction classes and
kinetic parameter roles
\cite{huard2018nanoscale,zhang2017investigation}.

\noindent\textbf{Neutral-Chlorine Surface Events.}
All neutral events are driven by the neutral-Cl supply control,
use a constant interaction probability, and use a unit
constant yield.
They describe progressive chlorination, reverse abstraction,
and low-probability spontaneous removal from highly
chlorinated surface states.
Table~\ref{tab:supp_neutral_catalog} lists the six configured
channels.
The ``Pool contribution'' column records the intermediate
species supplied to the Product Pool after activation of the
corresponding event.

\begin{table*}[!t]
\centering
\scriptsize
\setlength{\tabcolsep}{4pt}
\renewcommand{\arraystretch}{1.08}
\begin{tabular}{cllclc}
\toprule
ID
& Configured event
& Surface role
& $p_e^{\mathrm{const}}$
& Effect
& Pool contribution \\
\midrule
E1
& \texttt{Cl\_on\_Si\_to\_SiCl}
& $\mathrm{Si}\rightarrow\mathrm{SiCl}$
& $0.55$
& React
& $\mathrm{SiCl}$ \\

E2
& \texttt{Cl\_on\_SiCl\_to\_SiCl2}
& $\mathrm{SiCl}\rightarrow\mathrm{SiCl}_2$
& $0.30$
& React
& $\mathrm{SiCl}_2$ \\

E3
& \texttt{Cl\_on\_SiCl2\_to\_SiCl3}
& $\mathrm{SiCl}_2\rightarrow\mathrm{SiCl}_3$
& $0.20$
& React
& $\mathrm{SiCl}_3$ \\

E4
& \texttt{Cl\_on\_SiCl2\_to\_SiCl\_back}
& $\mathrm{SiCl}_2\rightarrow\mathrm{SiCl}$
& $0.02$
& React
& $\mathrm{SiCl}$ \\

E5
& \texttt{Cl\_on\_SiCl3\_etch}
& Removal from $\mathrm{SiCl}_3$
& $0.0001$
& Etch
& $\mathrm{SiCl}_3$ \\

E6
& \texttt{Cl\_on\_SiCl3\_to\_SiCl2}
& $\mathrm{SiCl}_3\rightarrow\mathrm{SiCl}_2$
& $0.08$
& React
& $\mathrm{SiCl}_2$ \\
\bottomrule
\end{tabular}
\caption{
Neutral-chlorine events in the benchmark Event Rules Library.
All six channels use the neutral-Cl supply control and a
constant unit yield.
}
\label{tab:supp_neutral_catalog}
\end{table*}

These probabilities are the values used by the released
benchmark configuration and should not be interpreted as a
verbatim reproduction of any single published reaction table.
The literature mechanism motivates the progressive
chlorination and abstraction structure, whereas the final
values reflect the reduced event representation used in the
benchmark simulator and neural prior compiler.

\noindent\textbf{Ion-Induced Events.}
Ion-induced channels gather the flux, incidence, and
mean-energy controls associated with their projectile.
Every such event uses the energy--angle-dependent probability
model with exponent
\begin{equation}
    n_e=0.5.
\end{equation}
For compact reporting, Table~\ref{tab:supp_ion_catalog}
denotes the probability parameters by
\begin{equation}
    \mathcal{Q}_e
    =
    \left(
        E^{\mathrm{th}}_e,
        E^{\mathrm{ref}}_e,
        p^0_e
    \right),
\end{equation}
where the two energy entries are reported in eV.
The angular identifier is either
$\mathrm{P}$ for the physical-sputtering angular law or
$\mathrm{C}$ for the chemical-sputtering angular law.

A Yamamura-style yield entry is abbreviated as
\begin{equation}
    \mathcal{Y}_e
    =
    \left(
        E^{Y,\mathrm{th}}_e,
        \theta^{\max}_e,
        Y^{\max}_e,
        Y^0_e
    \right),
\end{equation}
where energy is reported in eV and angle in degrees.
The symbol $1$ denotes a constant unit yield.
In the Pool column, ``$\times Y$'' indicates that the
product-pool contribution is scaled by the event-specific
yield; otherwise, the configured coefficient is one.

\begin{table*}[!t]
\centering
\scriptsize
\setlength{\tabcolsep}{3.2pt}
\renewcommand{\arraystretch}{1.06}
\begin{tabular}{cllcccccc}
\toprule
ID
& Projectile
& Target / branch
& Group
& $\mathcal{Q}_e$
& Ang.
& $\mathcal{Y}_e$
& Pool species
& Coeff. \\
\midrule
E7
& $\mathrm{Cl}^{+}$
& $\mathrm{Si}$ / P
& \texttt{physical}
& $(25,100,0.05)$
& P
& $(25,60,2.2,1.2)$
& $\mathrm{SiCl}$
& $Y$ \\

E8
& $\mathrm{Cl}^{+}$
& $\mathrm{SiCl}$ / P
& \texttt{both}
& $(35,100,0.10)$
& P
& $(35,60,2.8,1.5)$
& $\mathrm{SiCl}_2$
& $Y$ \\

E9
& $\mathrm{Cl}^{+}$
& $\mathrm{SiCl}$ / C
& \texttt{both}
& $(10,100,0.20)$
& C
& $1$
& $\mathrm{SiCl}_2$
& $1$ \\

E10
& $\mathrm{Cl}^{+}$
& $\mathrm{SiCl}_2$ / C
& \texttt{both}
& $(10,100,0.50)$
& C
& $1$
& $\mathrm{SiCl}_2$
& $1$ \\

E11
& $\mathrm{Cl}^{+}$
& $\mathrm{SiCl}_3$ / C
& \texttt{both}
& $(10,100,0.50)$
& C
& $1$
& $\mathrm{SiCl}_3$
& $1$ \\

E12
& $\mathrm{Cl}^{+}$
& Resist / P
& \texttt{physical}
& $(15,150,0.01)$
& P
& $(15,60,1.2,1.0)$
& $\mathrm{SiCl}$
& $Y$ \\
\midrule

E13
& $\mathrm{Cl}_2^{+}$
& $\mathrm{Si}$ / P
& \texttt{physical}
& $(25,100,0.02)$
& P
& $(25,60,2.2,1.2)$
& $\mathrm{SiCl}$
& $Y$ \\

E14
& $\mathrm{Cl}_2^{+}$
& $\mathrm{SiCl}$ / C
& \texttt{both}
& $(10,100,0.20)$
& C
& $1$
& $\mathrm{SiCl}_2$
& $1$ \\

E15
& $\mathrm{Cl}_2^{+}$
& $\mathrm{SiCl}_2$ / C
& \texttt{both}
& $(10,100,0.25)$
& C
& $1$
& $\mathrm{SiCl}_2$
& $1$ \\

E16
& $\mathrm{Cl}_2^{+}$
& $\mathrm{SiCl}_3$ / C
& \texttt{both}
& $(10,100,0.25)$
& C
& $1$
& $\mathrm{SiCl}_3$
& $1$ \\

E17
& $\mathrm{Cl}_2^{+}$
& Resist / P
& \texttt{physical}
& $(15,150,0.01)$
& P
& $(15,60,1.2,1.0)$
& $\mathrm{SiCl}$
& $Y$ \\
\midrule

E18
& $\mathrm{Ar}^{+}$
& $\mathrm{Si}$ / P
& \texttt{physical}
& $(25,100,0.05)$
& P
& $(25,60,2.2,1.2)$
& $\mathrm{SiCl}$
& $Y$ \\

E19
& $\mathrm{Ar}^{+}$
& $\mathrm{SiCl}$ / C
& \texttt{both}
& $(10,100,0.20)$
& C
& $1$
& $\mathrm{SiCl}$
& $1$ \\

E20
& $\mathrm{Ar}^{+}$
& $\mathrm{SiCl}_2$ / C
& \texttt{both}
& $(10,100,0.50)$
& C
& $1$
& $\mathrm{SiCl}_2$
& $1$ \\

E21
& $\mathrm{Ar}^{+}$
& $\mathrm{SiCl}_3$ / C
& \texttt{both}
& $(10,100,0.50)$
& C
& $1$
& $\mathrm{SiCl}_3$
& $1$ \\

E22
& $\mathrm{Ar}^{+}$
& Resist / P
& \texttt{physical}
& $(15,150,0.01)$
& P
& $(15,60,1.2,1.0)$
& $\mathrm{SiCl}$
& $Y$ \\
\bottomrule
\end{tabular}
\caption{
Ion-induced primary events in the benchmark Event Rules
Library.
For all channels, $n_e=0.5$.
The YAML label \texttt{both} identifies events assigned to the
coupled static mechanism prior.
}
\label{tab:supp_ion_catalog}
\end{table*}

The catalog explicitly separates physical and
chemically enhanced branches when both are supported for the
same projectile--target pair.
For example, Cl$^{+}$ interaction with SiCl is represented by
two distinct channels:
a higher-threshold physical branch and a lower-threshold
chemical branch.
This separation allows EIC to evaluate different kinetic
forms while allowing MAC to route their contributions through
a shared coupled-mechanism prior.

\noindent\textbf{Product-Mediated Redeposition Events.}
The remaining fifteen channels represent redeposition from
the Product Pool.
They are generated as the Cartesian product of three
intermediate product species and five target classes:
\begin{equation}
\begin{aligned}
    \mathcal{S}_{\mathrm{dep}}
    &=
    \{
        \mathrm{SiCl},
        \mathrm{SiCl}_2,
        \mathrm{SiCl}_3
    \},\\
    \mathcal{T}_{\mathrm{dep}}
    &=
    \{
        \mathrm{Si},
        \mathrm{SiCl},
        \mathrm{SiCl}_2,
        \mathrm{SiCl}_3,
        \mathrm{Resist}
    \}.
\end{aligned}
\end{equation}
Every pair
$(s,q)\in
\mathcal{S}_{\mathrm{dep}}\times
\mathcal{T}_{\mathrm{dep}}$
defines one deposition event, producing
$3\times5=15$ channels.
All use
\texttt{dep\_model=from\_products\_pool},
are assigned to the \texttt{chemical} group, and use the
configured sticking coefficient
\begin{equation}
    p^{\mathrm{stick}}_{s,q}
    =
    0.02.
\end{equation}

\begin{table}[!t]
\centering
\footnotesize
\setlength{\tabcolsep}{4pt}
\renewcommand{\arraystretch}{1.05}
\begin{tabular}{lccc}
\toprule
Pool species
& Target classes
& Channels
& $p^{\mathrm{stick}}$ \\
\midrule
$\mathrm{SiCl}$
& All five
& E23--E27
& $0.02$ \\

$\mathrm{SiCl}_2$
& All five
& E28--E32
& $0.02$ \\

$\mathrm{SiCl}_3$
& All five
& E33--E37
& $0.02$ \\
\bottomrule
\end{tabular}
\caption{
Product-mediated deposition events.
``All five'' denotes
$\{\mathrm{Si},\mathrm{SiCl},\mathrm{SiCl}_2,
\mathrm{SiCl}_3,\mathrm{Resist}\}$.
}
\label{tab:supp_deposition_catalog}
\end{table}

These deposition events do not gather an external process
flux.
Their base intensities are computed from the amount of the
required species accumulated in the Product Pool and the
corresponding sticking coefficient.
This preserves the directed dependency
\begin{equation}
    \text{primary event}
    \rightarrow
    \text{product pool}
    \rightarrow
    \text{deposition event}
\end{equation}
inside the executable transition.

\noindent\textbf{Reduced-Order Product Bookkeeping.}
The product entries in the library are computational
bookkeeping channels used to connect primary removal events
to downstream redeposition.
They should not be interpreted as a complete gas-phase
reaction mechanism or exact microscopic stoichiometry.
A numerical coefficient contributes a fixed multiple of the
primary-event intensity
$\boldsymbol{\lambda}^{p}_{t,e}$, whereas the special
coefficient \texttt{yield} contributes an amount proportional
to the event-specific sputtering yield.
This reduced representation retains the source--sink
structure needed by the Product Pool without introducing a
separate gas-phase transport model into \method{}.

In the current configuration, yield-scaled physical-removal
and resist-erosion channels are mapped to the configured
SiCl pool as a reduced-order closure.
This assignment is fixed before training and is applied
identically to every model run.
It therefore affects the structural dependency graph but is
not fitted using the benchmark trajectories.

\noindent\textbf{Parameter Provenance.}
The probability and reaction categories are informed by
published chlorine--silicon feature-profile mechanisms, which
distinguish direct physical sputtering from lower-threshold
chemical sputtering of chlorinated surfaces and include
finite-probability SiCl$_x$ redeposition
\cite{huard2018nanoscale,zhang2017investigation}.
However, the exact values in
Tables~\ref{tab:supp_neutral_catalog}--\ref{tab:supp_deposition_catalog}
are those stored in the released benchmark configuration.
This configuration, rather than the external literature
tables, defines the executable event catalog used to generate
all NEMSim structural inputs and experimental results.

\subsection{Kinetic Probability and Angular Models}
\label{sec:supp_kinetic_models}

The kinetic attributes in the Event Rules Library determine
the formula-guided component of each primary event intensity.
They specify how the formula-guided base intensity of each
event depends on particle supply, incident energy, and incidence
orientation before any learned state-dependent correction is
applied.
The implemented forms are differentiable adaptations of the
threshold-energy and angular models used in Monte Carlo
feature-profile simulation
\cite{huard2018nanoscale,zhang2017investigation}.

\noindent\textbf{Event-Specific Kinetic Inputs.}
The following equations give the component-wise realization
of the tensorized EIC computation in the main paper and are
evaluated in parallel over all primary event channels.
The quantities $f_{t,e}$, $E_{t,e}$, and $c_{t,e}$ are the
processed component-wise counterparts of
$\mathbf{u}^{f}_{t,e}$, $\mathbf{u}^{e}_{t,e}$, and
$\mathbf{u}^{d}_{t,e}$, respectively.

For each primary event $e$, the rule-defined control indices
select a supply input, a mean-energy input, and, when
applicable, a directional input from $\mathbf{u}_t$.
The nonnegative supply and effective energy are
\begin{equation}
\begin{gathered}
    f_{t,e}
    =
    \operatorname{softplus}
    \left(
        u_{t,\iota^{f}_e}
    \right),\\
    E_{t,e}
    =
    \operatorname{softplus}
    \left(
        u_{t,\iota^{E}_e}
    \right).
\end{gathered}
\label{eq:supp_kinetic_inputs}
\end{equation}
The directional control is first mapped to a nonnegative
concentration variable,
\begin{equation}
    k_{t,e}
    =
    \operatorname{softplus}
    \left(
        u_{t,\iota^{\theta}_e}
    \right),
\label{eq:supp_angle_concentration}
\end{equation}
and then converted into an effective incidence cosine:
\begin{equation}
    c_{t,e}
    =
    \frac{
        k_{t,e}+1
    }{
        k_{t,e}+2+\epsilon
    }.
\label{eq:supp_effective_cosine}
\end{equation}
Thus, $c_{t,e}\in(0,1)$ provides a compact differentiable
summary of the event-specific incidence orientation.

The configuration also reserves an energy-spread control
index.
The current deterministic kinetic compiler uses only the
mean-energy input; the energy-spread channel is retained in
the schema for future distributional or quadrature-based
extensions.

\noindent\textbf{Smooth Validity Gates.}
The source feature-profile model uses hard threshold
conditions for energy-activated reactions.
To preserve differentiability, the implementation replaces
each hard indicator with
\begin{equation}
    \mathcal{G}_{\tau}(x;a)
    =
    \sigma
    \left(
        \frac{x-a}{\tau}
    \right),
\label{eq:supp_smooth_gate}
\end{equation}
where $a$ is the corresponding threshold and $\tau$ controls
the transition width.
The default implementation uses an energy-gate temperature
$\tau_E=0.5$ and an angular-gate temperature
$\tau_c=0.05$.

\noindent\textbf{Interaction-Probability Models.}
For events with
\texttt{prob\_model=const}, the interaction probability is
\begin{equation}
    p_{t,e}
    =
    \operatorname{clip}_{[0,1]}
    \left(
        p_e^{\mathrm{const}}
    \right).
\label{eq:supp_constant_probability}
\end{equation}

For events with
\texttt{prob\_model=probModel}, the normalized energetic
activation is
\begin{equation}
    x_{t,e}
    =
    \operatorname{clip}_{[0,1]}
    \left(
        \frac{
            E_{t,e}-E^{\mathrm{th}}_e
        }{
            E^{\mathrm{ref}}_e
            -E^{\mathrm{th}}_e
            +\epsilon
        }
    \right).
\label{eq:supp_energy_activation}
\end{equation}
Consequently, the energetic contribution increases from zero
near $E^{\mathrm{th}}_e$ and saturates when
$E_{t,e}\geq E^{\mathrm{ref}}_e$.
The bounded probability is
\begin{equation}
\begin{aligned}
    p_{t,e}
    =
    \operatorname{clip}_{[0,1]}
    \Bigl[
        &p_e^0
        x_{t,e}^{\,n_e}
        F^{a_e}(c_{t,e})\\
        &\times
        \mathcal{G}_{\tau_E}
        (E_{t,e};E^{\mathrm{th}}_e)
        G_e^0
    \Bigr],
\end{aligned}
\label{eq:supp_energy_angle_probability}
\end{equation}
where
\begin{equation}
    G_e^0
    =
    \sigma
    \left(
        \frac{p_e^0}{\tau_E}
    \right)
\label{eq:supp_probability_validity}
\end{equation}
is a smooth parameter-validity gate, and
$a_e\in\{\mathrm{None},\mathrm{P},\mathrm{C}\}$
selects the angular model.
The benchmark configuration uses $n_e=0.5$ for all
energy-dependent events, consistent with the square-root
energy dependence used in the reference feature-profile
mechanism
\cite{zhang2017investigation}.

Compared with the source Monte Carlo expression, the
implemented form introduces three numerical safeguards:
smooth threshold activation, saturation at the reference
energy, and clipping to the probability interval.
These modifications preserve the intended rule dependence
while preventing unbounded values and discontinuous
gradients during neural-network optimization.

\noindent\textbf{Physical Angular Factor.}
For events marked with \texttt{angle\_kind=P}, the code uses
a Yamamura-style physical-sputtering angular profile.
Let
\begin{equation}
\begin{gathered}
    c_{\mathrm{P}}
    =
    \cos 60^{\circ},\\
    R_{\mathrm{P}}
    =
    \frac{1.0}{0.2}.
\end{gathered}
\label{eq:supp_physical_angle_constants}
\end{equation}
The corresponding shape parameters are
\begin{equation}
\begin{gathered}
    q_{\mathrm{P}}
    =
    -
    \frac{
        \log R_{\mathrm{P}}
    }{
        \log c_{\mathrm{P}}
        +1-c_{\mathrm{P}}
        +\epsilon
    },\\
    s_{\mathrm{P}}
    =
    q_{\mathrm{P}}c_{\mathrm{P}}.
\end{gathered}
\label{eq:supp_physical_angle_shape}
\end{equation}
Using
\begin{equation}
    c^{\epsilon}_{t,e}
    =
    \max(c_{t,e},\epsilon),
\end{equation}
the physical angular factor is
\begin{equation}
\begin{aligned}
    F^{\mathrm{P}}(c_{t,e})
    &=
    \left(
        c^{\epsilon}_{t,e}
    \right)^{-q_{\mathrm{P}}}\\
    &\quad\times
    \exp
    \left[
        -s_{\mathrm{P}}
        \left(
            \frac{1}{c^{\epsilon}_{t,e}}-1
        \right)
    \right]\\
    &\quad\times
    \mathcal{G}_{\tau_c}
    (c_{t,e};\epsilon).
\end{aligned}
\label{eq:supp_physical_angle_factor}
\end{equation}
This form equals approximately one at normal incidence,
increases toward an oblique-angle maximum, and is smoothly
suppressed near grazing incidence.
It corresponds to the physical-sputtering angular category
denoted by \texttt{P} in the source reaction mechanism
\cite{huard2018nanoscale}.

\noindent\textbf{Chemical Angular Factor.}
For events marked with \texttt{angle\_kind=C}, the code uses
a smoothed piecewise chemical-sputtering profile.
Let
\begin{equation}
    \theta_{t,e}
    =
    \arccos(c_{t,e}),
\end{equation}
and define
\begin{equation}
\begin{gathered}
    \theta_{\mathrm{C}}
    =
    45^{\circ},\\
    c_{\mathrm{C}}
    =
    \cos\theta_{\mathrm{C}}.
\end{gathered}
\label{eq:supp_chemical_angle_constants}
\end{equation}
The smooth transition weight is
\begin{equation}
    \omega^{\mathrm{C}}_{t,e}
    =
    \sigma
    \left(
        \frac{
            \theta_{\mathrm{C}}
            -\theta_{t,e}
        }{
            \tau_c
        }
    \right).
\label{eq:supp_chemical_angle_weight}
\end{equation}
The chemical angular factor is
\begin{equation}
\begin{aligned}
    F^{\mathrm{C}}(c_{t,e})
    &=
    \Biggl[
        \omega^{\mathrm{C}}_{t,e}\\
        &\quad+
        \left(
            1-\omega^{\mathrm{C}}_{t,e}
        \right)
        \frac{c_{t,e}}{c_{\mathrm{C}}}
    \Biggr]\\
    &\quad\times
    \mathcal{G}_{\tau_c}
    (c_{t,e};\epsilon).
\end{aligned}
\label{eq:supp_chemical_angle_factor}
\end{equation}
The factor is approximately constant for incidence angles
below $45^{\circ}$ and decreases proportionally to the
incidence cosine at larger angles.
For events with no angular dependence,
\begin{equation}
    F^{\mathrm{None}}(c)=1.
\label{eq:supp_no_angle_factor}
\end{equation}

\noindent\textbf{Yield Models.}
For events with
\texttt{yield\_model=const}, the nonnegative yield multiplier
is
\begin{equation}
    Y_{t,e}
    =
    \max
    \left(
        Y^{\mathrm{const}}_e,
        0
    \right).
\label{eq:supp_constant_yield}
\end{equation}

For events with
\texttt{yield\_model=yamamura}, the rule provides
$E^{Y,\mathrm{th}}_e$,
$\theta^{\max}_e$,
$Y^{\max}_e$, and
$Y^0_e$.
Define
\begin{equation}
\begin{gathered}
    c^{\max}_e
    =
    \cos\theta^{\max}_e,\\
    R^Y_e
    =
    \frac{
        Y^{\max}_e
    }{
        \max(Y^0_e,\epsilon)
    }.
\end{gathered}
\label{eq:supp_yield_ratio}
\end{equation}
The event-specific angular-shape parameters are
\begin{equation}
\begin{gathered}
    q^Y_e
    =
    -
    \frac{
        \log R^Y_e
    }{
        \log c^{\max}_e
        +1-c^{\max}_e
        +\epsilon
    },\\
    s^Y_e
    =
    q^Y_e c^{\max}_e.
\end{gathered}
\label{eq:supp_yield_shape_parameters}
\end{equation}
The resulting Yamamura-style yield multiplier is
\begin{equation}
\begin{aligned}
    Y_{t,e}
    &=
    \left(
        c^{\epsilon}_{t,e}
    \right)^{-q^Y_e}\\
    &\quad\times
    \exp
    \left[
        -s^Y_e
        \left(
            \frac{1}{c^{\epsilon}_{t,e}}-1
        \right)
    \right]\\
    &\quad\times
    \mathcal{G}_{\tau_E}
    (E_{t,e};E^{Y,\mathrm{th}}_e)\\
    &\quad\times
    \mathcal{G}_{\tau_c}
    (c_{t,e};\epsilon).
\end{aligned}
\label{eq:supp_yamamura_yield}
\end{equation}

In the current implementation,
$Y^{\max}_e$ and $Y^0_e$ determine the relative angular
shape through their ratio.
Equation~\eqref{eq:supp_yamamura_yield} is normalized to
approximately one at normal incidence before applying the
smooth gates; it is therefore an event-yield multiplier rather
than an independently calibrated absolute sputter yield.
Its scale is combined with the event probability and supply
inside the final base intensity.

\noindent\textbf{Normalized Energy Coordinates.}
The default model evaluates event energies and kinetic
thresholds in the normalized coordinate system used by the
process-control vector.
For an energy channel $j$, a physical parameter $v$ is mapped
as
\begin{equation}
    \mathcal{N}_{j}(v)
    =
    2
    \frac{
        v-\ell_j
    }{
        d_j
    }
    -1,
\label{eq:supp_energy_normalization}
\end{equation}
where $\ell_j$ and $d_j$ are the lower bound and range used
to normalize that control channel.
The same transformation is applied to
$E^{\mathrm{th}}_e$,
$E^{\mathrm{ref}}_e$, and
$E^{Y,\mathrm{th}}_e$ before evaluating the kinetic formulas.
This ensures that the event parameters and gathered energy
controls are expressed in the same coordinate system.

\noindent\textbf{Formula-Guided Base Intensity.}
After selecting the appropriate probability, angular, and
yield models, the primary base intensity is
\begin{equation}
    \lambda^{b}_{t,e}
    =
    f_{t,e}
    p_{t,e}
    Y_{t,e}.
\label{eq:supp_kinetic_base_intensity}
\end{equation}
Product-mediated deposition events do not evaluate
Eqs.~\eqref{eq:supp_constant_probability}
through~\eqref{eq:supp_yamamura_yield}.
Their dependency-driven intensities
$\boldsymbol{\lambda}^{\mathrm{dep}}_{t,e}$ are instead
computed by the Product Pool from the generated-product
amounts and rule-defined sticking coefficients.

Equation~\eqref{eq:supp_kinetic_base_intensity} provides the
fixed formula-guided component of EIC.
The bounded log-rate correction and the subsequent
product-dependency computation are detailed in
Sec.~\ref{sec:supp_eic}.
Thus, the kinetic models determine the admissible control and
event-parameter dependence, while the learned branch adapts
their magnitudes without replacing the underlying event-rule
structure.

\subsection{Mechanism-Family Mapping and Prior Construction}
\label{sec:supp_mechanism_mapping}

The Event Rules Library describes event-specific reactants,
targets, kinetic forms, and product dependencies.
However, many distinct events exert their effects through a
smaller set of recurring physical processes.
The rule compiler therefore maps the event catalog onto three
shared mechanism families,
\begin{equation}
\begin{gathered}
\mathcal{M}=\{1,2,3\},\\[-2pt]
(1,2,3)
\leftrightarrow
(\mathrm{physical},\mathrm{chemical},\mathrm{coupled}).
\end{gathered}
\label{eq:supp_mechanism_set}
\end{equation}
which are used as the shared mechanism families of the
Mechanism Attribution Compiler.

The mechanism families are functional modeling abstractions
rather than microscopic reaction labels directly observed in
the trajectory data.
Their interpretation is informed by the established separation
between direct momentum-transfer sputtering, surface-chemical
transformation, and ion-assisted chemical removal in
chlorine-based plasma etching
\cite{huard2018nanoscale,zhang2017investigation}.
Their executable definitions, however, are determined by the
compiled attributes of the benchmark Event Rules Library.

\noindent\textbf{Mechanism-Family Semantics.}
The physical family represents effects dominated by energetic
particle impact, momentum transfer, incidence geometry, and
direct removal of surface material.
Events assigned to this family commonly involve an ion
projectile, physical angular dependence, or a
Yamamura-style sputtering-yield model.
Representative examples include direct Ar$^{+}$ or
Cl$^{+}$ sputtering of Si and energetic erosion of resist.

The chemical family represents transformations governed
primarily by reactive-species availability and the local
chemical state of the surface.
It includes progressive chlorination, abstraction,
spontaneous chemical removal, and attachment of
product species.
These events are typically associated with neutral
projectiles, constant reaction probabilities, chemical
angular dependence, or product-mediated deposition.

The coupled family represents effects that cannot be
attributed to an isolated physical or chemical pathway.
It includes ion-induced removal of chlorinated surface
species, chemically enhanced sputtering, and dependencies
between product-generating and product-consuming events.
The coupled family therefore describes the interaction of
energetic activation, surface chemistry, and event-to-event
product transfer.

\begin{table*}[!t]
\centering
\footnotesize
\setlength{\tabcolsep}{4pt}
\renewcommand{\arraystretch}{1.18}

\begin{tabular}{
    p{0.13\textwidth}
    p{0.27\textwidth}
    p{0.29\textwidth}
    p{0.23\textwidth}
}
\toprule
Mechanism
& Functional interpretation
& Compiled rule evidence
& Representative event roles \\
\midrule

Physical
&
Direct energy or momentum transfer from an energetic
projectile to the surface.
&
Ion-driven event; physical angular type \texttt{P};
Yamamura-style yield; sputtering or resist-erosion role.
&
Direct sputtering of Si; physical removal from a surface;
energetic resist erosion. \\

Chemical
&
Surface transformation governed primarily by reactant
availability and surface composition.
&
Neutral-driven reaction; chemical angular type \texttt{C};
chlorination, abstraction, spontaneous reaction, or
deposition role.
&
Si chlorination; Cl abstraction; spontaneous removal;
SiCl$_x$ attachment. \\

Coupled
&
Response requiring interaction between energetic activation,
surface chemistry, or product-mediated event dependencies.
&
Rule group \texttt{both}; ion interaction with a chlorinated
target; product-source or product-sink relation.
&
Ion-assisted removal of SiCl$_x$; chemically enhanced
sputtering; generation and redeposition of products. \\

\bottomrule
\end{tabular}

\caption{
Mechanism-family interpretation and the event-rule attributes
used by the compiler.
The listed attributes provide structural evidence rather than
event-level supervision.
}
\label{tab:supp_mechanism_mapping}
\end{table*}

\noindent\textbf{Semantic Group-to-Family Mapping.}
Each event rule contains a categorical
\texttt{group} field with one of three values:
\texttt{physical}, \texttt{chemical}, or \texttt{both}.
The compiler maps these labels to the mechanism indices
\begin{equation}
    g_e
    =
    \begin{cases}
        1,
        & \texttt{group}_e=\texttt{physical},\\
        2,
        & \texttt{group}_e=\texttt{chemical},\\
        3,
        & \texttt{group}_e=\texttt{both}.
    \end{cases}
\label{eq:supp_group_mapping}
\end{equation}
The third label is mapped to the coupled family.
It does not denote an arithmetic average or an equal mixture
of the physical and chemical families.
Instead, it identifies an event whose modeled effect depends
on their interaction, such as ion-induced removal of a
chlorinated surface species.

This categorical mapping provides a default semantic anchor,
but it is not the complete dynamic event-to-mechanism
attribution.
The final attribution is allowed to vary with event intensity,
state, and process control, as described in
Sec.~\ref{sec:supp_mac}.

\noindent\textbf{Smoothed Static Attribution Prior.}
For each event, the compiler converts $g_e$ into a smoothed
mechanism prior
\begin{equation}
    \boldsymbol{\alpha}^{0}_e
    =
    \left[
        \alpha^{0}_{e,1},
        \ldots,
        \alpha^{0}_{e,M}
    \right],
\end{equation}
where $M=3$.
For smoothing coefficient $\epsilon_{\alpha}$,
\begin{equation}
    \alpha^{0}_{e,m}
    =
    \begin{cases}
        1-(M-1)\epsilon_{\alpha},
        & m=g_e,\\[1mm]
        \epsilon_{\alpha},
        & m\neq g_e.
    \end{cases}
\label{eq:supp_smoothed_mechanism_prior}
\end{equation}
The default implementation uses
\begin{equation}
    \epsilon_{\alpha}=0.05,
\end{equation}
giving a prior mass of $0.90$ to the rule-defined family and
$0.05$ to each alternative family.

Smoothing keeps the static rule prior strictly positive before
event-attribute compatibility is applied.
It therefore prevents the categorical group assignment itself
from imposing a hard zero.
The dominant prior entry nevertheless strongly biases each
event toward its rule-supported mechanism family.
Consequently, MAC can adapt an attribution when required by
the observed state and controls without discarding the
structural preference encoded by the event rule.

Stacking the event-wise priors gives
\begin{equation}
    \boldsymbol{\alpha}^{0}
    =
    \begin{bmatrix}
        (\boldsymbol{\alpha}^{0}_1)^{\top}\\
        \vdots\\
        (\boldsymbol{\alpha}^{0}_{N_e})^{\top}
    \end{bmatrix}
    \in
    \mathbb{R}^{N_e\times M}.
\label{eq:supp_prior_matrix}
\end{equation}
The matrix $\boldsymbol{\alpha}^{0}$ is compiled once and
registered as a fixed model buffer.

\noindent\textbf{Deterministic Mechanism-Role Signals.}
The semantic group supplies one dominant prior family, whereas
the detailed event attributes can provide evidence for
multiple functional roles.
The compiler therefore derives an additional multi-hot
mechanism-role vector
\begin{equation}
    \mathbf{r}^{\mathrm{mech}}_e
    =
    \left[
        r^{\mathrm{phys}}_e,
        r^{\mathrm{chem}}_e,
        r^{\mathrm{coup}}_e
    \right]
    \in
    \{0,1\}^{3}.
\label{eq:supp_mechanism_role_vector}
\end{equation}

Let the binary meta-features defined in
Sec.~\ref{sec:supp_rule_schema} indicate whether event $e$
is ion driven, neutral driven, uses physical or chemical
angular dependence, uses an angular yield model, represents
sputtering or reaction, acts on a chlorinated target, or
participates in a product dependency. Here,
$m^{\mathrm{physgrp}}_e$ and
$m^{\mathrm{both}}_e$
indicate the \texttt{physical} and \texttt{both} semantic
groups;
$m^{\mathrm{src}}_e$ and
$m^{\mathrm{sink}}_e$
indicate product-generation and product-consumption roles;
and
$m^{\mathrm{chlor}}_e$
indicates a chlorinated silicon target.
The remaining flags follow the event attributes described in
Sec.~\ref{sec:supp_rule_schema}.
The physical role signal is activated by
\begin{equation}
\begin{aligned}
    r^{\mathrm{phys}}_e
    =
    \max\bigl\{
        &m^{\mathrm{P}}_e,
        m^{\mathrm{yield}}_e,\\
        &m^{\mathrm{sput}}_e,
        m^{\mathrm{ion}}_e m^{\mathrm{physgrp}}_e
    \bigr\},
\end{aligned}
\label{eq:supp_physical_role_signal}
\end{equation}
where $m^{\mathrm{P}}_e$ identifies physical angular
dependence and $m^{\mathrm{yield}}_e$ identifies an
angle-dependent sputtering yield.

The chemical role signal is
\begin{equation}
\begin{aligned}
    r^{\mathrm{chem}}_e
    =
    \max\bigl\{
        &m^{\mathrm{C}}_e,
        m^{\mathrm{react}}_e,\\
        &m^{\mathrm{neutral}}_e,
        m^{\mathrm{dep}}_e
    \bigr\},
\end{aligned}
\label{eq:supp_chemical_role_signal}
\end{equation}
where $m^{\mathrm{C}}_e$ identifies chemical angular
dependence and $m^{\mathrm{dep}}_e$ identifies a
product-mediated deposition event.

The coupled role signal is
\begin{equation}
\begin{aligned}
    r^{\mathrm{coup}}_e
    =
    \max\bigl\{
        &m^{\mathrm{both}}_e,
        m^{\mathrm{src}}_e,
        m^{\mathrm{sink}}_e,\\
        &m^{\mathrm{ion}}_e
        m^{\mathrm{chlor}}_e
    \bigr\},
\end{aligned}
\label{eq:supp_coupled_role_signal}
\end{equation}
where $m^{\mathrm{src}}_e$ and $m^{\mathrm{sink}}_e$
indicate product-generation and product-consumption roles,
respectively.
The last term identifies ion-driven events acting on a
chlorinated surface state.

Equations~\eqref{eq:supp_physical_role_signal}--%
\eqref{eq:supp_coupled_role_signal} allow one event to
support multiple mechanism roles.
For example, an ion-assisted removal event may contain both
physical and coupled evidence, while a product-mediated
deposition event may contain chemical and coupled evidence.
This multi-role representation is distinct from the
single dominant family used to initialize
$\boldsymbol{\alpha}^{0}_e$.

Stacking all role vectors gives
\begin{equation}
    \mathbf{R}^{\mathrm{mech}}
    =
    \begin{bmatrix}
        (\mathbf{r}^{\mathrm{mech}}_1)^{\top}\\
        \vdots\\
        (\mathbf{r}^{\mathrm{mech}}_{N_e})^{\top}
    \end{bmatrix}
    \in
    \{0,1\}^{N_e\times M}.
\label{eq:supp_mechanism_role_matrix}
\end{equation}
Like $\boldsymbol{\alpha}^{0}$, this matrix is fixed before training.

\noindent\textbf{Treatment of Product-Mediated Events.}
Product-mediated deposition illustrates why the categorical
prior and the mechanism-role signals are kept separate.
Its local surface action is chemical attachment, so its
semantic group and dominant prior remain chemical.
However, its activation requires an upstream event to first
generate the corresponding product species.
The compiler therefore also activates its coupled
source--sink role.

Similarly, a physical-removal event that generates a
redepositable product retains a physical prior but contributes
to the coupled dependency signal through its product-source
attribute.
This design preserves the primary local action of each event
while representing inter-event dependencies through the
coupled pathway.

\noindent\textbf{Representative Mapping Examples.}
For a direct Ar$^{+}$-on-Si physical sputtering event,
the physical angular and yield attributes produce
\begin{equation}
    g_e=1,
\qquad
r^{\mathrm{phys}}_e=1.
\end{equation}
For a neutral Cl chlorination event,
\begin{equation}
    g_e=2,
\qquad
r^{\mathrm{chem}}_e=1.
\end{equation}
For an ion-induced event acting on SiCl$_x$ through the
chemical-sputtering branch,
\begin{equation}
    g_e=3,
\qquad
r^{\mathrm{chem}}_e
=
r^{\mathrm{coup}}_e
=
1.
\end{equation}
For a SiCl$_x$ redeposition event,
\begin{equation}
    g_e=2,
\qquad
r^{\mathrm{chem}}_e
=
r^{\mathrm{coup}}_e
=
1.
\end{equation}
These examples show that the dominant prior and the
executable role signals encode complementary information.

\noindent\textbf{Use in Mechanism Attribution.}
The static prior and deterministic mechanism-role signals
enter MAC at complementary stages.
The static prior
$\boldsymbol{\alpha}^{0}_e$
is combined with learned event-attribute compatibility to
construct the state-independent anchor
$\boldsymbol{\alpha}^{\mathrm{anc}}_e$.
The deterministic role vector
$\mathbf{r}^{\mathrm{mech}}_e$
then parameterizes the rule-induced dynamic correction.

At transition step $t$, MAC first bounds the final
event-intensity vector:
\begin{equation}
    I_{t,e}
    =
    \operatorname{Bnd}
    \left(
        \boldsymbol{\lambda}_t
    \right)_e.
\label{eq:supp_rule_intensity_summary}
\end{equation}
The component-wise rule-induced correction is
\begin{equation}
    b^{\mathrm{rule}}_{t,e,m}
    =
    c_m
    I_{t,e}
    r^{\mathrm{mech}}_{e,m},
\label{eq:supp_rule_correction_summary}
\end{equation}
where $c_m$ is the learned strength associated with
mechanism family $m$.
A lightweight context branch produces
$\mathbf{b}^{\mathrm{ctx}}_{t,e}$
from the normalized event parameters, compact state context,
and process control.

The bounded correction is
\begin{equation}
    \boldsymbol{\delta}_{t,e}
    =
    \tanh
    \left(
        \mathbf{b}^{\mathrm{rule}}_{t,e}
        +
        \kappa
        \mathbf{b}^{\mathrm{ctx}}_{t,e}
    \right),
\label{eq:supp_dynamic_correction_summary}
\end{equation}
and the final attribution is
\begin{equation}
    \alpha_{t,e,m}
    =
    \operatorname{Normalize}_{m}
    \left(
        \boldsymbol{\alpha}^{\mathrm{anc}}_e
        \odot
        \exp
        \left(
            \rho
            \boldsymbol{\delta}_{t,e}
        \right)
    \right).
\label{eq:supp_dynamic_attribution_summary}
\end{equation}
Normalization produces nonnegative attributions that sum to
one over the mechanism families.
The dynamic correction therefore adapts the
state-independent anchor while retaining its structural
support.

Neither
$\boldsymbol{\alpha}^{0}$
nor
$\mathbf{R}^{\mathrm{mech}}$
is optimized from trajectory data.
They do not encode observed mechanism labels, event
occurrences, or target attribution values.
The learnable event-attribute mapping, mechanism prototypes,
rule-correction strengths, context branch, and downstream
response modules are optimized only through the final
state-transition objective.
The mechanism mapping should therefore be interpreted as a
fixed executable organization of the event catalog rather
than intermediate supervision or a claim of uniquely
identifiable microscopic mechanisms.

\section{Additional Model Details}
\label{sec:supp_model}

The main paper presents the executable event-mechanism
pathway at the level required to explain the model design.
This section provides implementation-level formulations of
the three compilation modules.
All event-dependent constants and structural masks are
compiled from the fixed Event Rules Library before training.
All learnable mappings and corrections are optimized solely
through the final state-transition loss.

\subsection{Event Intensity Compiler}
\label{sec:supp_eic}

The Event Intensity Compiler (EIC) module maps the compact
state context $\bar{\mathbf{z}}_t$, process-control vector
$\mathbf{u}_t$, and fixed Event Rules Library to a
nonnegative intensity $\lambda_{t,e}$ for each event.
The following equations provide the component-wise
implementation of the tensorized EIC computation presented
in the main paper, with all event channels evaluated in
parallel.
EIC first evaluates the rule-defined kinetic relations, then
applies a bounded event-wise log-rate correction, and finally
propagates product-mediated dependencies through the Product
Pool.

\noindent\textbf{Rule-Defined Primary Intensity.}
For each primary event, the compiled control bindings derived
from $\mathcal{C}_e$ select its supply, mean-energy, and
directional inputs from $\mathbf{u}_t$.
The corresponding interaction probability $p_{t,e}$ and
yield $Y_{t,e}$ are evaluated using the fixed kinetic models
defined in Sec.~\ref{sec:supp_kinetic_models}.
Their product gives the rule-defined base intensity
\begin{equation}
    \lambda^{b}_{t,e}
    =
    f_{t,e}
    p_{t,e}
    Y_{t,e}.
\label{eq:supp_eic_base_intensity}
\end{equation}
This is the component-wise implementation of the tensorized
base-intensity computation
$\boldsymbol{\lambda}^{b}_{t,e}$ in the main paper.

Let $d_e\in\{0,1\}$ indicate whether event $e$ is
product-mediated.
The primary-event base intensity used by EIC is
\begin{equation}
    \lambda^{p,b}_{t,e}
    =
    (1-d_e)
    \lambda^{b}_{t,e}.
\label{eq:supp_eic_primary_base}
\end{equation}
Thus, externally controlled kinetic relations contribute only
to primary events, while product-mediated events receive no
direct control-driven base intensity.

\noindent\textbf{Selective State-Control Modulation.}
The rule-defined base intensity specifies the admissible
control-event dependence, while its realizable magnitude may
vary with the current morphology and process control.
EIC therefore constructs a compact state-control context from
$\bar{\mathbf{z}}_t$ and $\mathbf{u}_t$.

A state-derived representation is first computed as
\begin{equation}
    \mathbf{a}^{z}_t
    =
    h_z
    \left(
        \bar{\mathbf{z}}_t
    \right).
\label{eq:supp_eic_state_context}
\end{equation}
The process control provides a bounded multiplicative
modulation and an additive modulation:
\begin{equation}
\begin{gathered}
    \mathbf{s}_t
    =
    \gamma
    \tanh
    \left(
        h_s(\mathbf{u}_t)
    \right),\\
    \mathbf{d}_t
    =
    \gamma
    h_d(\mathbf{u}_t),
\end{gathered}
\label{eq:supp_eic_control_modulation}
\end{equation}
where $\gamma$ controls the direct influence of the
control-modulation branches.
A feature-wise selection gate is computed as
\begin{equation}
    \mathbf{g}_t
    =
    \sigma
    \left(
        h_g
        \left(
            [
                \bar{\mathbf{z}}_t,
                \mathbf{u}_t
            ]
        \right)
    \right).
\label{eq:supp_eic_selection_gate}
\end{equation}
The resulting bounded state-control context is
\begin{equation}
    \mathbf{a}_t
    =
    \tanh
    \left(
        \mathbf{a}^{z}_t
        \odot
        \left(
            1+\mathbf{s}_t
        \right)
        +
        \mathbf{d}_t
    \right)
    \odot
    \mathbf{g}_t.
\label{eq:supp_eic_fused_context}
\end{equation}
The outer hyperbolic tangent bounds the modulated features,
while $\mathbf{g}_t$ selectively retains the dimensions
relevant to event-intensity correction.

Each event is additionally represented by the compact
rate descriptor
\begin{equation}
    \boldsymbol{\psi}_e
    =
    \left[
        \frac{g_e}{3},\,
        \frac{j^{\mathrm{proj}}_e}{7},\,
        \frac{j^{\mathrm{tgt}}_e}{4},\,
        d_e,\,
        \frac{\mu^{p}_e}{2},\,
        \frac{\mu^{Y}_e}{2},\,
        \frac{a_e}{2}
    \right]
    \in
    \mathbb{R}^{7},
\label{eq:supp_eic_rate_descriptor}
\end{equation}
where $g_e$, $j^{\mathrm{proj}}_e$, and
$j^{\mathrm{tgt}}_e$ are the compiled mechanism-role,
projectile, and target identifiers;
$\mu^{p}_e$ and $\mu^{Y}_e$ identify the probability and
yield models; and $a_e$ identifies the angular model.
The categorical identifiers are normalized using their fixed
implementation ranges.

The descriptor $\boldsymbol{\psi}_e$ is used only by the
EIC rate-modulation branch.
It is distinct from the complete event descriptor
$\boldsymbol{\phi}_e$ used by MAC and MRC.
The rule-conditioned state-control context is
\begin{equation}
    \mathbf{q}_{t,e}
    =
    \left[
        h_c(\mathbf{a}_t),\,
        \boldsymbol{\psi}_e
    \right],
\label{eq:supp_eic_event_context}
\end{equation}
where $h_c$ is a lightweight context projection.

\noindent\textbf{Bounded Log-Rate Correction.}
Given the rule-conditioned context $\mathbf{q}_{t,e}$, EIC
predicts the component-wise log-rate correction
\begin{equation}
    \Delta\ell_{t,e}
    =
    \delta_{\max}
    \tanh
    \left(
        h_{\mathrm{rate}}
        \left(
            \mathbf{q}_{t,e}
        \right)
    \right),
\label{eq:supp_log_rate_modulation}
\end{equation}
where $\delta_{\max}>0$ sets the maximum magnitude of the
log-rate correction.
Consequently,
\begin{equation}
    -\delta_{\max}
    \leq
    \Delta\ell_{t,e}
    \leq
    \delta_{\max}.
\label{eq:supp_log_rate_bound}
\end{equation}
The corrected primary-event intensity is
\begin{equation}
    \lambda^{p}_{t,e}
    =
    \lambda^{p,b}_{t,e}
    \exp
    \left(
        \Delta\ell_{t,e}
    \right).
\label{eq:supp_primary_intensity}
\end{equation}
Equations~\eqref{eq:supp_log_rate_modulation} and
\eqref{eq:supp_primary_intensity} are the component-wise
implementation of
$\Delta\boldsymbol{\ell}_{t,e}$ and
$\boldsymbol{\lambda}^{p}_{t,e}$ in the main paper.

Consequently, a nonzero primary-event base intensity can
change only by a multiplicative factor in
\begin{equation}
    \left[
        \exp(-\delta_{\max}),\,
        \exp(\delta_{\max})
    \right].
\label{eq:supp_log_rate_factor_bound}
\end{equation}
The correction cannot introduce a primary event whose
rule-defined base intensity is zero.
The final layer of $h_{\mathrm{rate}}$ is initialized to zero,
so EIC initially reproduces the rule-defined primary-event
intensities.

\noindent\textbf{Product Pool Construction.}
Primary events may generate intermediate species that drive
downstream product-mediated events.
Let $\xi_{t,e,q}\geq 0$ denote the configured contribution
coefficient of event $e$ to product species $q$:
\begin{equation}
\xi_{t,e,q}
=
\begin{cases}
\xi^{\mathrm{const}}_{e,q},
& \text{fixed source},\\
Y_{t,e},
& \text{yield-scaled source},\\
0,
& \text{otherwise}.
\end{cases}
\label{eq:supp_eic_product_coefficient}
\end{equation}
Here, the zero case indicates that event $e$ does not generate species
$q$.
The nonzero source relations and coefficient types are obtained from
the fixed product-generation attributes compiled in the Event Rules
Library.

The transition-specific Product Pool is
\begin{equation}
    \mathcal{P}^{\mathrm{pool}}_{t,q}
    =
    \eta_q
    \sum_{e=1}^{N_e}
    \lambda^{p}_{t,e}
    \xi_{t,e,q},
\label{eq:supp_product_pool}
\end{equation}
where $\eta_q\geq 0$ is an optional species-wise redeposition
scale.
The default implementation fixes
\begin{equation}
    \eta_q=1
\label{eq:supp_default_pool_scale}
\end{equation}
for every represented product species.
Product availability therefore inherits the rule-defined
kinetics and bounded state-control correction of the upstream
primary events.
The Product Pool is a reduced dependency representation
inside EIC and is not supervised by microscopic
product-transport targets.

\noindent\textbf{Dependency-Driven Event Intensities.}
For a product-mediated event $e$, let $q(e)$ denote its
required Product Pool species and let
$p^{\mathrm{stick}}_e\geq 0$ denote its rule-defined sticking
coefficient.
Its dependency-driven base intensity is
\begin{equation}
    \lambda^{d,b}_{t,e}
    =
    d_e
    p^{\mathrm{stick}}_e
    \mathcal{P}^{\mathrm{pool}}_{t,q(e)}.
\label{eq:supp_dependency_base}
\end{equation}
The factor $d_e$ masks primary events and ensures that this
branch contributes only to product-mediated events.

The same bounded event-wise log-rate correction is then
applied:
\begin{equation}
    \lambda^{d}_{t,e}
    =
    \lambda^{d,b}_{t,e}
    \exp
    \left(
        \Delta\ell_{t,e}
    \right).
\label{eq:supp_dependency_intensity}
\end{equation}
This correction captures bounded state-control-dependent
variation in the dependency-driven intensity without
altering the rule-defined source-sink connectivity.

The code-level quantity $\lambda^{d}_{t,e}$ is the
component-wise implementation of the dependency-driven
intensity
$\boldsymbol{\lambda}^{\mathrm{dep}}_{t,e}$ used in the main
paper.
The different notation retains the intermediate naming used
by the implementation while making its correspondence to the
main formulation explicit.

\noindent\textbf{Final Event-Intensity Composition.}
The final event intensity is
\begin{equation}
    \lambda_{t,e}
    =
    \lambda^{p}_{t,e}
    +
    \lambda^{d}_{t,e}.
\label{eq:supp_final_intensity}
\end{equation}
This is the component-wise implementation of the tensorized
composition
\begin{equation}
    \boldsymbol{\lambda}_{t,e}
    =
    \boldsymbol{\lambda}^{p}_{t,e}
    +
    \boldsymbol{\lambda}^{\mathrm{dep}}_{t,e}
\label{eq:supp_final_intensity_correspondence}
\end{equation}
presented in the main paper.

Because the primary and product-mediated event sets are
disjoint, only one branch contributes to each event channel.
The construction preserves nonnegativity because the
rule-defined base intensities, Product Pool values, sticking
coefficients, and exponential corrections are nonnegative.
It also preserves rule-defined kinetic support:
a primary event cannot be introduced when its kinetic base
intensity is zero, and a product-mediated event cannot be
introduced when its required Product Pool species is absent.

EIC therefore retains the predefined kinetic thresholds,
event-control bindings, and source-product-sink connectivity,
while learning only bounded transition-specific log-rate
corrections.
All intermediate intensities and Product Pool values remain
latent and receive no direct supervision.
The learnable components that produce them are optimized only
through the final state-transition objective.

\subsection{Mechanism Attribution Compiler}
\label{sec:supp_mac}

The Mechanism Attribution Compiler (MAC) module converts
the final event intensities into attributions over the
predefined mechanism families.
Let
\begin{equation}
    \boldsymbol{\lambda}_t
    =
    \left[
        \lambda_{t,1},
        \ldots,
        \lambda_{t,N_e}
    \right]
    \in
    \mathbb{R}_{\geq 0}^{N_e}
\label{eq:supp_mac_event_intensity_vector}
\end{equation}
denote the final event-intensity vector compiled by EIC.
Given $\boldsymbol{\lambda}_t$, the compact state context
$\bar{\mathbf{z}}_t$, process-control vector $\mathbf{u}_t$,
and fixed Event Rules Library, MAC produces
\begin{equation}
    \boldsymbol{\alpha}_{t,e}
    =
    \left[
        \alpha_{t,e,1},
        \ldots,
        \alpha_{t,e,M}
    \right]
    \in
    \Delta^{M-1},
\label{eq:supp_mac_attribution_constraint}
\end{equation}
where
\begin{equation}
    \Delta^{M-1}
    =
    \left\{
        \mathbf{a}\in\mathbb{R}_{\geq 0}^{M}
        \;\middle|\;
        \sum_{m=1}^{M}a_m=1
    \right\}
\label{eq:supp_mac_simplex}
\end{equation}
is the simplex over the mechanism families.
The current implementation uses $M=3$ families, ordered as
physical, chemical, and coupled.
The following equations provide the component-wise
implementation of the tensorized MAC computation in the main
paper, with all event-mechanism pairs evaluated in parallel.

MAC first constructs a state-independent attribution anchor
from the event attributes and static mechanism prior.
It then applies a rule-induced correction driven by the
bounded event-intensity signal and a lightweight
state-control context correction.

\noindent\textbf{Prior-Guided Attribution Anchor.}
Each event descriptor $\boldsymbol{\phi}_e$, defined in
Sec.~\ref{sec:supp_rule_schema}, is embedded in a shared
prototype space:
\begin{equation}
\begin{gathered}
    \mathbf{d}_e
    =
    \operatorname{Norm}
    \left(
        h_{\mathrm{desc}}
        \left(
            \boldsymbol{\phi}_e
        \right)
    \right),\\
    \bar{\mathbf{p}}_m
    =
    \operatorname{Norm}
    \left(
        \mathbf{p}_m
    \right),\\
    \ell^{\mathrm{proto}}_{e,m}
    =
    \mathbf{d}_e^{\top}
    \bar{\mathbf{p}}_m,
\end{gathered}
\label{eq:supp_mac_prototype_scores}
\end{equation}
where $\operatorname{Norm}(\cdot)$ denotes
$\ell_2$ normalization and $\mathbf{p}_m$ is the learnable
prototype of mechanism family $m$.
Collecting the compatibility scores gives
\begin{equation}
    \boldsymbol{\ell}^{\mathrm{proto}}_e
    =
    \left[
        \ell^{\mathrm{proto}}_{e,1},
        \ldots,
        \ell^{\mathrm{proto}}_{e,M}
    \right].
\label{eq:supp_mac_prototype_score_vector}
\end{equation}

The prototype scores are converted into sparse
event-attribute compatibilities:
\begin{equation}
    \boldsymbol{\beta}_e
    =
    \operatorname{sparsemax}
    \left(
        \boldsymbol{\ell}^{\mathrm{proto}}_e
    \right)
    \in
    \Delta^{M-1}.
\label{eq:supp_mac_prototype_compatibility}
\end{equation}
The compatibility is combined multiplicatively with the
fixed static prior:
\begin{equation}
    \alpha^{\mathrm{anc}}_{e,m}
    =
    \frac{
        \beta_{e,m}
        \alpha^{0}_{e,m}
    }{
        \displaystyle
        \sum_{j=1}^{M}
        \beta_{e,j}
        \alpha^{0}_{e,j}
    }.
\label{eq:supp_mac_static_anchor}
\end{equation}
The resulting
$\boldsymbol{\alpha}^{\mathrm{anc}}_e$
depends on the fixed event attributes and learnable mechanism
prototypes, but remains independent of the current state,
process controls, and event intensities.

The static prior
$\boldsymbol{\alpha}^{0}_e$
is strictly positive before prototype matching.
The sparse compatibility may assign zero weight to some
mechanism families, so the resulting anchor can be sparse.
The subsequent multiplicative correction preserves the
support of this state-independent anchor.

\noindent\textbf{Bounded Event-Intensity Signal.}
Event intensities can differ substantially in magnitude.
Before constructing the dynamic correction, MAC converts the
final event-intensity vector into a bounded normalized signal.
For
$\mathbf{x}\in\mathbb{R}_{\geq 0}^{K}$,
define
\begin{equation}
\begin{gathered}
    r_i(\mathbf{x})
    =
    \log
    \left(
        1+\max(x_i,0)
    \right),\\
    \mu(\mathbf{x})
    =
    \operatorname{sg}
    \left(
        \frac{1}{K}
        \sum_{j=1}^{K}
        r_j(\mathbf{x})
    \right),
\end{gathered}
\label{eq:supp_mac_bounded_scale}
\end{equation}
where $\operatorname{sg}(\cdot)$ denotes stop-gradient.
The bounded operator used in the main paper is
\begin{equation}
    \operatorname{Bnd}(\mathbf{x})_i
    =
    \tanh
    \left(
        \frac{
            r_i(\mathbf{x})
        }{
            \mu(\mathbf{x})+\epsilon
        }
    \right).
\label{eq:supp_mac_bounded_operator}
\end{equation}
MAC applies this operator to the final event intensities:
\begin{equation}
    I_{t,e}
    =
    \operatorname{Bnd}
    \left(
        \boldsymbol{\lambda}_t
    \right)_e.
\label{eq:supp_mac_bounded_intensity}
\end{equation}
The logarithmic transformation compresses the dynamic range,
and the detached mean provides a common transition-specific
scale.
Within each event-intensity vector, the operator preserves
the ordering of nonnegative entries and maps them to
$[0,1)$.

Because $\lambda_{t,e}$ is the final output of EIC, the
bounded signal $I_{t,e}$ contains both primary-event and
product-mediated contributions.
Product-mediated effects therefore enter MAC through the
dependency-driven event intensities compiled by EIC.
MAC does not use the Product Pool as an additional dynamic
input.

\noindent\textbf{Rule-Induced Mechanism Correction.}
The Event Rules Library provides deterministic
mechanism-role signals derived from projectile type, angular
dependence, probability and yield attributes, reaction role,
chlorinated targets, and product dependencies.
Let
\begin{equation}
    \mathbf{r}^{\mathrm{mech}}_e
    =
    \left[
        r^{\mathrm{phys}}_e,\,
        r^{\mathrm{chem}}_e,\,
        r^{\mathrm{coup}}_e
    \right]
    \in
    \{0,1\}^{M}
\label{eq:supp_mac_role_signal}
\end{equation}
denote the multi-hot role vector defined in
Sec.~\ref{sec:supp_mechanism_mapping}.

The component-wise realization of the rule-induced
correction in the main paper is
\begin{equation}
\begin{aligned}
b^{\mathrm{rule}}_{t,e,m}
&=
\operatorname{RuleCorr}_{m}
\left(I_{t,e};\mathbf{r}_e\right)\\
&=
I_{t,e}
\begin{cases}
c_{\mathrm{phys}}r^{\mathrm{phys}}_e,
& m=1,\\
c_{\mathrm{chem}}r^{\mathrm{chem}}_e,
& m=2,\\
c_{\mathrm{coup}}r^{\mathrm{coup}}_e,
& m=3.
\end{cases}
\end{aligned}
\label{eq:supp_mac_rule_components}
\end{equation}
where $c_{\mathrm{phys}}$, $c_{\mathrm{chem}}$, and
$c_{\mathrm{coup}}$ are learnable mechanism-wise strengths.
The complete rule-induced correction is
\begin{equation}
    \mathbf{b}^{\mathrm{rule}}_{t,e}
    =
    \left[
        b^{\mathrm{rule}}_{t,e,1},
        \ldots,
        b^{\mathrm{rule}}_{t,e,M}
    \right]
    \in
    \mathbb{R}^{M}.
\label{eq:supp_mac_rule_correction}
\end{equation}

The physical role is induced by energetic-incidence and
physical-yield attributes, the chemical role by neutral and
reaction attributes, and the coupled role by ion-assisted or
product-mediated dependencies.
Because product-mediated events receive dependency-driven
intensities through EIC, their transition-specific dependency
strength enters MAC through $I_{t,e}$, while the fixed source
and sink attributes determine their coupled role support.

The rule-induced correction is structurally constrained:
if event $e$ has no predefined role for mechanism family
$m$, its intensity cannot reinforce that family through
$b^{\mathrm{rule}}_{t,e,m}$.

\noindent\textbf{State-Control Context Correction.}
The explicit rule correction is complemented by a lightweight
context branch:
\begin{equation}
    \mathbf{b}^{\mathrm{ctx}}_{t,e}
    =
    h_{\mathrm{ctx}}
    \left(
        \left[
            \bar{\mathbf{r}}_e,\,
            \bar{\mathbf{z}}_t,\,
            \mathbf{u}_t
        \right]
    \right)
    \in
    \mathbb{R}^{M},
\label{eq:supp_mac_context_correction}
\end{equation}
where $\bar{\mathbf{r}}_e$ is the normalized event-parameter
vector defined in Sec.~\ref{sec:supp_rule_schema}.
The final layer of $h_{\mathrm{ctx}}$ is initialized to zero,
so the context branch initially contributes no dynamic
correction.

\noindent\textbf{Bounded Dynamic Attribution.}
The rule-induced and context corrections are combined as
\begin{equation}
    \boldsymbol{\delta}_{t,e}
    =
    \tanh
    \left(
        \mathbf{b}^{\mathrm{rule}}_{t,e}
        +
        \kappa
        \mathbf{b}^{\mathrm{ctx}}_{t,e}
    \right),
\label{eq:supp_mac_dynamic_correction}
\end{equation}
where the default implementation uses
$\kappa=0.25$.
Consequently,
$-1<\delta_{t,e,m}<1$
for every mechanism family.

The global attribution-correction strength is constrained by
\begin{equation}
    \rho
    =
    \rho_{\max}
    \sigma
    \left(
        \rho_{\mathrm{raw}}
    \right),
\label{eq:supp_mac_dynamic_strength}
\end{equation}
where the default implementation uses
$\rho_{\max}=1$ and initializes
$\rho_{\mathrm{raw}}=-1$.

For a nonzero nonnegative mechanism vector
$\mathbf{v}\in
\mathbb{R}_{\geq 0}^{M}\setminus\{\mathbf{0}\}$,
define
\begin{equation}
    \operatorname{Normalize}_{m}
    \left(
        \mathbf{v}
    \right)
    =
    \frac{
        v_m
    }{
        \displaystyle
        \sum_{j=1}^{M}v_j
    }.
\label{eq:supp_mac_normalization}
\end{equation}
The final attribution is
\begin{equation}
    \alpha_{t,e,m}
    =
    \operatorname{Normalize}_{m}
    \left(
        \boldsymbol{\alpha}^{\mathrm{anc}}_e
        \odot
        \exp
        \left(
            \rho
            \boldsymbol{\delta}_{t,e}
        \right)
    \right).
\label{eq:supp_mac_final_attribution}
\end{equation}
The argument of
$\operatorname{Normalize}_{m}$
is nonzero because
$\boldsymbol{\alpha}^{\mathrm{anc}}_e$
contains at least one positive entry and the exponential
correction is strictly positive.
The update therefore preserves nonnegativity and ensures
\begin{equation}
    \sum_{m=1}^{M}
    \alpha_{t,e,m}
    =
    1.
\label{eq:supp_mac_attribution_sum}
\end{equation}
It retains the structural preference of the
state-independent anchor while adapting the relative
attribution according to the final event intensity, current
state, and process controls.

\noindent\textbf{Event-Mechanism Driver Composition.}
Finally, EMC composes event intensity and mechanism
attribution into
\begin{equation}
    T_{t,e,m}
    =
    \lambda_{t,e}
    \alpha_{t,e,m}.
\label{eq:supp_event_mechanism_driver}
\end{equation}
Equation~\eqref{eq:supp_event_mechanism_driver} is the
component-wise implementation of the tensorized composition
$\mathbf{T}_{t,e,m}
=
\boldsymbol{\lambda}_{t,e}
\odot
\boldsymbol{\alpha}_{t,e,m}$
in the main paper.
The resulting event-mechanism driver is passed to MRC.

\subsection{Mechanism Response Compiler}
\label{sec:supp_mrc}

The Mechanism Response Compiler (MRC) module converts the
event-mechanism drivers produced by EMC into a spatial
latent-state increment.
Rather than decoding each event independently, MRC aggregates
event-level drivers into a compact set of shared
mechanism-specific response modes and composes their
state-dependent spatial responses.

Let $R_b$ denote the number of response modes assigned to
each mechanism family.
The default implementation uses $M=3$ mechanism families and
$R_b=4$ response modes per family.
Let
\begin{equation}
    \mathbf{T}_t
    =
    \left[
        T_{t,e,m}
    \right]_{
        e=1,\ldots,N_e;\,
        m=1,\ldots,M
    }.
\label{eq:supp_mrc_driver_tensor}
\end{equation}
denote the event-mechanism drivers defined in
Eq.~\eqref{eq:supp_event_mechanism_driver}.
MRC computes
\begin{equation}
    \Delta\mathbf{z}_t
    =
    \mathrm{MRC}
    \left(
        \mathbf{z}_t,\,
        \bar{\mathbf{z}}_t,\,
        \mathbf{u}_t,\,
        \mathbf{T}_t
    \right).
\label{eq:supp_mrc_overview}
\end{equation}
The following equations provide the component-wise
implementation of the tensorized MRC computation in the main
paper.
All event, mechanism, and response-mode channels are
evaluated in parallel.

\noindent\textbf{Event-to-Modes Signature.}
Events attributed to the same mechanism family need not
produce identical responses.
MRC maps the fixed event descriptor
$\boldsymbol{\phi}_e$, defined in
Sec.~\ref{sec:supp_rule_schema}, to signature logits:
\begin{equation}
    \mathbf{L}^{S}_e
    =
    h_{\mathrm{sig}}
    \left(
        \boldsymbol{\phi}_e
    \right)
    \in
    \mathbb{R}^{M\times R_b},
\label{eq:supp_mrc_signature_logits}
\end{equation}
where $h_{\mathrm{sig}}$ is a two-layer MLP with a SiLU
activation.
For mechanism family $m$, the Event-to-Modes Signature is
\begin{equation}
    S_{e,m,r}
    =
    \frac{
        \exp
        \left(
            L^{S}_{e,m,r}
        \right)
    }{
        \displaystyle
        \sum_{j=1}^{R_b}
        \exp
        \left(
            L^{S}_{e,m,j}
        \right)
    },
\label{eq:supp_mrc_signature}
\end{equation}
and therefore
\begin{equation}
    \mathbf{S}_{e,m}
    =
    \left[
        S_{e,m,1},
        \ldots,
        S_{e,m,R_b}
    \right]
    \in
    \Delta^{R_b-1}.
\label{eq:supp_mrc_signature_simplex}
\end{equation}
The signature specifies how event $e$, when attributed to
mechanism family $m$, distributes its driver across the
available response modes.
It depends on the fixed event attributes and learnable
signature mapping, but remains independent of the current
state, process controls, and event intensities.

\noindent\textbf{Mechanism-Mode Drive.}
The event-mechanism drivers are aggregated using the
Event-to-Modes Signatures:
\begin{equation}
    P_{t,m,r}
    =
    \sum_{e=1}^{N_e}
    T_{t,e,m}
    S_{e,m,r}.
\label{eq:supp_mrc_mode_drive}
\end{equation}
Because
$T_{t,e,m}\geq 0$
and
$S_{e,m,r}\geq 0$,
the resulting mechanism-mode drive satisfies
$P_{t,m,r}\geq 0$.
This aggregation compresses the
$N_e\times M$
event-level drivers into
$M R_b$
mechanism-mode drives while retaining event-specific response
preferences through $S_{e,m,r}$.
The scalar $P_{t,m,r}$ is the component-wise realization of
the tensorized mechanism-mode drive
$\mathbf{P}_{t,m,r}$
used in the main paper.

\noindent\textbf{State-Control Realizability Gate.}
The effective response associated with a mechanism-mode drive
can vary with the current system state and process controls.
MRC therefore predicts mode-wise gate logits:
\begin{equation}
\begin{gathered}
    \mathbf{C}_t
    =
    h_{\mathrm{gate}}
    \left(
        \left[
            \bar{\mathbf{z}}_t,\,
            \mathbf{u}_t
        \right]
    \right)
    \in
    \mathbb{R}^{M\times R_b},\\
    G_{t,m,r}
    =
    \sigma
    \left(
        C_{t,m,r}
    \right).
\end{gathered}
\label{eq:supp_mrc_state_control_gate}
\end{equation}
Here, $h_{\mathrm{gate}}$ is a two-layer MLP with a SiLU
activation, and
\begin{equation}
    0
    <
    G_{t,m,r}
    <
    1.
\label{eq:supp_mrc_gate_bound}
\end{equation}
The gate modulates the realizable fraction of each
mechanism-mode response under the current state and process
controls.
The scalar $G_{t,m,r}$ is the component-wise realization of
the tensorized state-control gate
$\mathbf{G}_{t,m,r}$
in the main paper.

\noindent\textbf{Saturating Mode Response.}
To prevent the response amplitude from increasing without
bound as the mechanism-mode drive grows, MRC uses a
nonnegative saturating response law.
For each mechanism-mode pair,
\begin{equation}
\begin{gathered}
    A_{m,r}
    =
    \operatorname{softplus}
    \left(
        a^{\mathrm{raw}}_{m,r}
    \right)
    +
    \epsilon,\\
    K_{m,r}
    =
    \operatorname{softplus}
    \left(
        k^{\mathrm{raw}}_{m,r}
    \right)
    +
    \epsilon.
\end{gathered}
\label{eq:supp_mrc_saturation_parameters}
\end{equation}
The realized mode amplitude is
\begin{equation}
    w_{t,m,r}
    =
    A_{m,r}
    \frac{
        P_{t,m,r}
    }{
        K_{m,r}
        +
        P_{t,m,r}
        +
        \epsilon
    }
    G_{t,m,r}.
\label{eq:supp_mrc_saturating_response}
\end{equation}
Here, $K_{m,r}$ determines the drive scale at which
saturation occurs,
$A_{m,r}$ defines the upper response capacity before
state-control gating, and
$G_{t,m,r}$ determines the realizable fraction of that
capacity.
The amplitude is nonnegative and bounded:
\begin{equation}
    0
    \leq
    w_{t,m,r}
    <
    A_{m,r}.
\label{eq:supp_mrc_response_bound}
\end{equation}

The scalar $w_{t,m,r}$ is the component-wise realization of
the tensorized mode amplitude
$\mathbf{w}_{t,m,r}$
used in the main paper.

\noindent\textbf{State-Dependent Spatial Response Modes.}
MRC generates the spatial response patterns from the current
latent state $\mathbf{z}_t$.
A shared convolutional trunk first computes
\begin{equation}
    \mathbf{H}_t
    =
    \operatorname{SiLU}
    \left(
        \operatorname{GN}
        \left(
            \operatorname{Conv}_{3\times3\times3}
            \left(
                \mathbf{z}_t
            \right)
        \right)
    \right).
\label{eq:supp_mrc_shared_trunk}
\end{equation}
Each mechanism family then uses an independent pointwise
head:
\begin{equation}
    \widetilde{\mathbf{B}}_{t,m}
    =
    \operatorname{Conv}^{m}_{1\times1\times1}
    \left(
        \mathbf{H}_t
    \right).
\label{eq:supp_mrc_spatial_modes}
\end{equation}
The output is reshaped into the mode fields
$\{
\widetilde{\mathbf{B}}_{t,m,r}
\}_{r=1}^{R_b}$,
where each field has the same channel and spatial dimensions
as $\mathbf{z}_t$.

To separate the spatial pattern from its
transition-specific amplitude, each mode field is normalized
by its root-mean-square magnitude:
\begin{equation}
\begin{gathered}
    \nu_{t,m,r}
    =
    \left[
        \operatorname{Mean}_{c,i,j,k}
        \left(
            \widetilde{B}_{t,m,r,c,i,j,k}^{\,2}
        \right)
        +
        \epsilon
    \right]^{1/2},\\
    \mathbf{B}_{t,m,r}
    =
    \frac{
        \widetilde{\mathbf{B}}_{t,m,r}
    }{
        \nu_{t,m,r}
    }.
\end{gathered}
\label{eq:supp_mrc_normalized_basis}
\end{equation}
Thus,
$\mathbf{B}_{t,m,r}$
represents the state-dependent spatial response pattern,
whereas
$w_{t,m,r}$
controls its transition-specific amplitude.
This normalization reduces scale ambiguity between the
spatial response field and its scalar amplitude.

\noindent\textbf{Latent Response Composition.}
The mechanism-mode responses are composed into the latent
increment:
\begin{equation}
    \Delta\mathbf{z}_t
    =
    \sum_{m=1}^{M}
    \sum_{r=1}^{R_b}
    w_{t,m,r}
    \mathbf{B}_{t,m,r}.
\label{eq:supp_mrc_latent_increment}
\end{equation}
The updated latent representation is
\begin{equation}
    \widetilde{\mathbf{z}}_{t+1}
    =
    \mathbf{z}_t
    +
    \Delta\mathbf{z}_t.
\label{eq:supp_mrc_latent_update}
\end{equation}
The residual decoder subsequently predicts the state-space
increment:
\begin{equation}
    \widehat{\mathbf{x}}_{t+1}
    =
    \mathbf{x}_t
    +
    \mathrm{Dec}
    \left(
        \widetilde{\mathbf{z}}_{t+1},\,
        \mathbf{b}_t,\,
        \mathbf{x}_t
    \right),
\label{eq:supp_mrc_state_update}
\end{equation}
where $\mathbf{b}_t$ denotes the boundary feature produced by
the State Encoder.

The intermediate quantities are not assumed to be uniquely
identifiable microscopic variables.
Instead, they have constrained executable roles:
$S_{e,m,r}$ distributes event-mechanism drivers across
response modes,
$P_{t,m,r}$ aggregates their mechanism-mode contributions,
$G_{t,m,r}$ controls state-control realizability,
$w_{t,m,r}$ determines the bounded response amplitude, and
$\mathbf{B}_{t,m,r}$ provides the spatial response pattern.
Their functional contribution is examined through the
structured component ablations and mechanism-aligned
counterfactual analyses reported in the experiments.

\section{Experiments}
\label{sec:supp_experiments}

\subsection{Additional Benchmark Results}
\label{sec:supp_additional_benchmark_results}

We report additional results for the single-axis Control-OOD
settings and the Vis-60\% temporal-extrapolation setting.

\begin{table*}[t]
\centering
\footnotesize
\setlength{\tabcolsep}{5pt}

\begin{tabular}{llccc}
\toprule
Setting
& Method
& Avg. RMSE $\downarrow$
& Final RMSE $\downarrow$
& CD $\downarrow$ \\
\midrule

\multirow{6}{*}{OOD-Ar$^{+}$}
& ConvLSTM & 0.7710 & 0.8722 & 3.575 \\
& FNO      & 0.6109 & 0.7843 & 3.398 \\
& U-NO     & 0.5835 & 0.7687 & 3.272 \\
& CNO      & 0.2535 & 0.4253 & 3.241 \\
& DeepONet & 0.1055 & 0.0833 & 0.755 \\
& \textbf{NEMSim (ours)}
& \textbf{0.0218}
& \textbf{0.0509}
& \textbf{0.586} \\
\midrule

\multirow{6}{*}{OOD-Cl$^{+}$}
& ConvLSTM & 0.6237 & 1.0170 & 2.954 \\
& FNO      & 0.4964 & 0.6046 & 3.143 \\
& U-NO     & 0.9975 & 1.5883 & 3.535 \\
& CNO      & 0.1951 & 0.3920 & 1.989 \\
& DeepONet & 0.0922 & 0.1988 & 1.024 \\
& \textbf{NEMSim (ours)}
& \textbf{0.0268}
& \textbf{0.0642}
& \textbf{0.611} \\
\midrule

\multirow{6}{*}{Vis-60\%}
& ConvLSTM & 0.6333 & 0.8904 & 2.826 \\
& FNO      & 0.4677 & 0.5956 & 3.258 \\
& U-NO     & 0.5933 & 0.8139 & 3.796 \\
& CNO      & 0.0970 & 0.2287 & 1.450 \\
& DeepONet & 0.0847 & 0.1671 & 0.836 \\
& \textbf{NEMSim (ours)}
& \textbf{0.0180}
& \textbf{0.0661}
& \textbf{0.672} \\

\bottomrule
\end{tabular}

\caption{
Additional 100-step autoregressive rollout results under
single-axis Control-OOD and Vis-60\% temporal extrapolation.
The best result in each setting is shown in bold.
}
\label{tab:supp_additional_benchmark_results}
\end{table*}

NEMSim achieves the best performance across all three
additional settings and all reported metrics.
Compared with the strongest baseline in Avg. RMSE, NEMSim
reduces the error by 79.3\% under OOD-Ar$^{+}$, 70.9\%
under OOD-Cl$^{+}$, and 78.7\% under Vis-60\%.
These results complement the OOD-Joint and Vis-50\%
evaluations in the main paper, showing that the advantage
persists under both single-axis control extrapolation and a
less restrictive temporal-visibility setting.

\subsection{Computational Efficiency}
\label{sec:supp_efficiency}

\noindent\textbf{Profiling Protocol.}
We profile the one-step inference efficiency of all models on a single
NVIDIA H100 80GB HBM3 GPU using PyTorch 2.7.1 and CUDA 12.6.
To provide a controlled comparison across architectures, all models are
evaluated with a batch size of one and a standardized input resolution
of $2\times64\times64\times64$, where the two channels represent the
signed-distance fields.
Model parameters and inputs are stored in FP32, with TF32 Tensor Core
execution enabled and automatic mixed precision disabled.

Each model is warmed up for $50$ iterations, followed by $200$ measured
one-step forward passes.
Latency is measured using CUDA events, with
\texttt{torch.cuda.synchronize()} called before and after measurement.
The reported latency includes only model inference and excludes data
loading, host-to-device transfer, metric computation, and
post-processing.
Throughput is calculated as
\begin{equation}
    \mathrm{Throughput}
    =
    \frac{1000B}{t_{\mathrm{mean}}},
\end{equation}
where $B=1$ is the batch size and $t_{\mathrm{mean}}$ is the mean
latency in milliseconds.

Peak allocated GPU memory is measured after warm-up by resetting the
PyTorch peak-memory statistics and executing one complete synchronized
forward pass.
The reported value includes model parameters, inputs, and forward
activations.

\begin{table*}[!t]
\centering

\begingroup
\footnotesize
\setlength{\tabcolsep}{4pt}
\renewcommand{\arraystretch}{1.08}

\begin{tabular}{lccccc}
\toprule
Method
& GFLOPs/step
& Mean (ms) $\downarrow$
& P95 (ms) $\downarrow$
& Throughput (s$^{-1}$) $\uparrow$
& Peak mem. (MiB) $\downarrow$ \\
\midrule

NEMSim
& 41.688
& 11.439
& 11.442
& 87.42
& 330.380 \\

ConvLSTM
& 331.388
& 4.541
& 4.970
& 220.19
& 632.422 \\

CNO
& 159.410
& 10.960
& 11.218
& 91.24
& 336.233 \\

U-NO
& 1.061
& 3.579
& 3.620
& 279.43
& 183.034 \\

FNO
& 4.473
& 2.149
& 2.154
& 465.41
& 214.918 \\

DeepONet
& 131.081
& 3.091
& 3.104
& 323.51
& 590.570 \\

\bottomrule
\end{tabular}

\endgroup

\caption{
One-step inference efficiency at the standardized
$2\times64\times64\times64$ input resolution.
Latency and throughput are measured with a batch size of one.
Mean and P95 report latency in milliseconds.
Throughput is measured in samples per second.
Peak memory denotes the maximum GPU memory allocated by PyTorch
during one synchronized forward pass.
}
\label{tab:supp_efficiency}
\end{table*}

\noindent\textbf{Results.}
As shown in Table~\ref{tab:supp_efficiency}, \method{} is not the
fastest architecture in one-step inference, but maintains moderate
latency and memory usage despite executing the complete
event-mechanism compilation pathway.
Its peak allocated memory is comparable to CNO and substantially lower
than ConvLSTM and DeepONet.
\method{} also requires fewer floating-point operations than ConvLSTM,
CNO, and DeepONet, although its latency remains higher than the lighter
FNO and U-NO baselines.
This difference indicates that measured latency depends not only on the
operation count but also on the operator composition and execution
overhead of the structured compilation modules.

\subsection{Knowledge-Augmented Baselines}
\label{sec:supp_rulefeat}

\noindent\textbf{Experimental Objective.}
This experiment examines whether the performance gains of
\method{} can be explained by feature-level access to
rule-derived information, rather than by compiling that
information into an executable transition structure.
We construct a deterministic, control-conditioned RuleFeat
representation from the same Event Rules Library used by
\method{} and integrate it into CNO and DeepONet.
The resulting variants are denoted as CNO+Prior and
DeepONet+Prior, respectively.
RuleFeat provides compact rule-derived quantities while
preserving the native transition architectures of the two
baselines.

\noindent\textbf{RuleFeat Construction.}
The Event Rules Library contains $E=37$ ordered events and $M=3$
mechanism families.
For each event $e$, we calculate a formula-based event intensity
$\lambda^{\mathrm{RF}}_{t,e}$ using its rule-defined control bindings,
probability or rate expression, yield expression, and product-mediated
dependencies.
For product-dependent deposition events, the deterministic product
relations in the library are first evaluated to construct the
corresponding Product Pool.
The Product Pool is used internally to calculate the dependent event
intensities but is not directly supplied to either baseline.

The resulting RuleFeat vector is defined as
\begin{equation}
\begin{aligned}
G^{\mathrm{RF}}_{t,e,m}
&=
\lambda^{\mathrm{RF}}_{t,e}\alpha^{0}_{e,m},\\
\mathbf{f}^{\mathrm{rule}}_{t}
&=
\left[
\boldsymbol{\lambda}^{\mathrm{RF}}_{t};
\operatorname{vec}\!\left(\mathbf{G}^{\mathrm{RF}}_{t}\right)
\right]
\in\mathbb{R}^{148},
\end{aligned}
\label{eq:rulefeat}
\end{equation}
where
$\boldsymbol{\lambda}^{\mathrm{RF}}_{t}\in\mathbb{R}^{37}$ contains the
control-conditioned event intensities,
$\boldsymbol{\alpha}^{0}\in\mathbb{R}^{37\times3}$ denotes the static
event--mechanism prior, and
$\mathbf{G}^{\mathrm{RF}}_{t}\in\mathbb{R}^{37\times3}$ contains their
element-wise interactions.
The events follow the fixed ordering specified by the Event Rules
Library.
The total RuleFeat dimension is therefore
$37+37\times3=148$.

RuleFeat depends on the normalized control vector but not on the current
morphology state.
Because the control vector is fixed within each benchmark trajectory,
RuleFeat remains constant throughout the corresponding autoregressive
rollout, although it is deterministically recomputed during each forward
pass.
Its construction contains no trainable parameters or random operations.
Raw event names, projectile and target identities, semantic descriptor
flags, explicit source--sink tensors, and intermediate Product Pool
features are not directly supplied to the knowledge-augmented
baselines.

\noindent\textbf{CNO+Prior.}
For CNO+Prior, denoted as CNO+RuleFeat in the implementation, the
$148$-dimensional RuleFeat vector is processed by a trainable
$148\!\rightarrow\!64\!\rightarrow\!8$ projection network with a GELU
activation. The resulting eight-dimensional representation is spatially
broadcast to match the morphology-field resolution. It is then
concatenated with the two-channel morphology state, the eight-channel
control embedding, and three coordinate channels before the CNO lifting
block. The input consequently increases from $13$ to $21$ channels,
whereas the channel widths, depth, and subsequent CNO blocks remain
unchanged. RuleFeat is injected only once at the model input.

The RuleFeat projection contains $10{,}056$ trainable parameters.
Expanding the first $3\times3\times3$ convolution in the lifting block
from $13$ to $21$ input channels introduces an additional $6{,}912$
parameters. CNO+Prior therefore adds $16{,}968$ parameters in total,
increasing the parameter count from $1{,}733{,}908$ for CNO to
$1{,}750{,}876$ for CNO+Prior.

\noindent\textbf{DeepONet+Prior.}
For DeepONet+Prior, denoted as DeepONet+RuleFeat in the implementation,
the $148$-dimensional RuleFeat vector is processed by a trainable
$148\!\rightarrow\!64\!\rightarrow\!32$ projection network with a GELU
activation.
The resulting 32-dimensional representation is concatenated with the
$14$-dimensional control vector and supplied to the condition encoder of
the DeepONet branch network.
The state encoder, coordinate-based trunk network, branch--trunk inner
product, and output head remain unchanged.
RuleFeat is therefore available only to the branch pathway and is not
directly provided to the trunk network.

The RuleFeat projection and the corresponding increase in the condition
encoder input introduce $15{,}712$ trainable parameters.
The total parameter count therefore increases from $1{,}720{,}162$ for
the standard DeepONet to $1{,}735{,}874$ for DeepONet+Prior.

\noindent\textbf{Controlled Comparison.}
For each baseline family, the standard and knowledge-augmented variants
use the same underlying backbone.
The only architectural difference is the RuleFeat projection and fusion
operation described above.
Both variants receive the same two-channel morphology state and the same
$14$-dimensional normalized control vector.
They also use identical trajectory identifiers, field supervision, and
evaluation metrics within each experimental setting.

CNO+Prior and DeepONet+Prior receive a deterministic representation
derived from the same Event Rules Library used by \method{}.
However, they do not receive the complete semantic descriptor tensors
of \method{} and do not instantiate its executable EIC--MAC--MRC
pathway.
In particular, the knowledge-augmented baselines do not perform learned
state-dependent event-intensity modulation, dynamic event-to-mechanism
attribution, explicit mechanism-mode response composition, or
saturating response execution.
The comparison therefore separates the benefit of accessing
rule-derived information from the benefit of compiling that information
into a structured event--mechanism transition structure.

\noindent\textbf{Evaluation Protocol.}
The comparison is conducted under OOD-Joint and Vis-50\%,
using the same trajectory-level partitions as the
corresponding main experiments.
OOD-Joint contains $202$ training, $50$ validation, $53$
test, and $195$ unused trajectories, whereas Vis-50\% uses
the fixed $300/100/100$ train/validation/test partition and
exposes the first $50$ transitions of each training
trajectory.

All standard, knowledge-augmented, and NEMSim variants use
identical trajectory identifiers within each setting.
Evaluation follows the common protocol in Sec.~S2.5: prediction starts from the
ground-truth state at step $0$ and proceeds through a
$100$-step autoregressive rollout without subsequent
ground-truth correction, teacher forcing, or prediction
clipping.
We report Avg. RMSE, Final RMSE, and final-step Chamfer
Distance.

\subsection{Data Efficiency}
\label{sec:supp_data_efficiency}

\noindent\textbf{Experimental Protocol.}
We evaluate \method{}, DeepONet, and DeepONet+Prior using nested subsets
of the fixed full-500 OOD-Joint training partition.
The complete partition contains $202$ training, $50$ validation, $53$
test, and $195$ unused trajectories.
The $10\%$, $20\%$, $50\%$, and $100\%$ settings retain $20$, $40$,
$101$, and $202$ training trajectories, respectively.
Each selected trajectory contributes all $100$ consecutive one-step
transitions; therefore, the fractions refer exclusively to the number
of training trajectories rather than temporal or spatial subsampling.

The reduced subsets are constructed by stratifying the training
trajectories using a $3\times3$ grid over the log-transformed Ar$^{+}$
and Cl$^{+}$ incident-flux controls.
Deterministic sampling with seed $0$ produces nested subsets satisfying
\begin{equation}
    \mathcal{D}_{10\%}
    \subset
    \mathcal{D}_{20\%}
    \subset
    \mathcal{D}_{50\%}
    \subset
    \mathcal{D}_{100\%}.
\end{equation}
All models use exactly the same trajectory identifiers at each data
fraction, while the validation and test partitions remain fixed at
$50$ and $53$ trajectories, respectively.
Model architectures are kept unchanged across all fractions.
The maximum training budget is fixed at $350$ epochs, with an initial
learning rate of $5\times10^{-4}$ and a minimum learning-rate floor of
$1\times10^{-5}$; the number of optimizer steps is therefore not matched
across data fractions.

Evaluation follows the standard full-500 OOD-Joint protocol.
Each model performs a $100$-step autoregressive rollout on the same $53$
test trajectories, starting from the ground-truth state at step $0$ and
without subsequent ground-truth correction.

\begin{table}[t]
\centering
\footnotesize
\setlength{\tabcolsep}{3.2pt}
\begin{tabular}{c l c c c}
\toprule
Data & Method
& Avg. RMSE $\downarrow$
& Final RMSE $\downarrow$
& CD $\downarrow$ \\
\midrule

\multirow{3}{*}{$10\%$}
& DeepONet
& 0.1955 & 0.2853 & 2.609 \\
& DeepONet+Prior
& 0.1975 & 0.2660 & 2.902 \\
& \textbf{\method{}}
& \textbf{0.0967} & \textbf{0.2097} & \textbf{1.167} \\
\midrule

\multirow{3}{*}{$20\%$}
& DeepONet
& 0.1906 & 0.2239 & 2.924 \\
& DeepONet+Prior
& 0.1755 & 0.2331 & 1.978 \\
& \textbf{\method{}}
& \textbf{0.0902} & \textbf{0.1936} & \textbf{1.095} \\
\midrule

\multirow{3}{*}{$50\%$}
& DeepONet
& 0.1533 & 0.2208 & 1.394 \\
& DeepONet+Prior
& 0.1708 & 0.2113 & 1.416 \\
& \textbf{\method{}}
& \textbf{0.0861} & \textbf{0.1863} & \textbf{1.047} \\
\midrule

\multirow{3}{*}{$100\%$}
& DeepONet
& 0.1243 & 0.2120 & 1.122 \\
& DeepONet+Prior
& 0.1165 & 0.2161 & 1.088 \\
& \textbf{\method{}}
& \textbf{0.0511} & \textbf{0.1141} & \textbf{0.851} \\
\bottomrule
\end{tabular}
\caption{
Data-efficiency results under OOD-Joint.
Training-data fractions denote percentages of the original training
trajectories.
}
\label{tab:supp_data_efficiency}
\end{table}

\noindent\textbf{Results.}
As shown in Table~\ref{tab:supp_data_efficiency}, \method{} achieves the
lowest Avg. RMSE, Final RMSE, and CD at every training-data fraction.
With only $10\%$ of the training trajectories, it obtains an Avg. RMSE
of $0.0967$, which remains lower than DeepONet and DeepONet+Prior trained
with the complete training partition.
Its persistent advantage over DeepONet+Prior indicates that access to
rule-derived features alone does not reproduce the data efficiency of
the executable event--mechanism transition structure.

\subsection{Robustness to Perturbed Event Rules}
\label{sec:supp_rule_robustness}

\noindent\textbf{Experimental Protocol.}
We evaluate the robustness and functional contribution of the encoded
event knowledge by perturbing different components of the Event Rules
Library.
All variants retain the same model topology, parameter capacity,
training data, and evaluation protocol as the full model.
Each ablated or perturbed variant is retrained from scratch, and its
configuration remains unchanged throughout training, validation, and
testing.
Each randomized configuration is generated once using a fixed random
seed to ensure reproducibility.

\noindent\textbf{\method{} (Full).}
This variant uses the complete Event Rules Library without perturbation,
including the original rate and yield parameters, event descriptors,
event-mechanism priors, product dependencies, and mechanism-role
attributes.
It serves as the reference model.

\noindent\textbf{Structure-only.}
This variant retains the same model topology and parameter capacity
while removing the physical and chemical semantics encoded by the
Event Rules Library.
The rule-derived event descriptors and event parameters are replaced by
fixed randomly initialized embeddings, the event-mechanism prior is
replaced by a uniform prior, and the original product and dependency
metadata are removed.
Consequently, the internal event and mechanism slots no longer
correspond to specific physical or chemical processes.
The randomly initialized rule representations remain fixed during
training, while the neural components are trained normally.
This variant tests whether the structured computational architecture
alone is sufficient without meaningful event-rule information.

\noindent\textbf{Mechanism Shuffle.}
We randomly select approximately \(30\%\) of the events
(\(12\) out of \(37\)) and reassign the dominant entries of their
static event-mechanism priors to different mechanism families.
Each modified prior row remains normalized.
Event rate and yield expressions, event descriptors, product
dependencies, deterministic mechanism-role signals, and the model
architecture remain unchanged.
This perturbation disrupts the correspondence between events and their
intended mechanism priors while preserving the dimensionality and
normalization of the prior matrix.

\noindent\textbf{Event Drop.}
We randomly select approximately \(30\%\) of the events
(\(12\) out of \(37\)) and deactivate them by setting their
event-intensity masks to zero.
The affected event intensities therefore remain zero throughout model
execution.
Because the masking is applied before product aggregation and
dependency propagation, the removed events no longer contribute to
event-mechanism drivers, generated products, downstream
product-mediated events, mechanism responses, or the final state
update.
The event slots and network parameterization are retained, so this
perturbation changes the available event knowledge without changing
model capacity.

\noindent\textbf{Rate/Yield Noise.}
We independently perturb the positive continuous parameters appearing
in the event rate and yield expressions using multiplicative noise.
Each affected parameter \(p\) is replaced by
\begin{equation}
    \widetilde{p}
    =
    p\left(1+\epsilon\right),
    \qquad
    \epsilon\sim\mathcal{U}(-0.3,0.3).
\end{equation}
The perturbed quantities include the continuous threshold, reference,
exponent, probability, yield, angular-response, and sticking
parameters used by the event rules.
Parameters with bounded physical ranges are clipped to their valid
intervals.
Event identities, categorical expression types, control dependencies,
product dependencies, and event-mechanism assignments remain
unchanged.
This setting evaluates robustness to imperfect numerical rule
parameters without modifying the underlying rule structure.

\noindent\textbf{Evaluation.}
All variants are evaluated on the same OOD-Joint test partition using
autoregressive rollout from the ground-truth state at step \(0\).
No ground-truth correction is applied after initialization.
We report average rollout RMSE, final-step RMSE, and final-step Chamfer
distance.
Comparison with \method{} (Full) assesses the contribution of correct
event semantics beyond the retained model topology and evaluates the
sensitivity of the model to missing, mismatched, or numerically
inaccurate event knowledge.

\subsection{Mechanism-Aligned Counterfactual Analysis}
\label{sec:supp_mechanism_alignment}

We perform a counterfactual control-intervention experiment to examine
the correspondence between process controls, internal mechanism
responses, and predicted state changes. The experiment uses the frozen
best-rollout checkpoint trained under the Control-ID setting. All $100$
test trajectories from the fixed full-$500$ IID partition are included,
and no model parameters are updated during this analysis.

For each trajectory, the ground-truth state at step $0$ is used as the
common initial state. The model first performs a one-step prediction
using the original control vector. Additional one-step predictions are
then obtained after perturbing the Cl$^+$ incident flux, the neutral Cl
incident flux, or both. All other control variables and the initial
state remain unchanged.

Let $s_{+}$ and $s_{\mathrm{n}}$ denote the multiplicative factors
applied to the Cl$^+$ and neutral-Cl fluxes, respectively. We define
the intervention set as
\begin{equation}
    \mathcal{I}
    =
    \{0.5,1.0,1.5\}^{2}
    \setminus
    \{(1.0,1.0)\}.
    \label{eq:mechanism_interventions}
\end{equation}
Thus, the experiment contains eight interventions. A factor of $1.5$
denotes an increased flux, a factor of $0.5$ denotes a decreased flux,
and a factor of $1.0$ leaves the corresponding control unchanged. The
perturbed physical fluxes are processed using the same transformation
and range normalization employed during model training. Applying all
eight interventions to the $100$ test trajectories yields $800$
trajectory--intervention pairs.

For each mechanism family $m\in
\{\mathrm{physical},\mathrm{chemical},\mathrm{coupled}\}$, the
mechanism response strength is calculated by aggregating the
event--mechanism transition intensities:
\begin{equation}
    M_m
    =
    \sum_e T_{e,m},
    \label{eq:mechanism_response_strength}
\end{equation}
where $T_{e,m}$ is the routed contribution from event $e$ to mechanism
family $m$. The response change relative to the unperturbed prediction
from the same trajectory is
\begin{equation}
    \Delta M_m(\%)
    =
    100
    \frac{
        M_m^{\mathrm{cf}}-M_m^{\mathrm{base}}
    }{
        M_m^{\mathrm{base}}
    },
    \label{eq:mechanism_response_change}
\end{equation}
where ``base'' and ``cf'' denote the original and counterfactually
perturbed controls, respectively. The reported response matrix is
obtained by averaging $\Delta M_m$ over the $100$ test trajectories for
each intervention and mechanism family.

To quantify the corresponding predicted state change, we construct a
continuous material-removal proxy from the silicon SDF channel. Negative
SDF values represent the solid silicon phase, and an increase in the
SDF value corresponds to motion toward or outside the solid region. For
a one-step prediction, the removal proxy is defined as
\begin{equation}
    R
    =
    \sum_{\mathbf{v}}
    \max\left(
        \widehat{\phi}^{\,\mathrm{Si}}_{1}(\mathbf{v})
        -
        \phi^{\mathrm{Si}}_{0}(\mathbf{v}),
        0
    \right),
    \label{eq:material_removal_proxy}
\end{equation}
where $\mathbf{v}$ indexes the spatial voxels. This continuous
definition captures SDF changes even when the predicted interface does
not cross the zero level set within one step. Its relative change is
computed as
\begin{equation}
    \Delta R(\%)
    =
    100
    \frac{
        R^{\mathrm{cf}}-R^{\mathrm{base}}
    }{
        R^{\mathrm{base}}
    }.
    \label{eq:material_removal_change}
\end{equation}

For the mechanism--state alignment analysis, the chemical response is
used for interventions that modify only the neutral-Cl flux. The
coupled response is used for interventions that modify the Cl$^+$ flux,
including the joint Cl$^+$--Cl interventions. Each scatter point
corresponds to one trajectory--intervention pair.

Pearson correlation is calculated separately for four intervention
groups. The Cl$^+$-increasing group contains the Cl$^+$ increase and
the two joint interventions in which Cl$^+$ is increased. The
Cl$^+$-decreasing group is constructed analogously. Each of these
groups contains $3\times100=300$ observations. The neutral-Cl-only
increase and decrease interventions are evaluated separately, with
$100$ observations in each group.

\subsection{Ablation Studies}
\begin{table}[!h]
\centering
{\footnotesize
\begin{tabular}{
@{}l
@{\hspace{4pt}}l
@{\hspace{4pt}}c
@{\hspace{4pt}}c
@{\hspace{4pt}}c
@{}
}
\toprule
Setting
& Variant
& Avg. RMSE $\downarrow$
& Final RMSE $\downarrow$
& CD $\downarrow$ \\
\midrule

\multirow{4}{*}{Control-ID}
& Full
& \textbf{0.0134}
& \textbf{0.0316}
& \textbf{0.519} \\

& w/o EIC
& 0.0612
& 0.1282
& 0.881 \\

& w/o MAC
& 0.0611
& 0.1234
& 0.828 \\

& w/o MRC
& 0.0600
& 0.1258
& 0.827 \\

\midrule

\multirow{4}{*}{OOD-Joint}
& Full
& \textbf{0.0511}
& \textbf{0.1141}
& \textbf{0.851} \\

& w/o EIC
& 0.1229
& 0.2549
& 1.207 \\

& w/o MAC
& 0.1232
& 0.2552
& 1.216 \\

& w/o MRC
& 0.1224
& 0.2526
& 1.217 \\

\midrule

\multirow{4}{*}{Vis-50\%}
& Full
& \textbf{0.0195}
& \textbf{0.0659}
& \textbf{0.621} \\

& w/o EIC
& 0.0709
& 0.1583
& 1.542 \\

& w/o MAC
& 0.0797
& 0.1880
& 1.247 \\

& w/o MRC
& 0.0861
& 0.2254
& 1.649 \\

\bottomrule
\end{tabular}
}
\caption{
Ablation results under Control-ID, OOD-Joint, and Vis-50\% settings.
}
\label{tab:supp_ablation_results}
\end{table}

Table~\ref{tab:supp_ablation_results} evaluates the contribution of the main structured components in \method{} under Control-ID, OOD-Joint, and Vis-50\% settings.
\textbf{w/o EIC} removes event intensity compiler, so event activities are no longer explicitly compiled from event-rule attributes and process controls.
\textbf{w/o MAC} removes event-to-mechanism attribution, so activated events are not routed through mechanism families before response composition.
\textbf{w/o MRC} removes the response-compilation path before latent modes update, including state--control gating, event-to-mode signatures, mechanism-mode drives, and the saturating response law, while keeping the subsequent latent update and decoder path.

Removing any structured component consistently degrades performance across all settings, with larger drops under OOD-Joint and Vis-50\%.
This indicates that event intensity compilation, event-to-mechanism attribution, and mechanism-response compilation jointly support robust extrapolative rollout in \method{}.

\section{Implementation and Reproducibility Details}
\label{sec:supp_reproducibility}

\subsection{Model and Training Configuration}
\label{sec:supp_training}

NEMSim employs a three-level 3D encoder with a stem width of $12$ and
pyramid widths of $(12,18,24)$. The encoded state is represented using
$18$ latent channels. The event responses are routed into three
mechanism families, corresponding to physical, chemical, and coupled
mechanisms, and the mechanism-conditioned response operator uses a
basis rank of $4$. The response readout has a hidden width of $24$.

The model is optimized using AdamW with an initial learning rate of
$5\times10^{-4}$ and a weight decay of $10^{-5}$. A linear warm-up is
applied for the first $10$ epochs, followed by cosine annealing to a
minimum learning rate of $10^{-5}$. Training minimizes the voxel-wise
mean squared error between the predicted and ground-truth next states.
Gradients are clipped to a maximum norm of $1.0$.

All experiments use a maximum budget of $350$ epochs with a batch size
of $24$. We use random seed $0$, perform validation every $10$ epochs, and apply early stopping with a patience of $20$ validation rounds.

\subsection{Baseline Implementations}
\label{sec:supp_baselines}

All baselines receive the same two-channel 3D state representation and
the same normalized $14$-dimensional control vector as \method{}.
Each architecture is adapted to operate on 3D volumetric
state-transition data.
Their principal architectural settings and control-injection strategies
are summarized in Table~\ref{tab:supp_baseline_impl}.

\begin{table*}[t]
    \centering
    \small
    \setlength{\tabcolsep}{4pt}
    \renewcommand{\arraystretch}{1.12}
    \caption{Implementation details of the data-driven baselines.
    Parameter counts include all trainable parameters. Complex-valued
    Fourier coefficients are counted as two real-valued parameters.}
    \label{tab:supp_baseline_impl}
    \begin{tabular}{p{1.7cm}p{4.3cm}p{4.3cm}p{3.2cm}c}
        \toprule
        Method
        & 3D adaptation
        & Control injection
        & Core configuration
        & Params (M) \\
        \midrule

        ConvLSTM
        & Four stacked ConvLSTM3D layers followed by a 3D convolutional readout.
        & The projected control vector is spatially broadcast and concatenated with every input state.
        & Four layers with hidden widths $(28,28,28,28)$.
        & 0.634 \\

        FNO
        & 3D Fourier layers operate along all three spatial dimensions, with a pointwise residual path in each operator block.
        & A projected control field is broadcast and concatenated with the input state before lifting.
        & Width $14$, four Fourier blocks, and $6\times6\times6$ retained modes.
        & 1.358 \\

        U-NO
        & The seven-layer U-shaped neural-operator topology is extended to 3D using dense 3D Fourier operators and multiresolution skip connections.
        & A projected control field is broadcast and concatenated with the state and spatial coordinates before lifting.
        & Widths $(8,16,32)$ and $4\times4\times4$ Fourier modes.
        & 2.432 \\

        CNO
        & The Lift--Encoder--ResNet--ED Link--Decoder--Project topology is extended to 3D. The implementation applies alias-controlled activations only at resolution-changing operations.
        & A projected control field is broadcast and concatenated with the state and spatial coordinates before the lifting block.
        & Widths $(22,44,88)$, two residual blocks per stage, and an oversampling factor of $1.5$.
        & 1.734 \\

        DeepONet
        & A 3D convolutional branch encoder is combined with a coordinate-based trunk network through a branch--trunk basis contraction.
        & The control vector is projected into a spatial feature map and concatenated with the state inside the branch encoder.
        & Branch widths $(32,64,128)$, basis dimension $192$, and a three-layer trunk with width $160$.
        & 1.720 \\

        \bottomrule
    \end{tabular}
\end{table*}

Details of the knowledge-augmented CNO+Prior and
DeepONet+Prior variants are provided in
Sec.~\ref{sec:supp_rulefeat}.

\subsection{Evaluation and Computational Details}
\label{sec:supp_evaluation}

\noindent\textbf{Evaluation and Checkpoint Selection.}
All models are trained exclusively with one-step supervision. Each
optimization step minimizes the prediction error from the state at
step $t$ to that at step $t+1$, and no multi-step rollout loss is used
for gradient-based optimization.

Validation is performed every $10$ epochs using a complete $100$-step
autoregressive rollout. The checkpoint achieving the lowest mean
validation-rollout MSE is retained for final evaluation. Thus, rollout
performance is used only for model selection and does not participate
in gradient-based training.

All evaluation rollouts begin from the ground-truth state at step $0$.
After initialization, each predicted state is recursively used as the
input to the next prediction step, without teacher forcing or subsequent
ground-truth correction. Every model is evaluated over a complete
$100$-step autoregressive rollout. This protocol is also used for
temporal extrapolation: although only the first $50$ transitions are
available during training in the Vis-$50\%$ setting, evaluation still
starts from step $0$ and continues for all $100$ prediction steps.
All data partitions are defined at the trajectory level over the full
set of $500$ cases.

All field and surface metrics follow the implementation and settings
defined in Sec.~S2.5.

\noindent\textbf{Computational Resources.}
All models were trained and evaluated on a single NVIDIA H100 GPU with
80\,GB of HBM3 memory.
The server is equipped with two AMD EPYC 9J14 96-core processors,
providing 192 physical cores and 384 logical CPUs, and 503\,GiB of
system memory.
The server runs Ubuntu~24.04.3 LTS (Noble Numbat).
The software environment uses Python~3.12.11, PyTorch~2.7.1 compiled
with CUDA~12.6, cuDNN~9.5.1, and NVIDIA driver~580.95.05.

\section{Additional Qualitative Results}
\label{sec:supp_qualitative}

We provide additional rollout visualizations under OOD-Joint and
Vis-50\%.
Each figure compares the predicted 3D surfaces and representative SDF
slices at steps $10$, $25$, $50$, $75$, and $100$.
Surface colors show signed deviations from the ground-truth surface.
The SDF panels show signed field errors together with the ground-truth
and predicted zero-level contours.

\begin{figure*}[!t]
    \centering
    \includegraphics[width=\textwidth]
    {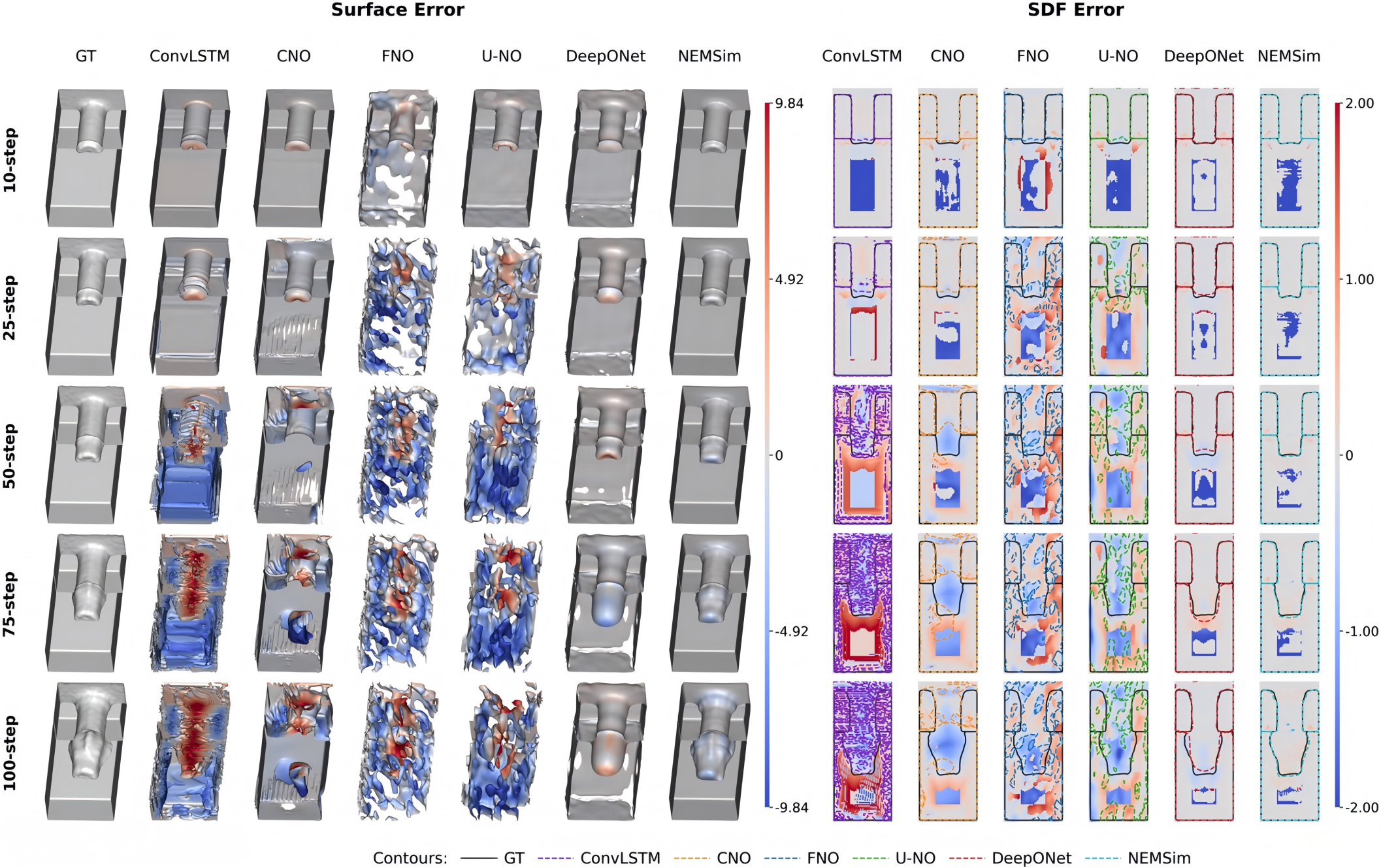}
    \caption{
    Additional qualitative comparison under OOD-Joint.
    Rows correspond to autoregressive rollout steps
    $10$, $25$, $50$, $75$, and $100$.
    The left block compares predicted surfaces colored by signed
    surface deviation, while the right block shows representative
    SDF errors and zero-level contours.
    NEMSim maintains closer agreement with the ground-truth morphology
    and contour throughout the rollout, whereas the baselines exhibit
    increasing bottom, sidewall, and global shape deviations.
    }
    \label{fig:supp_qualitative_ood}
\end{figure*}

Under OOD-Joint, the baseline errors accumulate progressively as the
rollout advances.
ConvLSTM and CNO develop substantial geometric deviations around the
trench bottom and sidewalls, while FNO and U-NO produce increasingly
irregular surface structures and spatially distributed SDF errors.
DeepONet preserves the overall morphology more effectively than these
baselines but still shows noticeable late-stage bottom and contour
misalignment.
In contrast, \method{} retains a coherent surface and keeps most
remaining errors localized near the evolving interface through step
$100$.

\begin{figure*}[!t]
    \centering
    \includegraphics[width=\textwidth]
    {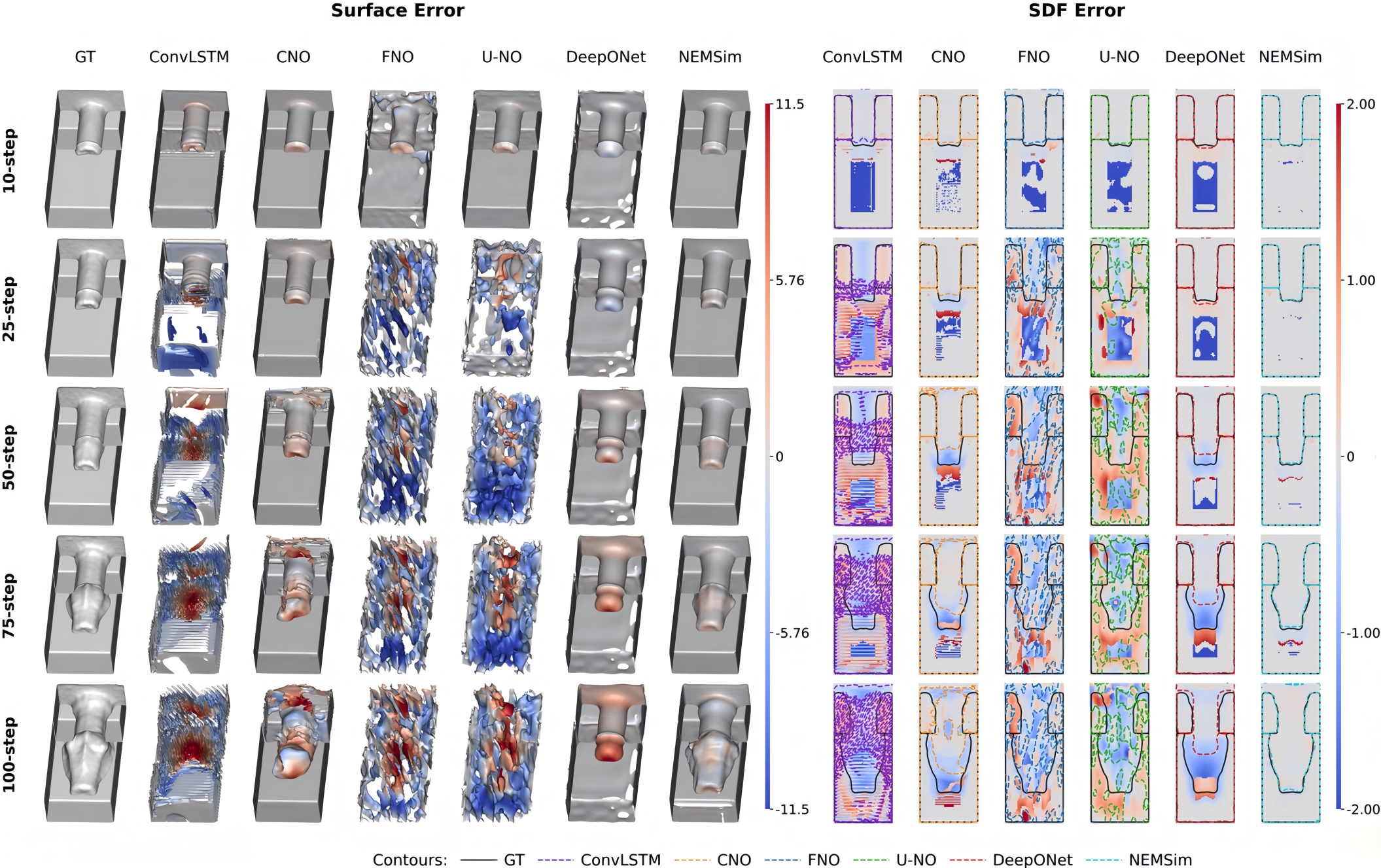}
    \caption{
    Additional qualitative comparison under Vis-50\%.
    Rows correspond to autoregressive rollout steps
    $10$, $25$, $50$, $75$, and $100$.
    Steps $75$ and $100$ lie beyond the temporal range exposed during
    training.
    The left block compares signed surface deviations, and the right
    block compares SDF errors and zero-level contours.
    NEMSim better preserves the evolving trench geometry after entering
    the temporally extrapolative stage.
    }
    \label{fig:supp_qualitative_vis50}
\end{figure*}

Under Vis-50\%, the differences become more pronounced after step $50$.
Several baselines undergo rapid error growth during the extrapolative
stage, producing distorted surfaces, unstable local structures, or
large contour displacement.
DeepONet continues to capture the coarse morphology but accumulates
systematic errors near the trench bottom.
\method{} remains substantially closer to the ground truth at steps
$75$ and $100$, with limited interface-localized errors and more
accurate bottom and sidewall contours.
These observations are consistent with the lower rollout-wide RMSE and
final-step geometric errors reported in the quantitative evaluation.

\end{document}